\pdfoutput=1

\documentclass[11pt]{article}

\usepackage[final]{emnlp2021}

\usepackage{times}
\usepackage{latexsym}

\usepackage[T1]{fontenc}

\usepackage[utf8]{inputenc}

\usepackage{microtype}

\usepackage{amsfonts}
\usepackage{amsmath}
\usepackage[pdftex]{graphicx}
\usepackage{caption}
\usepackage{subcaption}
\usepackage{multirow}
\usepackage{booktabs}
\usepackage{comment}

\title{Is BERT a Cross-Disciplinary Knowledge Learner? \\
A Surprising Finding of Pre-trained Models' Transferability}

\author{Wei-Tsung Kao \and Hung-yi Lee\\
   Graduate Institute of Communication Engineering, National Taiwan University \\
  \texttt{\{r09942067, hungyilee\}}@ntu.edu.tw\\}

\begin{document}
\maketitle
\begin{abstract}
This paper investigates whether the power of the models pre-trained on text data, such as BERT, can be transferred to general token sequence classification applications. To verify pre-trained models' transferability, we test the pre-trained models on text classification tasks with meanings of tokens mismatches, and real-world non-text token sequence classification data, including amino acid, DNA, and music. We find that even on non-text data, the models pre-trained on text converge faster, perform better than the randomly initialized models, and only slightly worse than the models using task-specific knowledge. We also find that the representations of the text and non-text pre-trained models share non-trivial similarities.
\end{abstract}

\section{Introduction}
In recent NLP research, pre-trained masked language models (MLMs) such as BERT~\citep{devlin-etal-2019-bert} are widely used by practitioners. After pre-trained on large corpora such as Wikipedia, the models can be fine-tuned quickly on NLP tasks like text classification and question answering and generalize well on small datasets such as RTE in GLUE~\citep{glue}. To apply and improve BERT in a more specialized domain such as scientific articles or clinical data, several MLMs are proposed by pre-training on the domain-specific text data~\citep{lee2020biobert, beltagy-etal-2019-scibert}. 
The concept of MLM can also be extended to other disciplines (maybe non-linguistic) as long as the input is discrete. For example, \citet{proteinBERT} pre-train MLMs called PLUS on amino acid sequence data and achieve state-of-the-art performance on several protein classification tasks. 

This paper examines whether the model pre-trained on large text corpora, such as BERT, can be efficiently adapted to data with numbers of tokens, token distribution, labels, and structure very different from natural language (the target data could even be non-text). We refer to this ability as \textbf{discipline adaptability}\footnote{We use the term \textit{discipline adaptation} instead of \textit{domain adaptation}. In NLP, domain adaptation usually refers to the setting like the transfer from general text to specialized text data such as scientific articles. We use the term \textit{discipline} to emphasize data with very different distribution and structure.}. Previous work \citep{papadimitriou-jurafsky-2020-learning} only shows that language models (LMs) pre-trained on non-text data can be adapted to LMs of human languages. This work is the first to examines if the pre-trained MLMs can learn the relation between the label and the data never seen during pre-training. 
Our contributions are the following.
\begin{itemize}
    \item We propose settings to examine the \textit{discipline} \textit{adaptability} of the pre-trained models. We find that BERT, BERT-Chinese, ALBERT, and RoBERTa can reduce training loss much more quickly, generalize better than the randomly initialized models on the non-text data, and are just slightly worse than the models using prior knowledge within each discipline. 
    \item Our analyses indicate that before fine-tuning, the similarity between BERT and the MLM-like model pre-trained on the non-text data is much higher than the one between BERT and the randomly-initialized model, which helps to explain the success of BERT within the non-text disciplines. Furthermore, our extensive investigation of several hypotheses about attention similarity, hierarchical structure in the non-text data, and training stability indicates that these hypotheses are not sufficient to explain the \textit{discipline adaptability} of BERT.
\end{itemize}
We believe the findings of \textit{discipline adaptation} will intrigue the NLP community to ponder what is learned in the pre-training procedure. 
The findings can also be helpful to the disciplines that large-scale datasets are not available, which are essential for practitioners.

\section{Method}
\label{sec:method}
To examine the \textit{discipline adaptability} of the models pre-trained on text corpora, we fine-tune the pre-trained models on two types of downstream data. 
The first type (section \ref{method:in}) are synthetic datasets, which are generated by permuting tokens in common NLP datasets. So the meaning of each token is changed. 
The second type of data (section \ref{method:out}) is a more challenging situation in which the downstream tasks are not relevant to human language. 

\subsection{Synthetic data}
\label{method:in}

\begin{figure}[t]
		\centering
		\begin{subfigure}[b]{0.45\linewidth}
			\centering
			\includegraphics[width=0.8\linewidth]{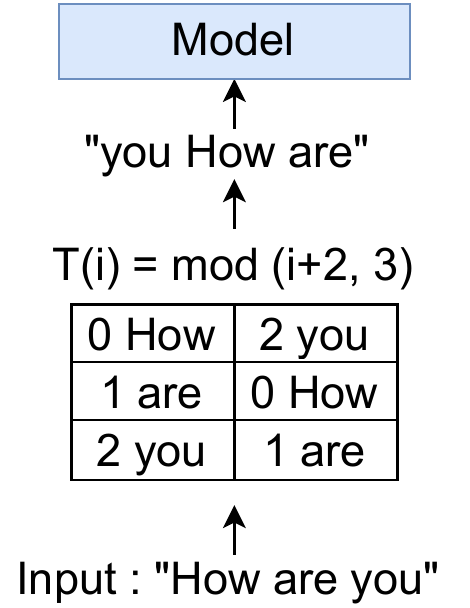}
			\caption{permutation on text data}
			\label{fig:shift_c}
		\end{subfigure}
		\begin{subfigure}[b]{0.45\linewidth}
			\centering
			\includegraphics[width=0.8\linewidth]{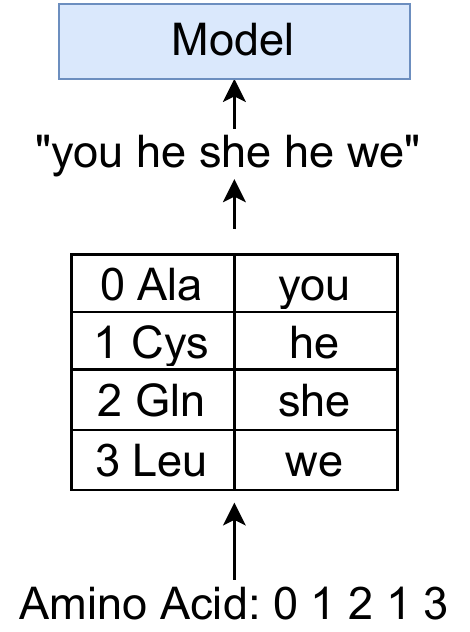}
			\caption{Amino acid sequence}
			\label{fig:non-text_input}
		\end{subfigure}
		\label{fig:method}
		\caption{Examples of (a) token permutation  
		(b) amino acid sequence input for protein classification.}
\end{figure}

We first test the models on synthetic text sequence classification data. 
The synthetic data is generated as below. Given a text sequence classification dataset, we define a deterministic one-to-one mapping $T$ that changes a subword token $x_i$ in a text sequence to another subword token $T(x_i)$, as shown in figure \ref{fig:shift_c}. 
To elaborate on the design of synthetic text data, consider the tokens in the text corpora as nodes in a graph and the relationship among tokens as edges.
Then the graph of the synthetic data is an isomorphism of the original graph. 
Hence, the structure of the synthetic data (the structure of the graph) is identical to the original one, and the tasks of the synthetic datasets are as difficult as the original tasks (if the pre-training procedure is not considered). 
Suppose the pre-trained model can still outperform the model trained from scratch on this artificial data. This indicates that the pre-trained models can transfer knowledge to the downstream tasks with meanings of tokens completely different from pre-trained corpora. 
And therefore, it is probable that we can further take advantage of the pre-trained model when processing real-world non-text data.

In our experiments, we first pre-train the model on normal text corpora, and then we fine-tune and test the model on the synthetic data. We choose $T(i) = (i+1000)\ \text{mod}\ D$, where $D$ is the vocabulary size of the model. We have also tried generating the mappings randomly. The results are similar and left in the appendix.
For a real example, the sentence "his healthy sense of satire is light and fun..." in GLUE dataset will be changed to "canadian franzme 1988pia leader watch sports czech at at at"\footnote{The token "." maps to the token "at" in the examples, so there are three consecutive "at" in the synthetic sentence.}. 

\subsection{Real-world non-text data}
\label{method:out}
To further validate the $\textit{discipline adaptability}$ of the pre-trained model, we fine-tune the pre-trained model on real-world non-text data. 
In these downstream tasks, both the token distributions and the number of tokens could be very different from the text data for pre-training. 
So this is a more difficult setting to evaluate the transferability of pre-trained models.

To process non-text data by BERT, we map each token of the non-text data to one subword token as in figure \ref{fig:non-text_input}. 
In the following experiments, the (deterministic) mapping table is generated randomly because we find that different mappings lead to similar results as long as we do not map the non-text tokens to the unused tokens of the pre-trained models. 
We add a randomly initialized linear classifier on top of the pre-trained model in the fine-tuning phase without randomly initializing any pre-trained parameters, including the embedding layer. Then we fine-tune the whole model. 

\section{Experiment}

\subsection{Setup}
\label{exp:setup}
We use GLUE dataset to generate the synthetic data. The validation sets are used to test the models. WNLI is excluded as in \citep{devlin-etal-2019-bert}; For the real-world non-text data, we include the following tasks with different numbers of tokens, token distributions, and structures:

Protein classification (3 tasks):  \textit{Localization} (Loc.)~\citep{localization}, \textit{Stability} (Stab.)~\citep{stability}, and \textit{Fluorescence} (Flu.)~\citep{fluorescence} used in~\citet{proteinBERT}. The input is amino acid sequences consisting of 20 different tokens.

DNA classification (4 tasks): \textit{H3}, \textit{H4},  \textit{H3K9ac} from~\citet{H3}, and \textit{Splice} from~\citet{splice} used in~\citet{hilbert-cnn}. The input is DNA sub-sequences consisting of 4 different tokens.

Music composer classification (1 task): We use \textit{MAESTRO-v1} dataset~\citep{music}. The input is pitch sequences consisting of 128 different tokens.

The pre-trained models used in the experiments include BERT-base-uncased, BERT-base-Chinese, ALBERT-base-v1, and RoBERTa-base. The randomly initialized (trained from scratch) models have the same architectures as BERT-base. The experiments on BERT-large are left in the appendix due to space limitations. The models are initialized by the default distribution widely adopted for pre-training the models (e.g., $\mathcal{N}(0, 4\times10^{-4} )$ for BERT-base). For detailed hyperparameters, please refer to the appendix. For simplicity, we use "pre-trained models" to refer to the above models pre-trained on the natural language if not specified.

\begin{figure}[t]
    \centering
    \begin{subfigure}[t]{0.49\linewidth}
			\centering
			\includegraphics[width=\linewidth]{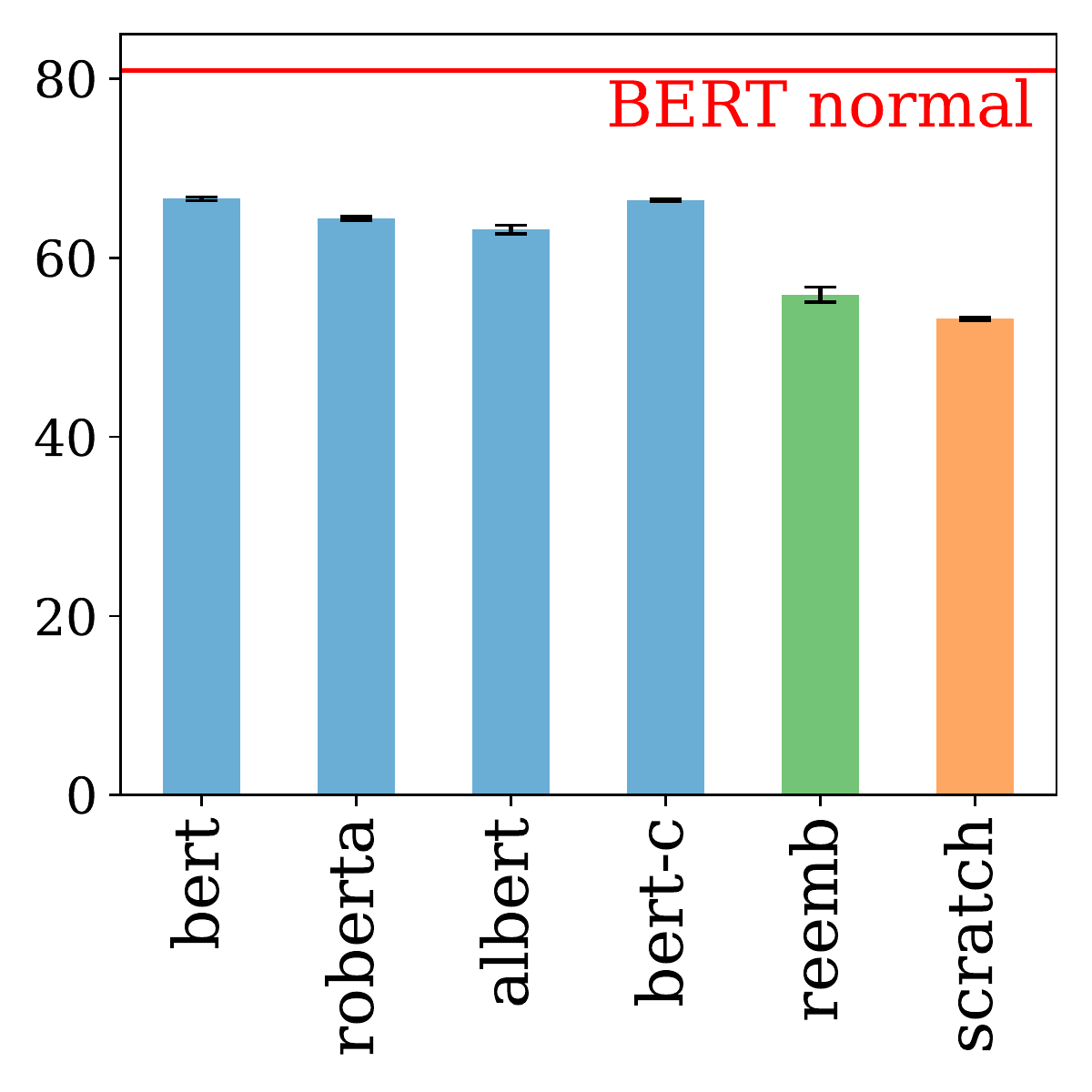}
			\caption{Synthetic GLUE (8 tasks)}
			\label{fig:glue}
		\end{subfigure}
		\begin{subfigure}[t]{0.49\linewidth}
			\centering
			\includegraphics[width=\linewidth]{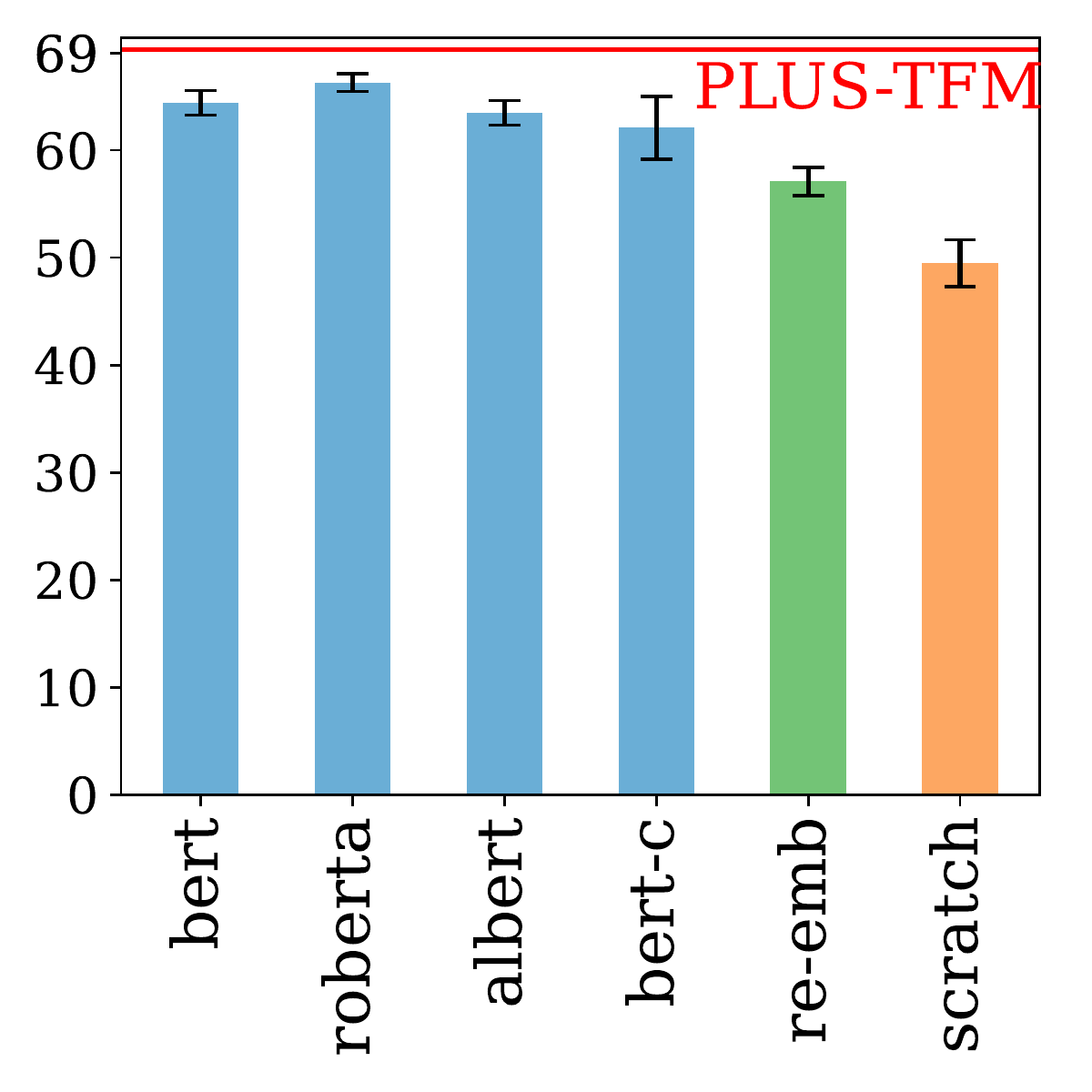}
			\caption{Protein (3 tasks)}
			\label{fig:protein}
		\end{subfigure}
		\begin{subfigure}[t]{0.49\linewidth}
			\centering
			\includegraphics[width=\linewidth]{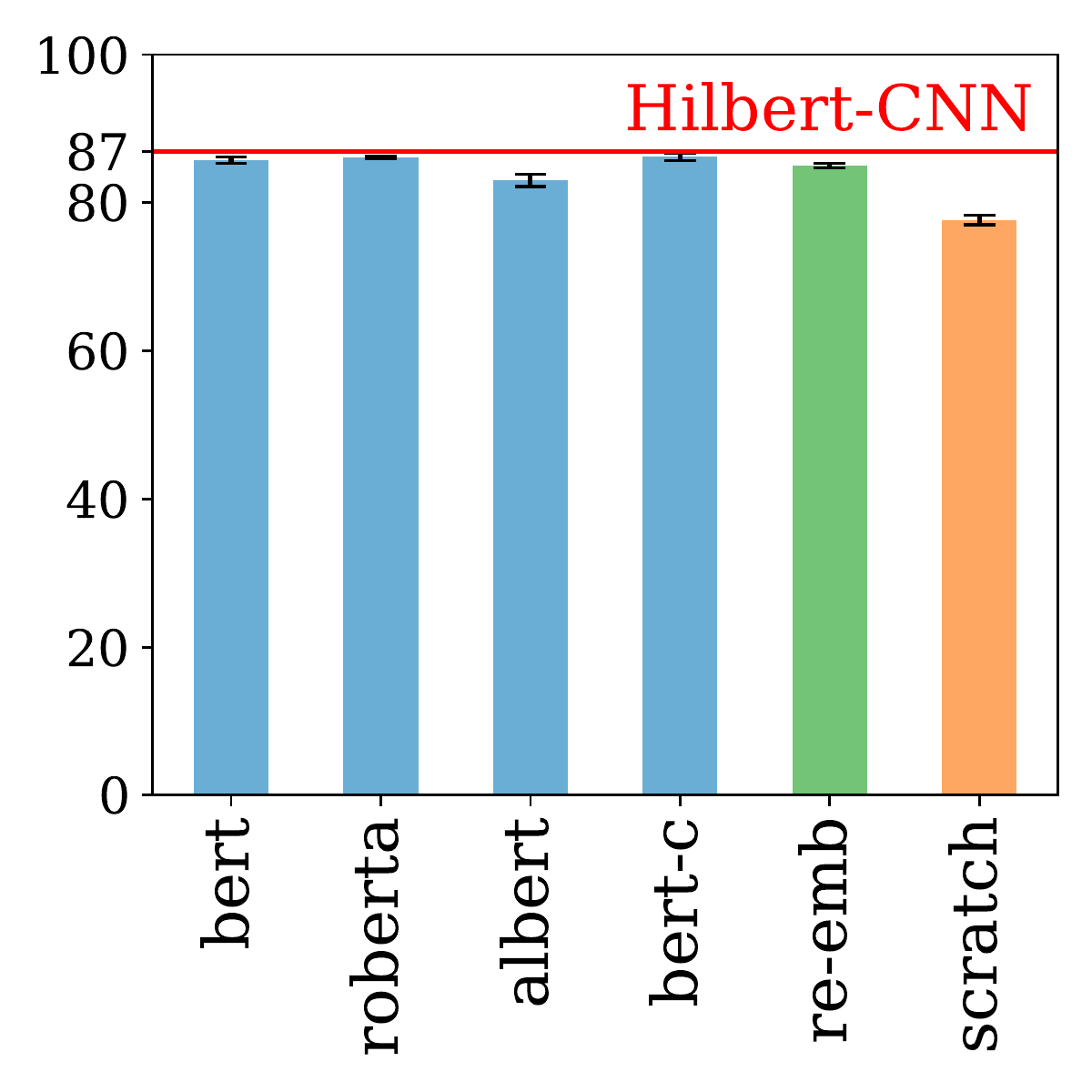}
			\caption{DNA (4 tasks)}
			\label{fig:dna}
		\end{subfigure}
		\begin{subfigure}[t]{0.49\linewidth}
			\centering
			\includegraphics[width=\linewidth]{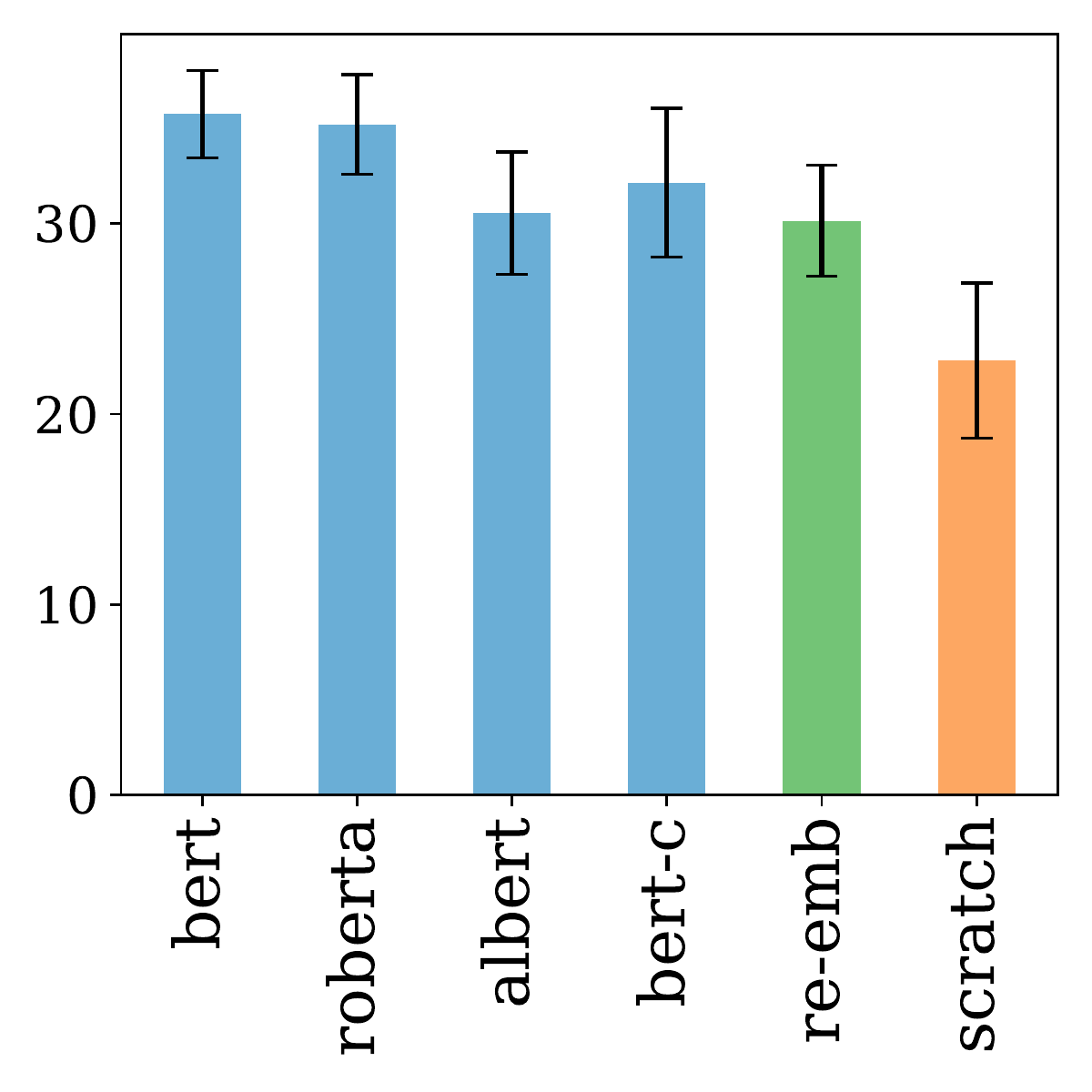}
			\caption{music (1 task)}
			\label{fig:music}
		\end{subfigure}
    \caption{The average scores (y-axis) of the pre-trained models and the ones trained from scratch within each discipline. The higher the scores, the better the models. The black error bars stand for standard deviation among random seeds. The red lines are the performance of the discipline-specific models. "-c": chinese.}
    \label{fig:main}
\end{figure}

\subsection{Results}
\label{exp:in}

Figure \ref{fig:main} shows the average scores of the pre-trained models (blue bars) and the models trained from scratch (orange bars) within each discipline. 
The means (bars) and standard deviations (black error bars) of figure \ref{fig:glue} are calculated over three random seeds, and the ones in figure \ref{fig:protein}, \ref{fig:dna}, and \ref{fig:music} are calculated over six independent runs (with different token mappings).
The GLUE score of BERT fine-tuned on normal GLUE acts as the \textit{discipline-specific} top-line (red  lines); 
For protein classification, the discipline-specific model is PLUS-TFM~\citep{proteinBERT}, which is a 12-layer transformer MLM pre-trained on protein sequence; 
For DNA classification, the discipline-specific model is Hilbert-CNN~\citep{hilbert-cnn}; 
For music composer classification, we use all the classes in the dataset to classify. 
But previous works~\citep{kim2020deep, spijker2020classifying} use only part of the classes, so no discipline-specific models are available. 
Detailed scores of each task within each discipline are left in the appendix.

The pre-trained models outperform the trained from scratch models in all disciplines.
The phenomenon is general over pre-trained models with different model structures (ALBERT), pre-training objectives, amount of pre-training data (RoBERTa), and different natural languages (BERT-Chinese).
Furthermore, the pre-trained models perform just slightly worse than PLUS-TFM and Hilbert-CNN without using any discipline-specific knowledge. The standard deviations of most models and disciplines are small, which implies that the effect of different token mappings is marginal.

At first sight, fine-tuning the pre-trained models on synthetic GLUE seems equivalent to randomly initializing the word embedding layer and then fine-tuning the pre-trained models on normal GLUE, which we called re-embedding (re-emb). 
If the equivalence is true, an explanation for the performance gain of the pre-trained models is just that the intermediate layers are already trained.  
Nevertheless, figure \ref{fig:glue} shows the equivalence does not hold. 
re-emb (green bar) degrades the performance. For the non-text data, the performance of re-emb is also worse than the models with all pre-trained parameters in figure \ref{fig:protein}, \ref{fig:dna}, and \ref{fig:music}.
Accordingly, the pre-trained word embedding layer benefits the non-text downstream tasks even though the meanings of the tokens are different from pre-training. 
We also find that using unused tokens of the pre-trained models even makes the performance degenerate to the trained from scratch baseline.

\section{Discussion}
The results in section \ref{exp:in} validate the potential of the pre-trained models as strong cross-disciplinary knowledge learners. The success of the pre-trained models could stem from better generalization ability or better training loss dynamics. In this section, we analyze the contribution of the pre-training in terms of optimization and generalization. In addition, we try to explain the success of the pre-trained models by comparing the representations from BERT and PLUS on the protein data.

\subsection{Training speed}

\begin{figure}[t]
		\centering
		\begin{subfigure}[t]{0.49\linewidth}
			\centering
			\includegraphics[width=\linewidth]{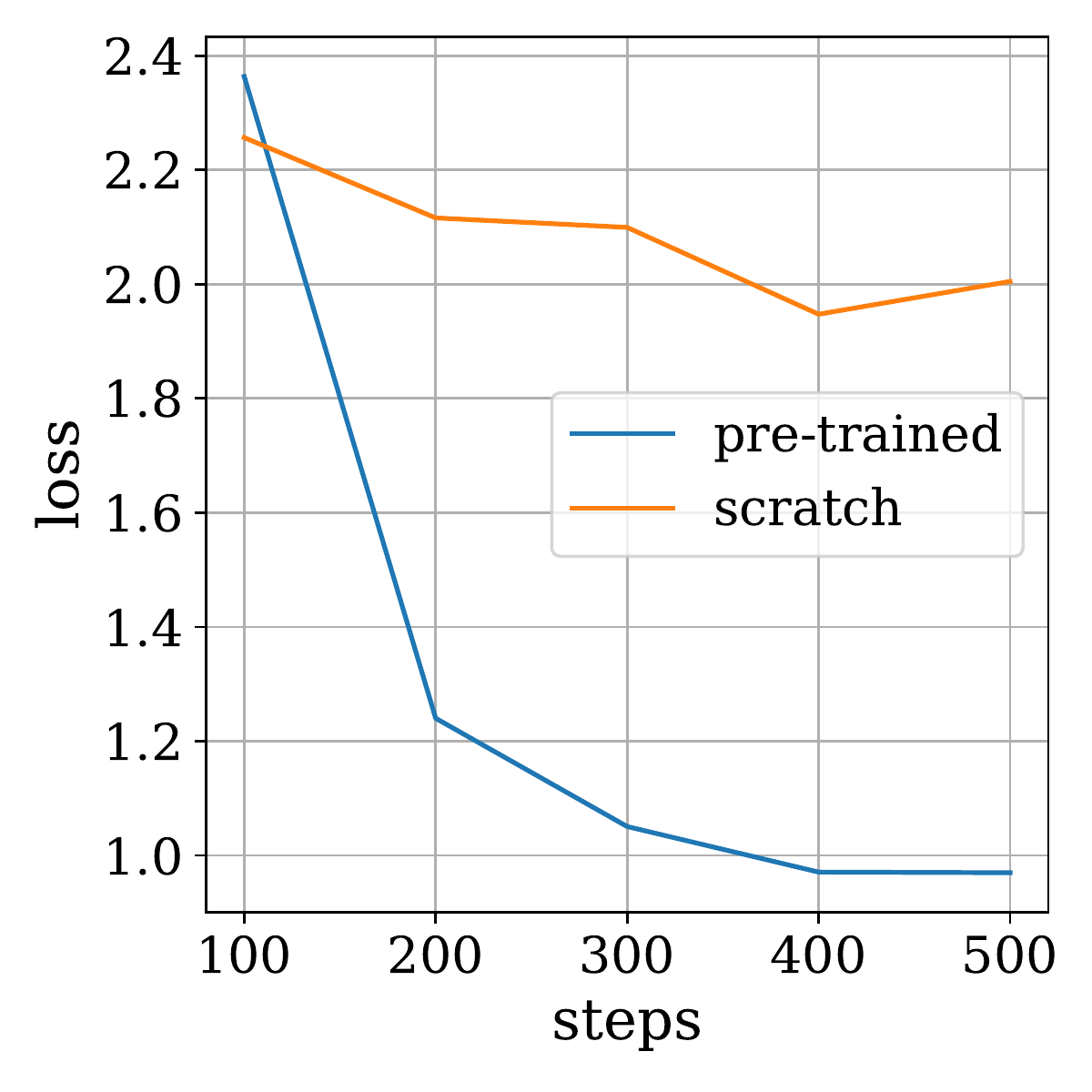}
			\caption{synthetic GLUE (STS-B)}
			\label{fig:sts-b_loss}
		\end{subfigure}
		\begin{subfigure}[t]{0.49\linewidth}
			\centering
			\includegraphics[width=\linewidth]{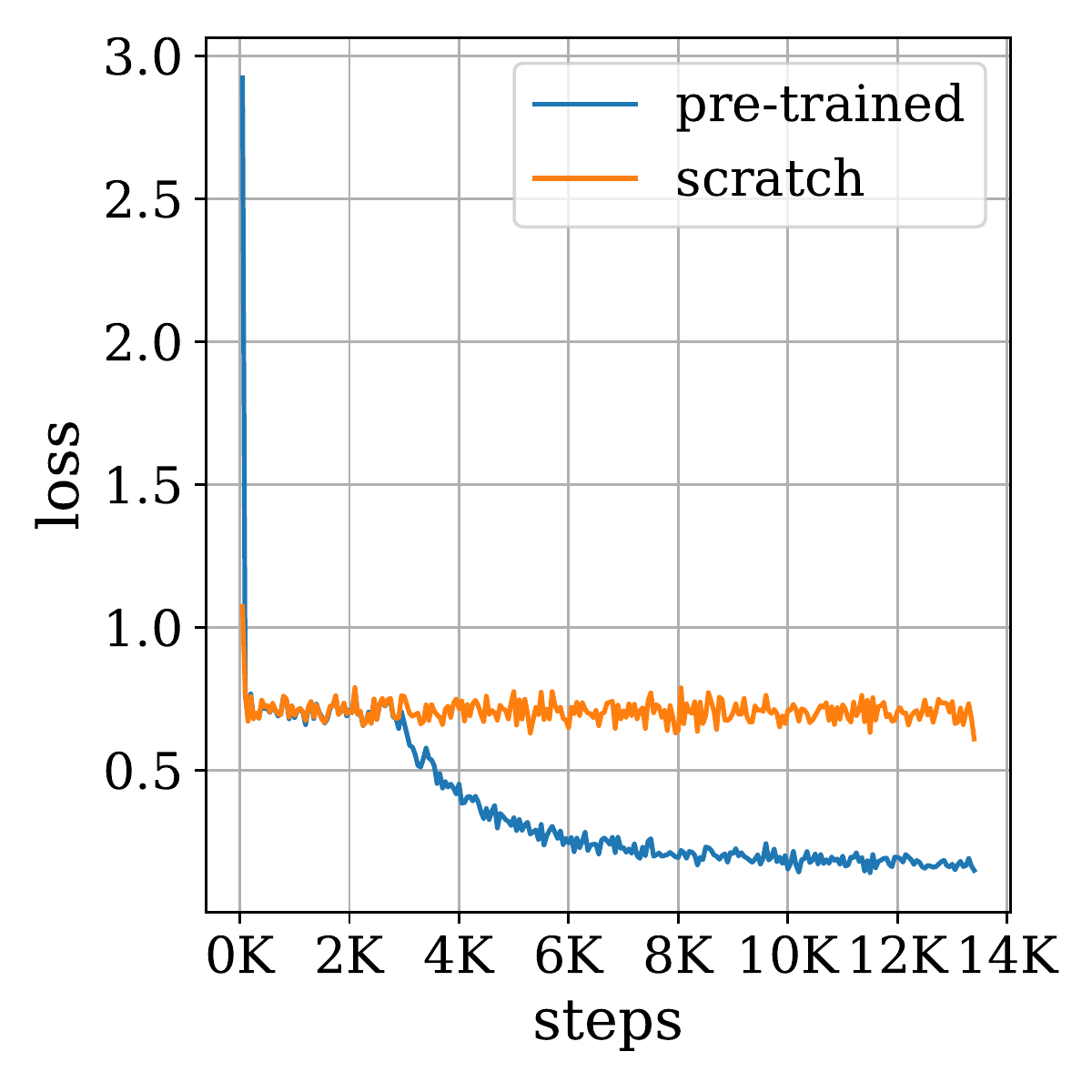}
			\caption{fluorescence}
			\label{fig:flu_loss}
		\end{subfigure}
		\caption{Training loss of BERT (blue lines) and the models trained from scratch (orange lines).}
		\label{fig:train_speed}
\end{figure}

Figure \ref{fig:train_speed} shows that BERT always reduces the training loss faster than the models trained from scratch. For \textit{fluorescence} task in figure \ref{fig:flu_loss}, the model trained from scratch seems stuck at local minimum rapidly, while BERT gets out of local minimum as fine-tuning proceeds. For a small dataset like \textit{STS-B} in figure \ref{fig:sts-b_loss}, BERT can reduce the training loss in only hundreds of steps, but the training loss of the model trained from scratch is still high. 
The results of the other tasks are similar and left in the appendix.

\subsection{Generalization ability}

\begin{figure}[t]
    \centering
    \begin{subfigure}[t]{0.49\linewidth}
		\centering
		\includegraphics[width=\linewidth]{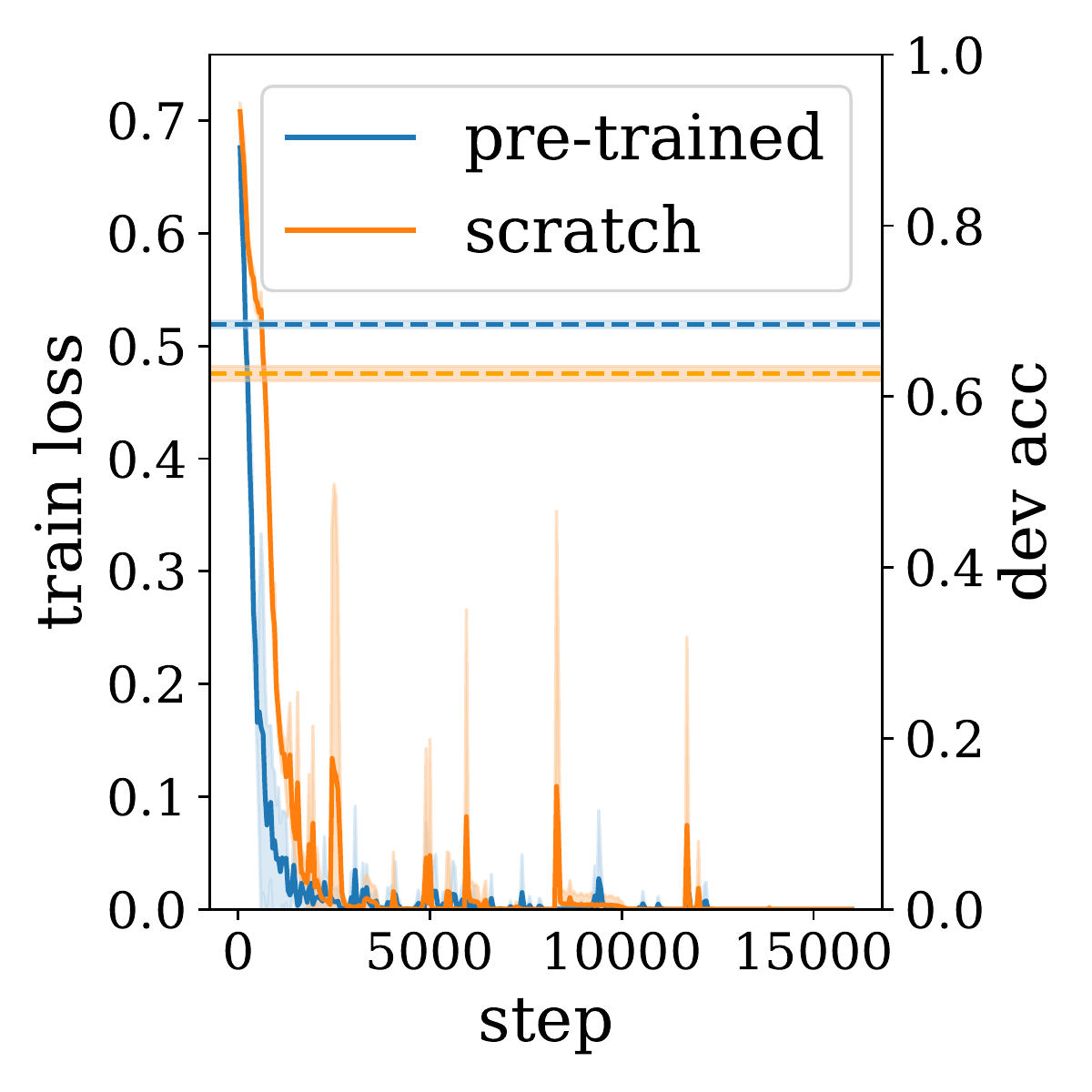}
		\caption{H3K9ac}
		\label{fig:small_1}
	\end{subfigure}
	\begin{subfigure}[t]{0.49\linewidth}
		\centering
		\includegraphics[width=\linewidth]{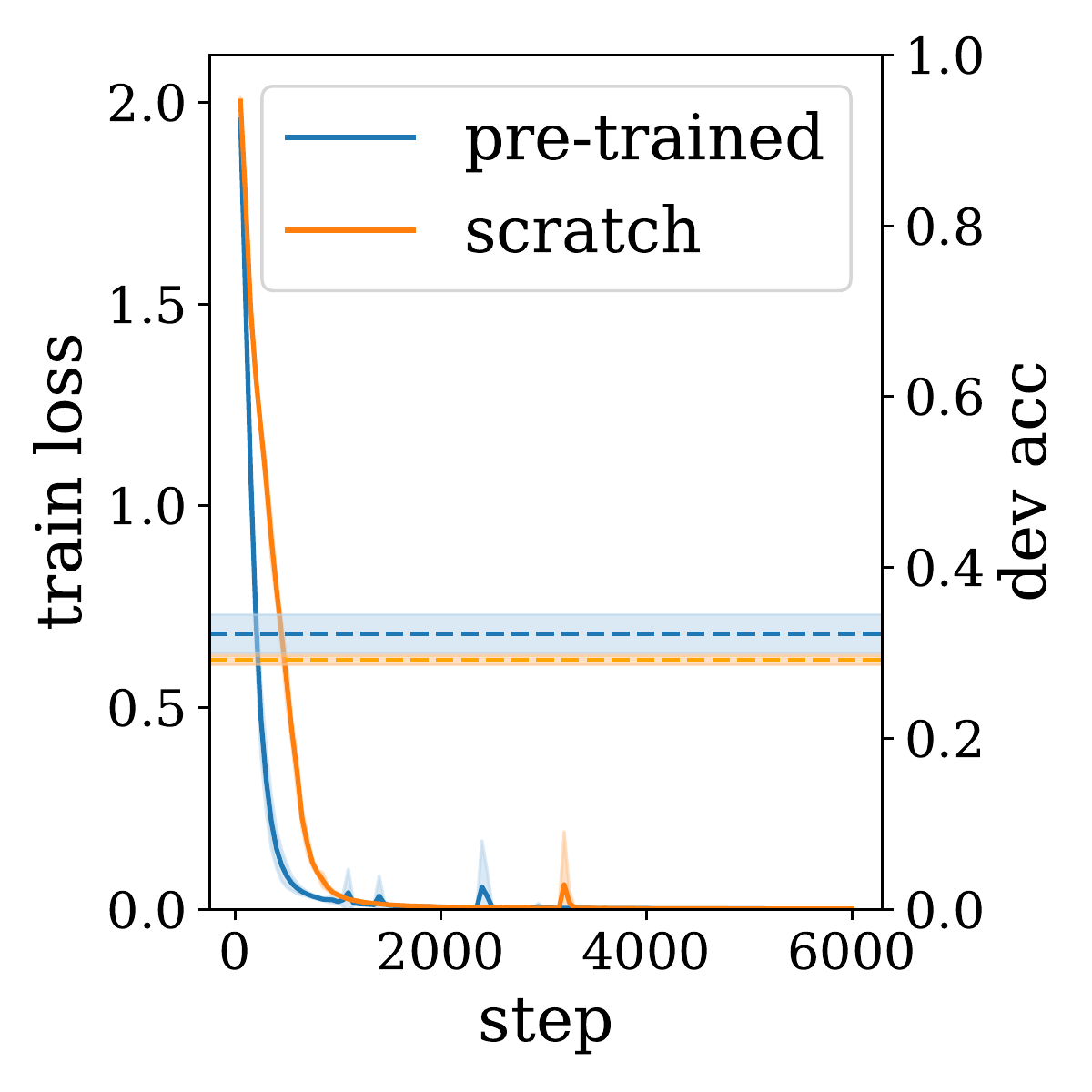}
		\caption{localization}
		\label{fig:small_2}
	\end{subfigure}
    \caption{The training curve (the solid lines) and the validation results at the end of training (the dashed lines) of pre-trained BERT (blue) and the model trained from scratch (orange). The solid lines stand for mean and the shaded areas stand for standard deviation over six independent runs.}
    \label{fig:small_data}
\end{figure}

Then we further examine the generalization ability of pre-trained models. We train all the models on only 1\% of the non-text data. In this way, both pre-training models and models training from scratch can converge to almost zero loss. And we compare their validation performance to know their generalization ability. 
Figure \ref{fig:small_data} show the results of one DNA dataset and one protein dataset. The results of the other tasks are similar and left in the appendix. Under the setting of 1 \% training data, the training losses of the pre-trained models and the models trained from scratch both converge to zero. And the pre-trained models still surpass the trained from scratch ones on the validation sets. Therefore, model pre-training improves the model generalization ability in \textit{discipline adaptation}.

\subsection{Representation similarity}

\begin{table}[t]
    \small
    \centering
    \begin{tabular}{lccc}
        \toprule
         & Flu. & Stab. & Loc. \\
        \midrule
        BERT - PLUS & \textbf{0.729} & \textbf{0.634} & \textbf{0.504}\\ 
        BERT - random & 0.598 & 0.545 & 0.362\\ 
        PLUS - random & 0.461 & 0.405 & 0.322\\ 
        random - random & 0.434 & 0.388 & 0.387\\
        \bottomrule
    \end{tabular}
    \caption{PWCCA similarity (a value in $[-1,1]$) between the representations of the last layer of the models on protein data. All the models are not fine-tuned. "random" means the randomly initialized models with the same architecture of BERT.}
    \label{tab:pwcca}
\end{table}

To explain the success of the text pre-trained models on the non-text data, we apply Projection Weighted Canonical Correlation Analysis (PWCCA)~\cite{pwcca} on the representations of BERT and PLUS-TFM. 
The results in table \ref{tab:pwcca} show that before fine-tuning, the similarity between BERT and PLUS is much higher than the similarity between BERT and the randomly initialized model. 
The behavior of BERT is different from the randomly initialized models when processing the non-text data, even though BERT is pre-trained only on natural language, and this could be one of the reasons behind BERT's \textit{discipline adaptability}.

\subsection{Hypotheses}

To elaborate on the reason behind the \textit{discipline adaptability} of the pre-trained MLM, we have tried to explore several possibilities. However, they are not sufficient to explain the phenomenon. In the next sub-sections, we summary these experiments. Some detailed results are left in the appendix. 

\subsubsection{Attention similarity}
\begin{figure}[t]
    \centering
    \begin{subfigure}[t]{0.49\linewidth}
        \centering
        \includegraphics[width=\linewidth]{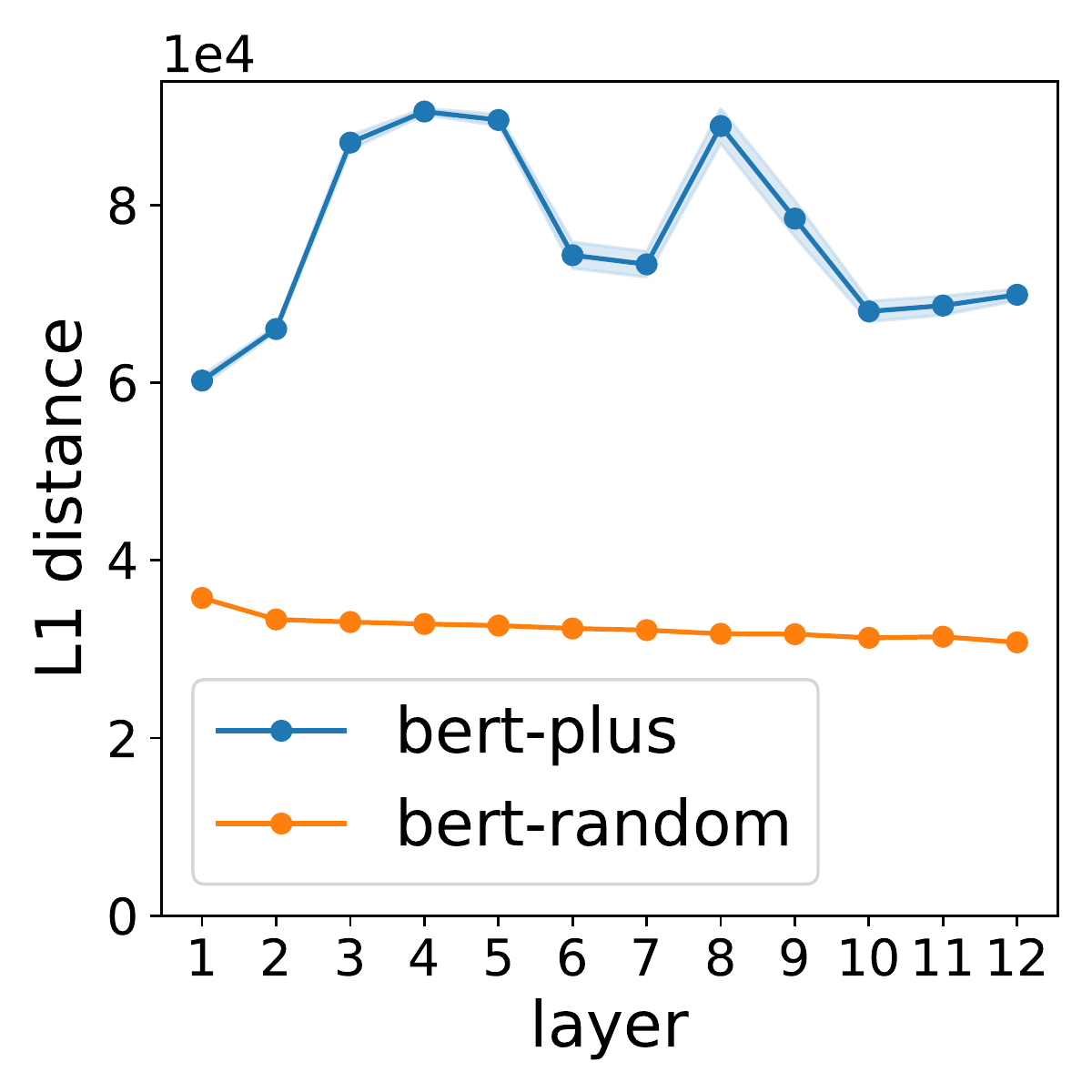}
        \caption{w/o fine-tuning}
    \end{subfigure}
    \begin{subfigure}[t]{0.49\linewidth}
        \centering
        \includegraphics[width=\linewidth]{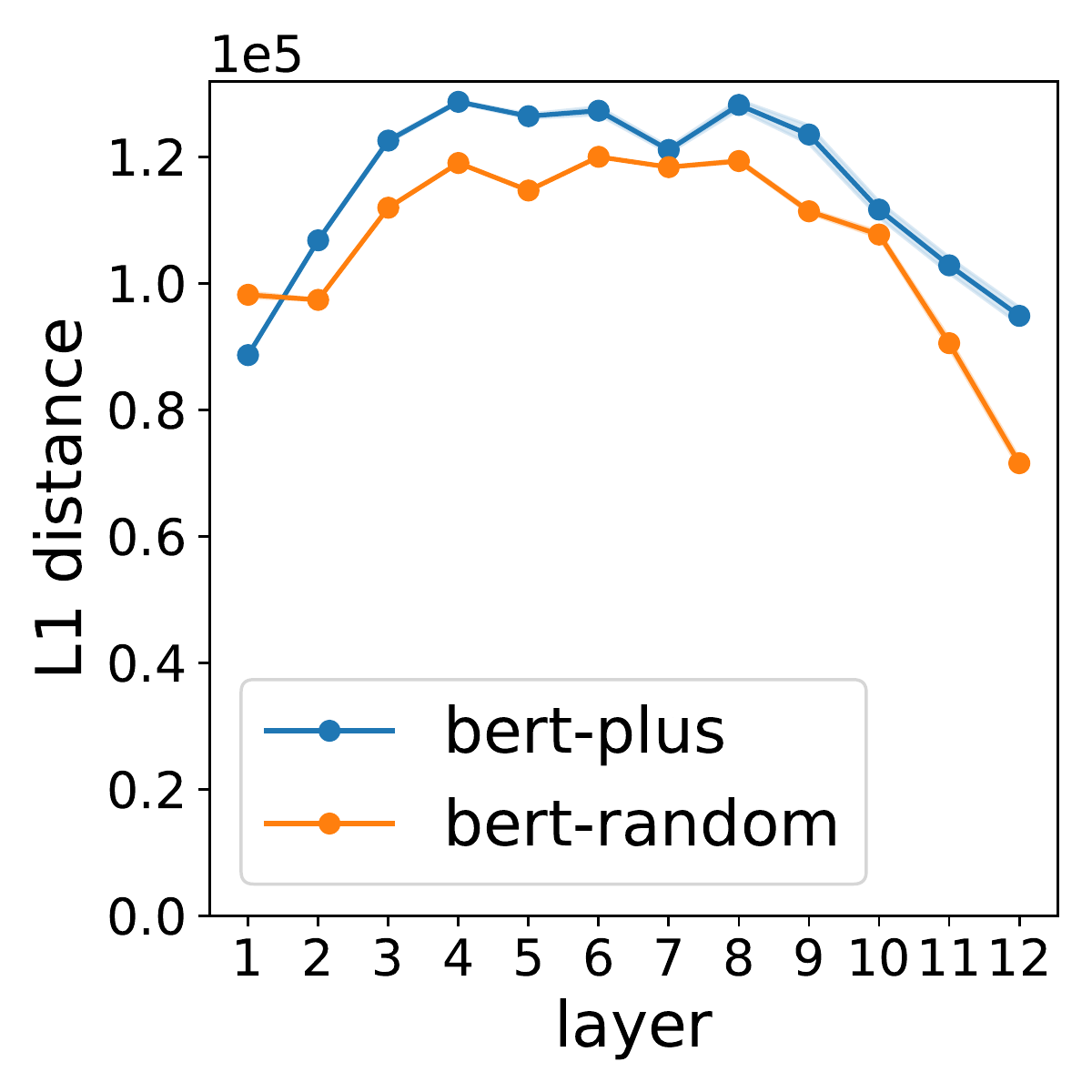}
        \caption{fine-tuning on localization}
    \end{subfigure}
    \caption{Average $L_1$ distance of the attention maps from different models after applying matching algorithm.}
    \label{fig:smatching}
\end{figure}
We have examined the similarity of the attention map between BERT, PLUS, and the randomly initialized model. For one input data, we first extract their attention maps in each layer. For the attention maps in the same layer, we use the Hungarian algorithm to find the minimum L1 distance matching between the maps from the different models. The average distance of the matching represents the similarity of attention patterns in each layer. The results on the protein data are shown in fig \ref{fig:smatching}. In almost all layers, the distance between BERT and PLUS is larger than the one between BERT and random, no matter whether they are fine-tuned or not. From this viewpoint, we may not consider that PLUS and BERT share common attention patterns.

\subsubsection{Properties of the pre-training data}

We would like to investigate which properties of the pre-training data result in \textit{discipline adaptability}. We used the following data used in \citep{chiang2020pre} to pre-train MLMs:
\begin{itemize}
    \item uniform: Tokens in a sentence are sampled \textit{i.i.d} from uniform distribution over all tokens.
    
    \item flat or nesting parentheses: Tokens in a sentence are generated randomly, recursively, while hierarchically matched.
    
    \item Kannada~\citep{ortiz-suarez-etal-2020-monolingual}: Kannada is a language spoken by people in southwestern India.
\end{itemize} 

\begin{table}[t]
    \small
    \centering
    \begin{tabular}{lccc|c}
        \toprule
         & Flu. & Stab. & Loc. & Avg.\\
        \midrule
        scratch & 29.4 & 59.6 & 56.6 & 48.5\\
        uniform & 36.6 & 53.7 & 57.7 & 49.3\\ 
        flat & 36.8 & 56.3 & 57.8 & 50.3\\ 
        nesting & 47.9 & 62.7 & 60.5 & 57.0\\ 
        Kannada & 47 & 71.3 & 62.8 & \textbf{60.4}\\
        \bottomrule
    \end{tabular}
    \caption{Protein classification results of the models pre-trained on the artifical datasets and human language.}
    \label{tab:artificial}
\end{table}

However, as shown in table \ref{tab:artificial}, the models pre-trained on the artificial data perform worse on the protein classification than the one pre-trained on natural languages. So natural language may indeed share similarities with protein, while the hierarchical structure only may not be enough to explain the \textit{discipline adaptability}.

\subsubsection{Gradient stability}

We also examine whether BERT satisfies the following criteria about training stability or not:
\begin{itemize}
    \item \citet{saxe2014exact} claim that if the singular values of the output-input Jacobian matrix of the model initialization are all equal to 1 (called dynamical isometry), then the model can avoid gradient vanish or gradient explode and be trained better.
    
    \item \citet{sankararaman2020impact} show that negatively correlated gradients produced by different data would slow down the convergence.
    
    \item \citet{liu-etal-2020-understanding} observe that a large variance of the output of transformer under parameter perturbation would make the training procedure unstable.
\end{itemize}
On the synthetic GLUE, BERT does not fit these criteria better, or even worse than the gaussian initialization. Although BERT is optimized better even on the non-text data, the above theories fail to elaborate the optimization properties of BERT. The detailed results are left in the appendix.

\section{Conclusion}

In this paper, we investigate the potential of BERT as a cross-disciplinary knowledge learner. By fine-tuning BERT on the synthetic text data with meanings of tokens changed and the non-text data, we verify that BERT can be adapted to data of different disciplines efficiently and generalizes well. Besides, we discover the non-trivial similarity between the models pre-trained on text and protein before fine-tuning by PWCCA, which helps to explain the reasons behind BERT's \textit{discipline adaptability}. We hope that the proposed settings can act as new analysis tools for researchers and provide new insight into the power of pre-trained models.

\section*{Broader impact}

The results of this paper are helpful for practitioners of other disciplines when large-scale pre-training datasets are unavailable. The \textit{discipline adaptability} of the pre-trained models also helps to reduce computational costs since we may not need to pre-train one model for each discipline. We think that the results in this paper will not cause any ethical issues.

\section*{Acknowledgements}

We thank National Center for High-performance Computing (NCHC) of National Applied Research
Laboratories (NARLabs) in Taiwan for providing computational and storage resources.

\bibliography{anthology,ref}
\bibliographystyle{acl_natbib}

\appendix

\section{Hyperparameters for experiments}
\label{app:hyper}

The transformer models used in the experiments are 12-layer, 768-hidden, 12-attention heads models if not specified. The total number of parameters is 110M, which is the same size as BERT-base. For BERT-large-uncased (bert-l) and the large model trained from scratch (scratch-l) in the appendix, the total number of parameters is 340M. PLUS-TFM has the same structure as BERT-base and the total number of parameters is 110M. For the Hilbert-CNN model, the total number of parameters is 961K according to the original paper.

We use Adam optimizer for all experiments in the paper, and the learning rate is set to $10^{-5}$. The optimizer is chosen by applying grid search on MRPC from GLUE dataset, and the learning rate is chosen by applying grid search on MRPC and the validation set of \textit{fluorescence} protein classification task. We search learning rate from $10^{-4}$ to $10^{-7}$. We uniformly sample 5 points in this range, and further sample 5 points between $10^{-5}$ and $10^{-6}$. We search optimizers including Adagrad, Adam, Adamax, RAdam, and NovoGrad for three independent runs. The parameters that make the randomly initialized 12-layer transformer models achieve the highest F1 score on the MRPC training set and the highest Spearman correlation on the \textit{fluorescence} validation set are chosen (which are also the best for the re-emb setting). We do not use gradient clipping and warm-up, so the learning rate schedule is the same as linear learning rate decay. All models are trained with batch size 32 on two RTX 2080-Ti (GLUE dataset) or one Tesla V100 GPU (protein classification, DNA classification, and music classification). For GLUE dataset, we use the validation set of GLUE as testing set and evaluate the final checkpoint. For all the non-text datasets, we select the best checkpoints on the validation set and evaluate on the testing set. 

\section{Full results on synthetic GLUE}
\begin{table*}[t]

	\centering
	\begin{tabular}{llllllllll}
		\toprule
		  &  MNLI & QQP & QNLI & SST-2 & CoLA & STS-B & MRPC & RTE & avg \\
		 & m/mm-acc & F1 & acc & acc & mcc & spr & F1 & acc \\
		\midrule
		Normal data\\
		\midrule
		BERT & 84.0/84.3 & 87.4 & 91.2 & 92.2 & 55.0 & 86.7 & 85.5 & 62.1 & 80.9 \\
		re-emb & 66.1/66.8 & 78.7 & 64.5 & 80.0 & 0.0 & 19.0 & 78.6 & 49.0 & 55.9\\
		\midrule
		\textit{permutation}\\
		\midrule
		bert & 68.6/69.9 & 81.2 & 80.0 & 79.6 & 0.0 & 77.8 & 82.9 & 60.2 & 66.6 \\
		roberta & 66.4/67.5 & 77.8 & 79.2 & 76.2 & 0.0 & 73.9 & 83.0 & 56.0 & 64.4 \\
	    albert & 65.9/67.5 & 79.5 & 67.5 & 71.3 & 0.0 & 71.6 & 81.4 & 53.2 & 63.2 \\
	    bert-c & 68.9/70.0 & 81.7 & 80.2 & 77.5 & 0.0 & 76.1 & 85.3 & 58.5 & 66.4 \\
		scratch & 61.4/62.1 & 69.3 & 61.1 & 81.0 & 0.0 & 8.3 & 81.3 & 54.2 & 53.2 \\
		\midrule
		bert-l & 44.5/44.7 & 26.8 & 60.4 & 60.7 & 0.0 & 73.5 & 82.3 & 54.3 & 49.7 \\
		scratch-l & 40.6/40.9 & 21.7 & 50.2 & 80.9 & 0.0 & 9.2 & 81.2 & 50.5 & 41.7 \\
		\midrule
		
		\textit{random mapping}\\
		\midrule
		BERT & 68.2/68.6 & 80.6 & 79.7 & 78.7 & 0.0 & 75.6 & 83.4 & 58.5 & 65.9\\ 
		scratch & 61.5/62.0 & 69.0 & 61.5 & 79.6 & 0.0 & 8.3 & 81.3 & 51.0 & 52.7\\

		\bottomrule
	\end{tabular}
	\caption{Full results on GLUE validation set (averaged over three random seeds). The evaluation metrics are listed below the task names. Normal data means the models are fine-tuned on the normal GLUE. Permutation means the models are fine-tuned on the synthetic GLUE. Random mapping means the token mapping is generated randomly. "avg": The average score (GLUE score). "m/mm": MNLI matched/mismatched set. "spr": Spearman correlation. "mcc": Matthews correlation coefficients. "re-emb": Randomly initializing the word embedding layer of BERT and fine-tuning the BERT.}
	\label{app:glue}
\end{table*}

\begin{table*}[t]

	\centering
	\begin{tabular}{llllllllll}
		\toprule
		  &  MNLI & QQP & QNLI & SST-2 & CoLA & STS-B & MRPC & RTE  \\
		 & acc & F1 & acc & acc & mcc & spr & F1 & acc \\
		\midrule
		permutation\\
		\midrule
		BERT-l & 21.2/21.3 & 46.5 & 17.0 & 16.9 & 0.0 & 3.8 & 0.9 & 6.7  \\
		scratch-l & 14.4/14.7 & 37.6 & 0.6 & 0.7 & 0.0 & 0.4 & 0.0 & 3.1  \\
		\bottomrule
	\end{tabular}
	\caption{The standard deviations of the large models on GLUE validation set.}
	\label{app:glue_std}
\end{table*}

The full results on synthetic GLUE dataset are shown in table \ref{app:glue}. The pre-trained models (including the large model) outperform the models trained from scratch except for SST-2 and CoLA. For SST-2, pre-trained models generalize worse than the models trained from scratch. For CoLA, all models fail to be trained. But for the other six tasks, pre-trained models outperform the models trained from scratch. The standard deviations of most of the models are small than 2 except for the RTE dataset and the large models. For RTE, the maximum standard deviation is 5.24 (ALBERT). For the large models, the standard deviations are much larger and listed in table \ref{app:glue_std}. When we generate the token mappings randomly, the results are similar. This indicates that the effect of different token mappings is marginal. 

For BERT with word embedding layer randomly initialized and then fine-tuned (re-emb), the performance is worse than the one using the whole pre-trained weights, which indicates that even pre-trained word embedding is necessary.

\section{Testing and validation performance on non-text data}

\begin{table*}[t]

	\centering
	\begin{tabular}{cccccccc}
		\toprule
		  & \multicolumn{3}{c}{Protein}  & \multicolumn{4}{c}{DNA}\\
		  \cmidrule(lr){2-4} \cmidrule(lr){5-8} 
		  & localization & stability & fluorescence & H3 & H4 & H3K9ac & Splice \\
		\midrule
		specific & 69.0 & 76.0 & 63.0 & 87.3 & 87.3 & 79.1 & 94.1 \\
		\midrule
		bert & 64.1 {\small (0.4)} & 70.6 {\small(3.9)} & 58.4 {\small(3.2)} & 83.6 {\small(1.3)} & 85.9 {\small(0.9)} & 77.2 {\small(0.9)} & 96.4 {\small(0.8)} \\
		roberta & 65.4 {\small(1.0)} & 73.6 {\small(1.7)} & 59.8 {\small(1.6)} & 84.2 {\small(0.7)} & 86.5 {\small(0.4)} & 78.9 {\small(0.6)} & 94.7 {\small(0.3)} \\
		albert & 65.1 {\small(0.6)} & 70.3 {\small(2.8)} & 55.0 {\small(2.4)} & 83.5 {\small(0.3)} & 86.4 {\small(0.5)} & 78.2 {\small(0.6)} & 84.0 {\small(3.6)} \\
		bert-c & 63.1 {\small(0.8)} & 69.7 {\small(3.0)} & 53.4 {\small(8.4)} & 84.0 {\small(1.3)} & 86.2 {\small(0.9)} & 77.9 {\small(0.4)} & 96.8 {\small(0.8)} \\
		re-emb & 62.8 {\small(0.5)} & 71.0 {\small(2.3)} & 37.4 {\small(4.4)} & 83.0 {\small(1.2)} & 83.9 {\small(0.8)} & 77.5 {\small(0.4)} & 95.6 {\small(0.4)} \\
		scratch & 57.8 {\small(0.6)} & 62.2 {\small(5.3)} & 28.4 {\small(1.6)} & 76.1 {\small(0.7)} & 66.6 {\small(1.3)} & 72.6 {\small(0.3)} & 95.3 {\small(1.2)} \\
		\midrule
		bert-l & 63.0 {\small(0.4)} & 23.2 {\small(41.8)} & 44.6 {\small(18.6)} & 70.5 {\small(14.8)} & 63.5 {\small(7.3)} & 65.0 {\small(9.4)} & 82.6 {\small(18.7)} \\
		scratch-l & 58.3 {\small(0.5)} & 59.7 {\small(4.3)} & 11.6 {\small(3.5)} & 76.7 {\small(0.5)} & 58.7 {\small(2.5)} & 67.1 {\small(7.9)} & 95.6 {\small(0.8)} \\
		\bottomrule
	\end{tabular}
	\caption{Testing results of protein classification and DNA classification. The metric is Spearman correlation for \textit{fluorescence} and \textit{stability}. And the metric is accuracy for all the other tasks. The number in the parenthesis is the standard deviation (calculated over six independent runs with different token mappings). "specific": the \textit{discipline-specific} models.}
	\label{tab:all_task}
\end{table*}

\begin{table}[t]

	\centering
	\begin{tabular}{cccc}
		\toprule
		  & Protein & DNA & Music\\
		\midrule
		bert  & 64.4 {\small(1.2)} & 85.8 {\small(0.4)} & 35.7 {\small(2.3)}\\
		roberta & 66.3 {\small(0.8)} & 86.1 {\small(0.2)} &  35.2 {\small(2.6)}\\
		albert & 63.5 {\small(1.2)} & 83.0 {\small(0.9)} &  30.5 {\small(3.2)}\\
		bert-c & 62.1 {\small(2.9)} & 86.2 {\small(0.5)} &  32.1 {\small(3.9)}\\
		re-emb & 57.1 {\small(1.3)} & 85.0 {\small(0.3)} &  30.1 {\small(2.9)}\\
		scratch & 49.5 {\small(2.2)} & 77.7 {\small(0.7)} &  22.8 {\small(4.1)}\\
		\midrule
		bert-l & 43.6 {\small(14.7)} & 70.4 {\small(6.7)} &  30.8 {\small(4.0)}\\
		scratch-l & 43.2 {\small(2.6)} & 74.5 {\small(1.8)} & 26.0 {\small(5.0)} \\
		\bottomrule
	\end{tabular}
	\caption{The testing results of music composer classification, the average score of DNA classification, and the average score of protein classification. The numbers in the parenthesis are the standard deviations calculated over six independent runs with different token mappings.}
	\label{tab:all_task_mean}
\end{table}

\begin{table*}[t]

	\centering
	\begin{tabular}{ccccccccc}
		\toprule
		  & \multicolumn{3}{c}{Protein}  & \multicolumn{4}{c}{DNA}\\
		  \cmidrule(lr){2-4} \cmidrule(lr){5-8} 
		  & localization & stability & fluorescence & H3 & H4 & H3K9ac & Splice \\
		\midrule
		bert & 69.6 {\small(0.9)} & 70.6 {\small(0.8)} & 57.0 {\small(4.3)} & 83.5 {\small(0.7)} & 87.0 {\small(0.6)} & 78.4 {\small(0.7)} & 95.9 {\small(0.7)} \\
		roberta & 71.1 {\small(1.1)} & 68.8 {\small(0.9)} & 59.2 {\small(2.7)} & 86.7 {\small(0.6)} & 87.6 {\small(0.2)} & 79.5 {\small(0.6)} & 95.2 {\small(0.5)} \\
		albert & 69.0 {\small(0.8)} & 66.3 {\small(1.9)} & 50.9 {\small(5.3)} & 85.6 {\small(0.9)} & 87.0 {\small(0.3)} & 79.3 {\small(0.6)} & 82.5 {\small(3.1)} \\
		bert-c & 69.0 {\small(0.7)} & 75.4 {\small(0.5)} & 50.8 {\small(9.3)} & 83.6 {\small(1.3)} & 87.5 {\small(0.7)} & 79.2 {\small(0.8)} & 96.5 {\small(0.5)} \\
		re-emb & 67.9 {\small(0.4)} & 67.3 {\small(1.7)} & 36.9 {\small(3.2)} & 82.8 {\small(0.5)} & 85.3 {\small(0.7)} & 78.4 {\small(0.8)} & 95.4 {\small(0.9)} \\
		scratch & 59.9 {\small(0.8)} & 63.6 {\small(0.6)} & 29.4 {\small(1.3)} & 75.4 {\small(0.2)} & 66.5 {\small(0.8)} & 71.9 {\small(0.2)} & 96.0 {\small(0.3)} \\
		\midrule
		bert-l & 69.3 {\small(1.0)} & 37.5 {\small(27.5)} & 42.5 {\small(18.4)} & 70.1 {\small(14.6)} & 64.0 {\small(8.1)} & 64.1 {\small(9.4)} & 82.2 {\small(19.6)}\\
		scratch-l & 60.9 {\small(0.7)} & 61.6 {\small(2.1)} & 13.0 {\small(2.9)} & 75.7 {\small(0.3)} & 58.6 {\small(2.1)} & 66.5 {\small(7.9)} & 96.0 {\small(0.6)} \\ 
		\bottomrule
	\end{tabular}
	\caption{Validation results of protein classification and DNA classification. The metric is Spearman correlation for \textit{fluorescence} and \textit{stability}. And the metric is accuracy for all the other tasks. The numbers in the parenthesis are the standard deviations calculated over six independent runs with different token mappings.}
	\label{tab:valid}
\end{table*}

\begin{table}[t]
	
	\centering
	\begin{tabular}{cccc}
		\toprule
		  & Protein & DNA & Music\\
		\midrule
		bert & 65.7 {\small(1.6)} & 86.2 {\small(0.4)} &  43.2 {\small(4.2)}\\
		roberta & 66.4 {\small(0.8)}  & 87.2 {\small(0.2)} &  41.1 {\small(3.6)}\\
		albert & 62.1 {\small(1.9)} & 83.6 {\small(0.8)} &  36.3 {\small(3.3)}\\
		bert-c & 65.1 {\small(3.3)} & 86.7 {\small(0.6)} &  42.2 {\small(3.4)}\\
		re-emb & 57.4 {\small(1.4)} & 85.5 {\small(0.4)} &  39.5 {\small(2.8)}\\
		scratch & 51.0 {\small(0.6)} & 77.5 {\small(0.2)} &  31.0 {\small(2.3)}\\
		\midrule
		bert-l & 49.8 {\small(9.0)} & 70.1 {\small(7.1)} &  43.3 {\small(3.5)}\\
		scratch-l & 45.2 {\small(0.9)} & 74.2 {\small(1.7)} & 34.3 {\small(2.6)} \\
		\bottomrule
	\end{tabular}
	\caption{Validation results of music composer classification, average score of DNA classification, and average score of protein classification. The numbers in the parenthesis are the standard deviations calculated over six independent runs with different token mappings.}
	\label{tab:valid_mean}
\end{table}

Table \ref{tab:all_task} and \ref{tab:valid} show the full testing and validation results on each non-text classification task. Table \ref{tab:all_task_mean} and \ref{tab:valid_mean} show the average scores of each discipline. For most of the tasks, the text pre-trained models outperform the models trained from scratch and the re-emb models on both the testing and validation set.

\section{Training loss for the other tasks}

\begin{figure*}[t]
    \centering
    \begin{subfigure}[t]{0.245\linewidth}
			\centering
			\includegraphics[width=\linewidth]{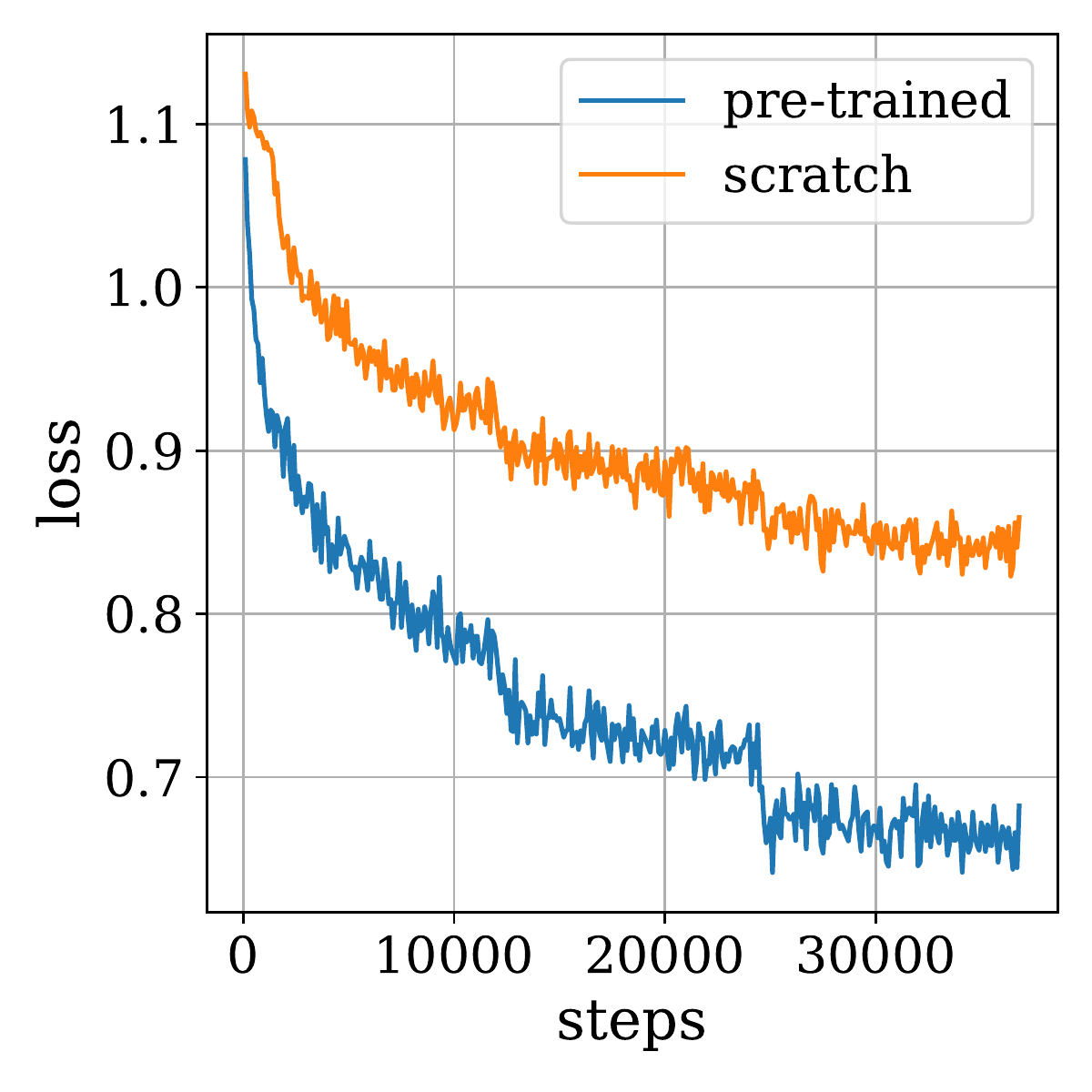}
			\caption{MNLI}
	\end{subfigure}
	\begin{subfigure}[t]{0.245\linewidth}
			\centering
			\includegraphics[width=\linewidth]{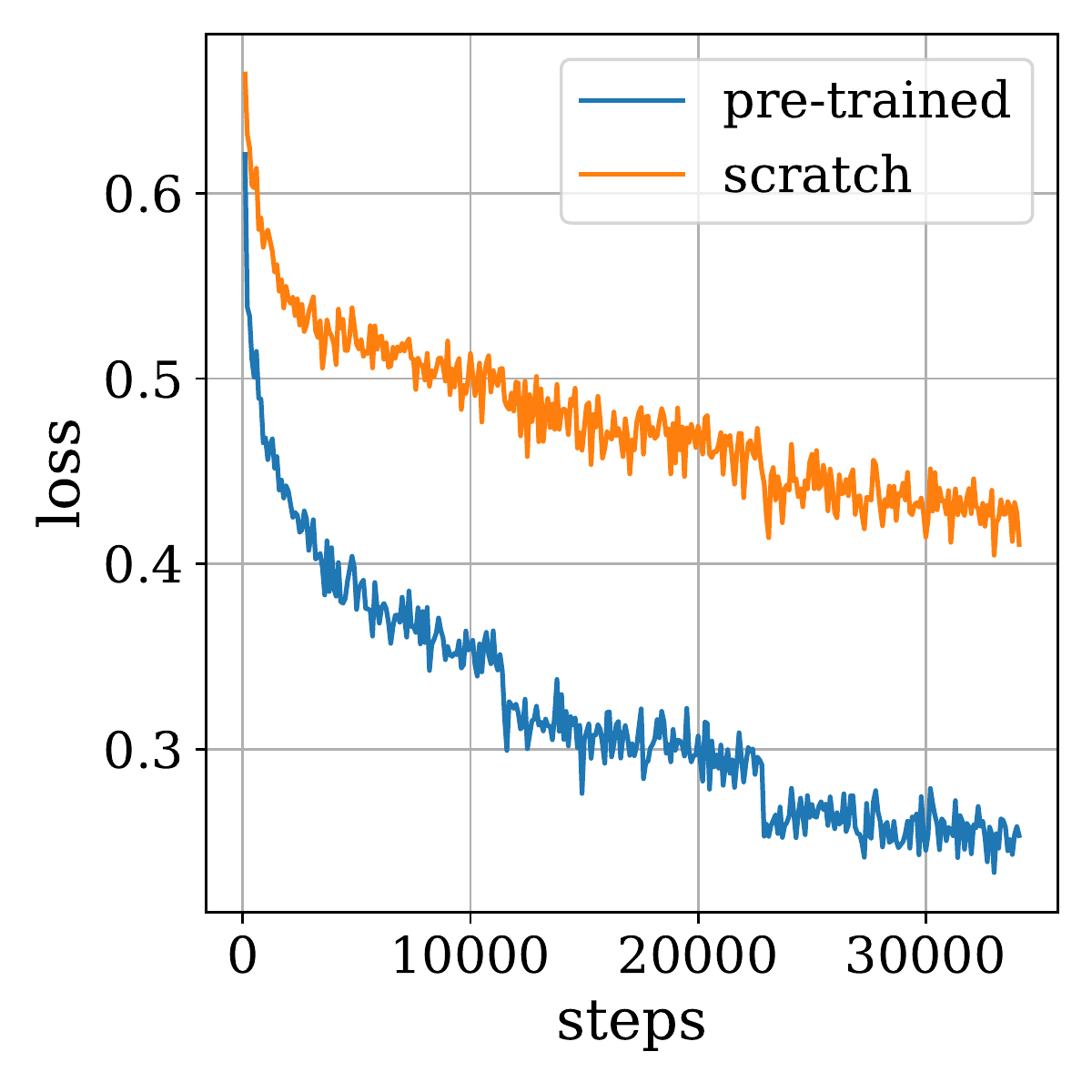}
			\caption{QQP}
	\end{subfigure}
	\begin{subfigure}[t]{0.245\linewidth}
			\centering
			\includegraphics[width=\linewidth]{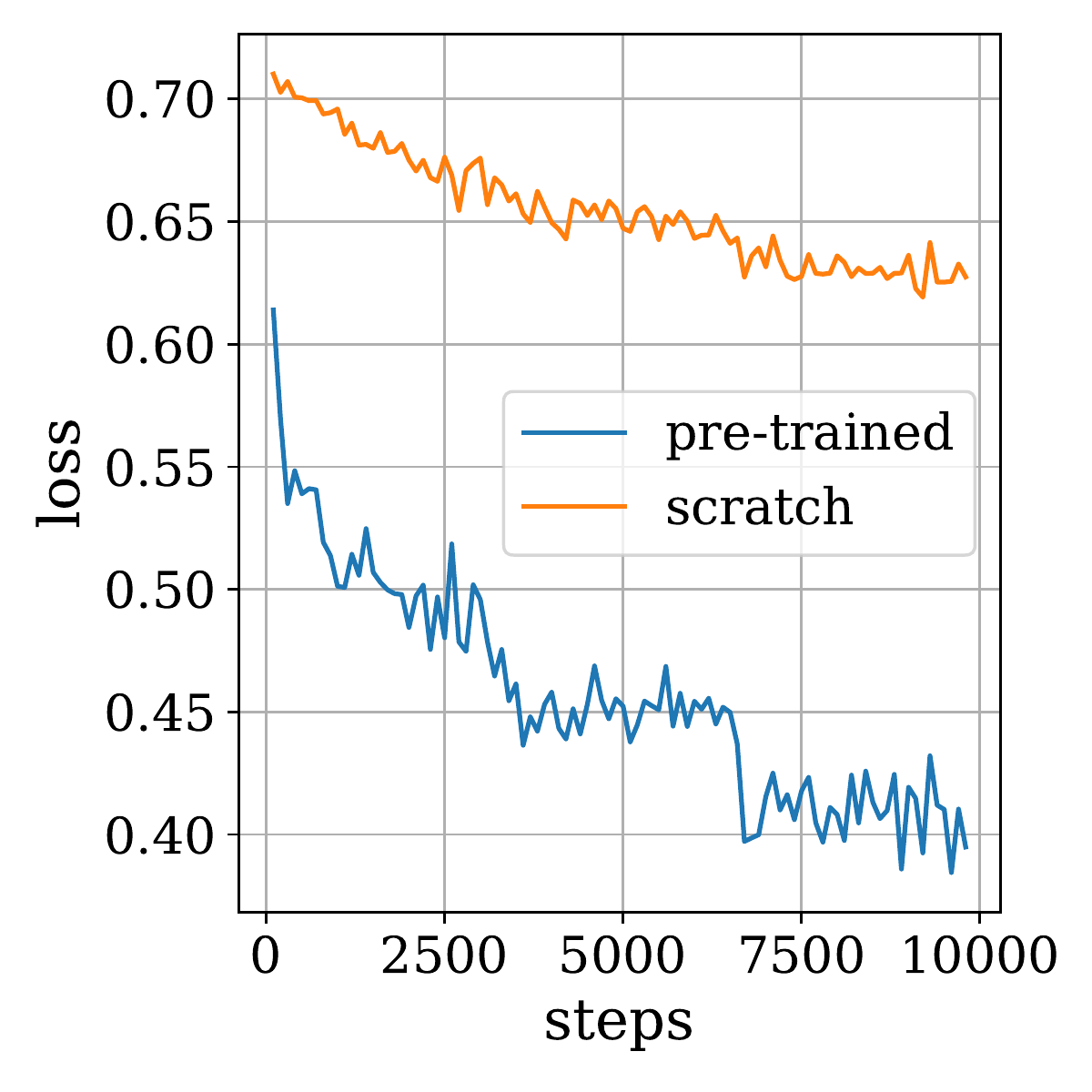}
			\caption{QNLI}
	\end{subfigure}
	\begin{subfigure}[t]{0.245\linewidth}
			\centering
			\includegraphics[width=\linewidth]{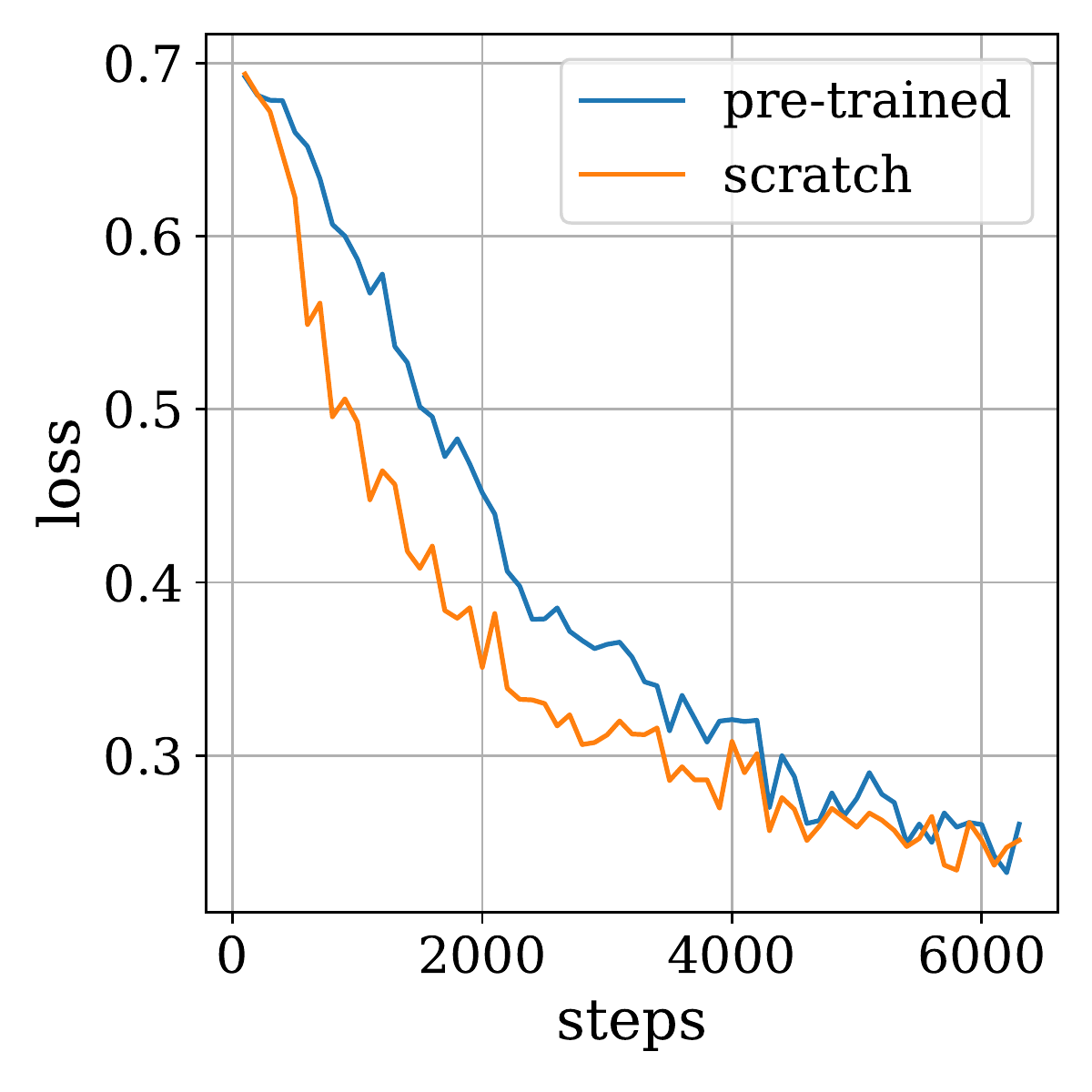}
			\caption{SST-2}
	\end{subfigure}
	\begin{subfigure}[t]{0.245\linewidth}
			\centering
			\includegraphics[width=\linewidth]{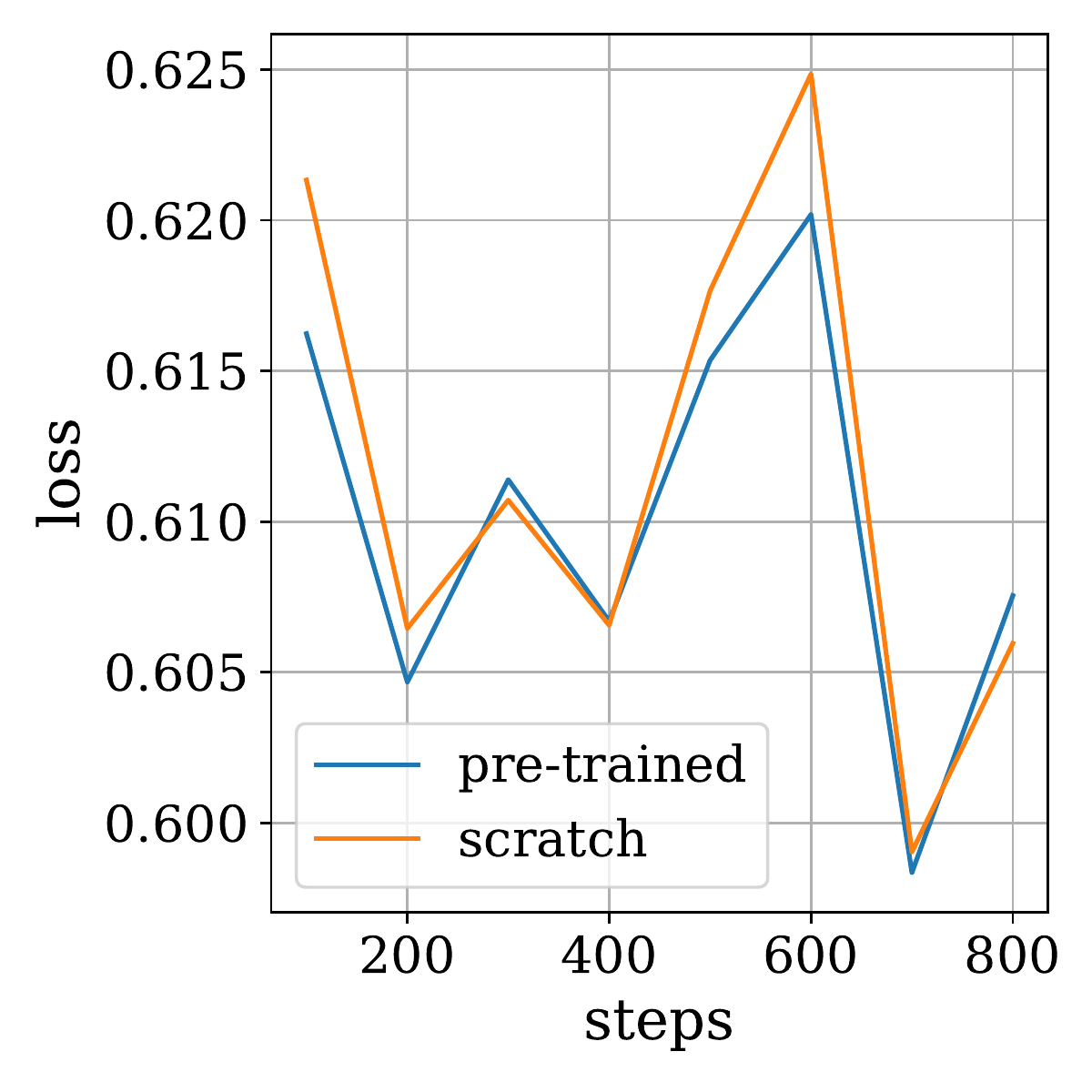}
			\caption{CoLA}
	\end{subfigure}
	\begin{subfigure}[t]{0.245\linewidth}
			\centering
			\includegraphics[width=\linewidth]{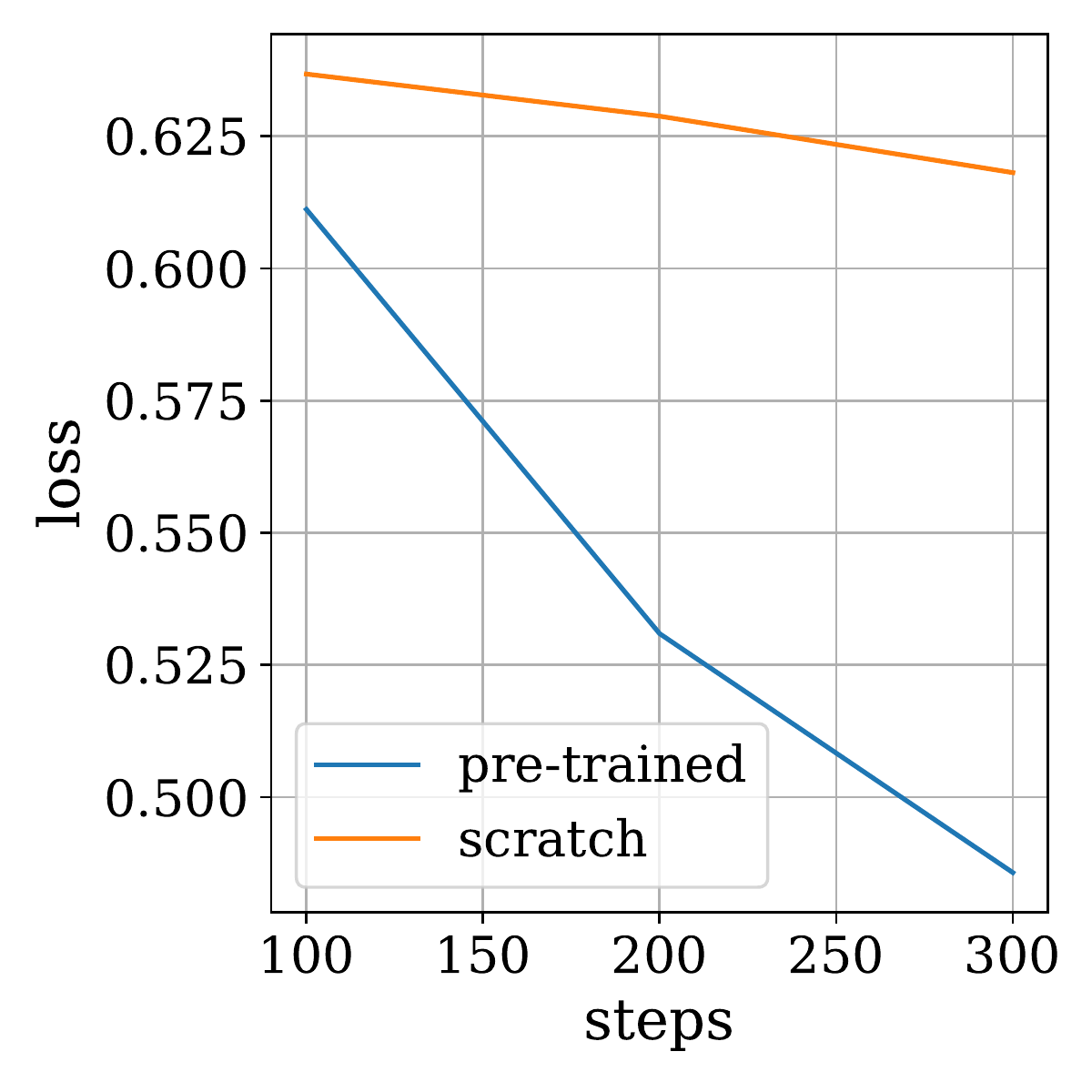}
			\caption{MRPC}
	\end{subfigure}
	\begin{subfigure}[t]{0.245\linewidth}
			\centering
			\includegraphics[width=\linewidth]{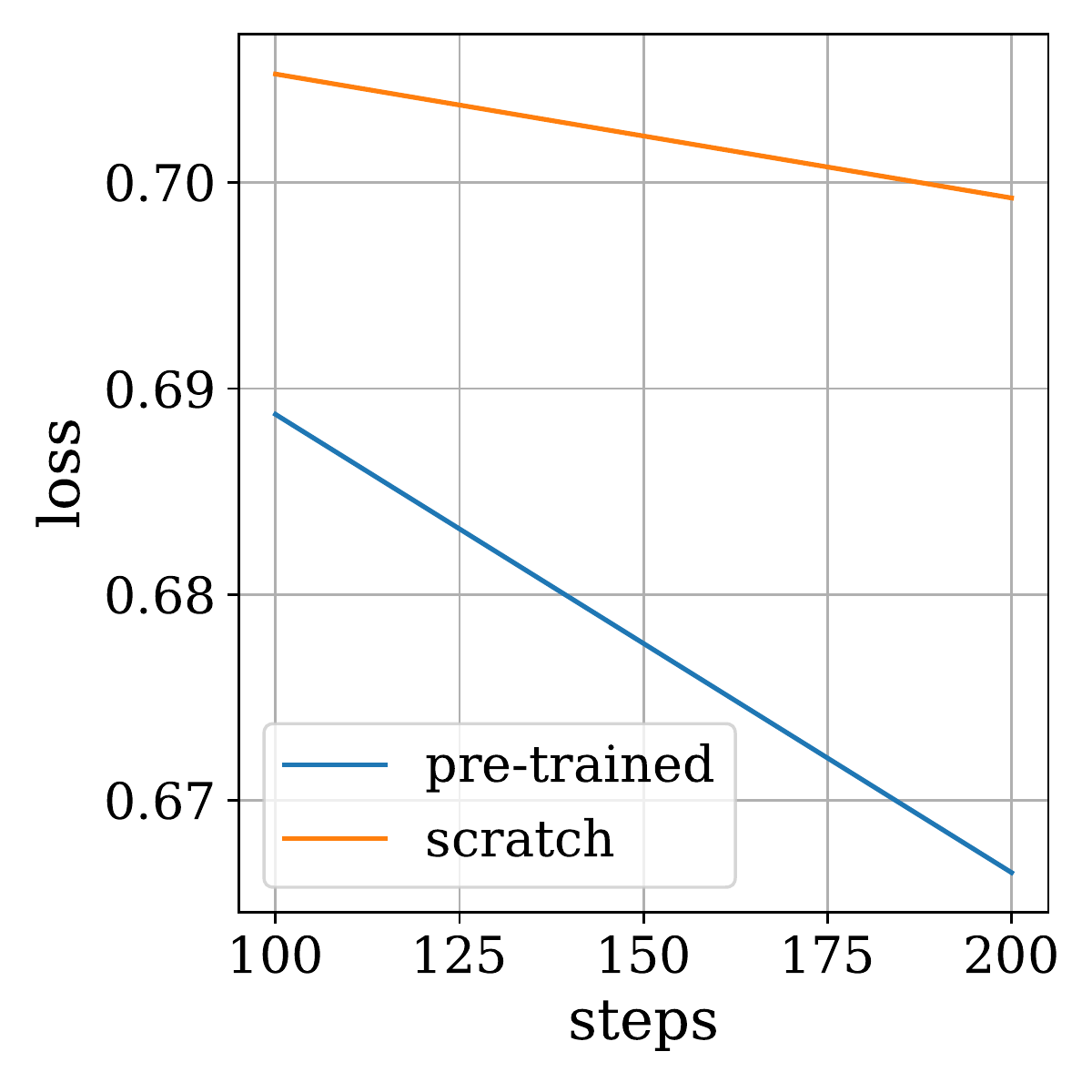}
			\caption{RTE}
	\end{subfigure}
    
    \caption{Training loss of BERT (blue line) and the models trained from scratch (orange line) on the other GLUE tasks.}
    \label{app:train_speed_glue}
\end{figure*}

\begin{figure*}[t]
        \centering
        \begin{subfigure}[t]{0.245\linewidth}
			\centering
			\includegraphics[width=\linewidth]{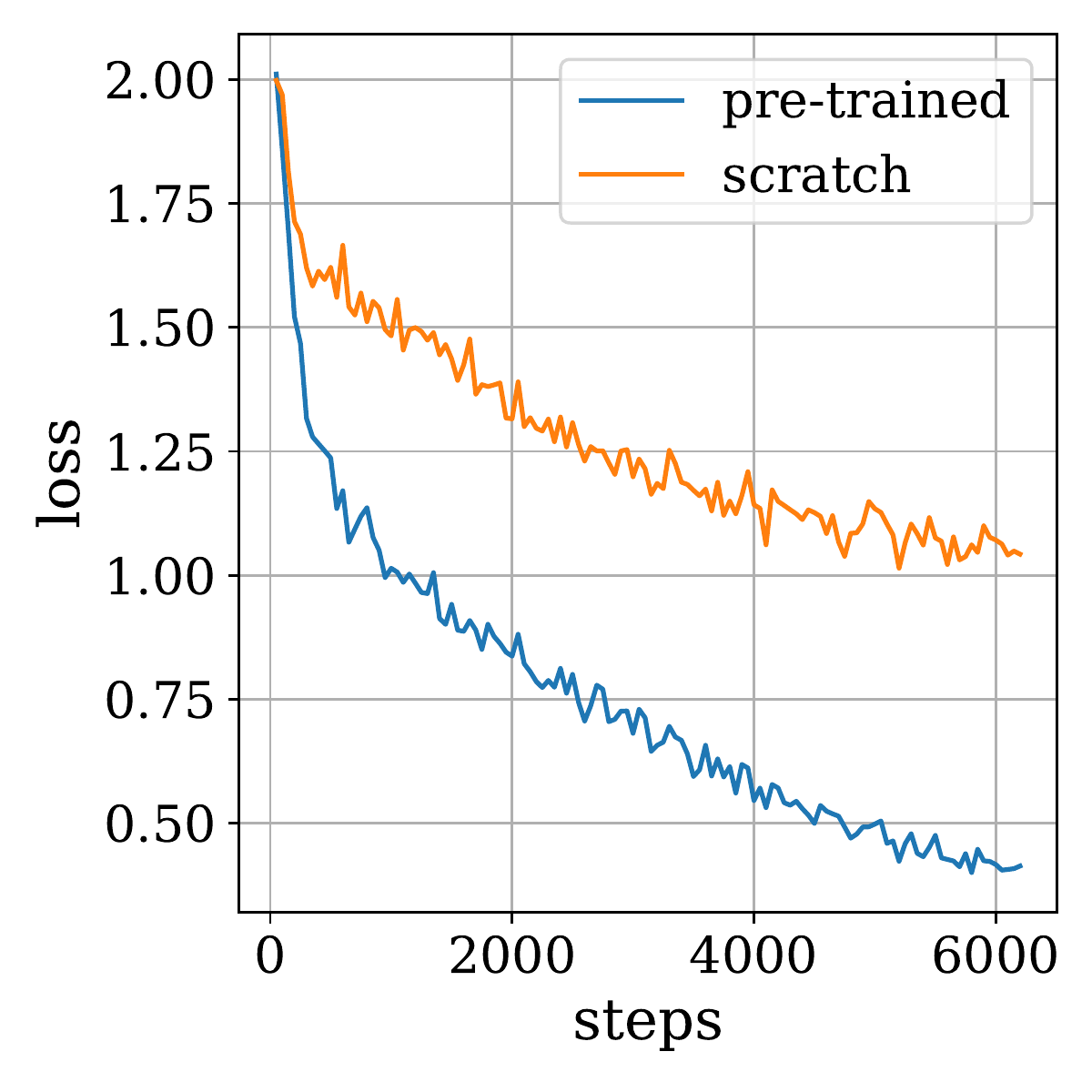}
			\caption{localization}
		\end{subfigure}
		\begin{subfigure}[t]{0.245\linewidth}
			\centering
			\includegraphics[width=\linewidth]{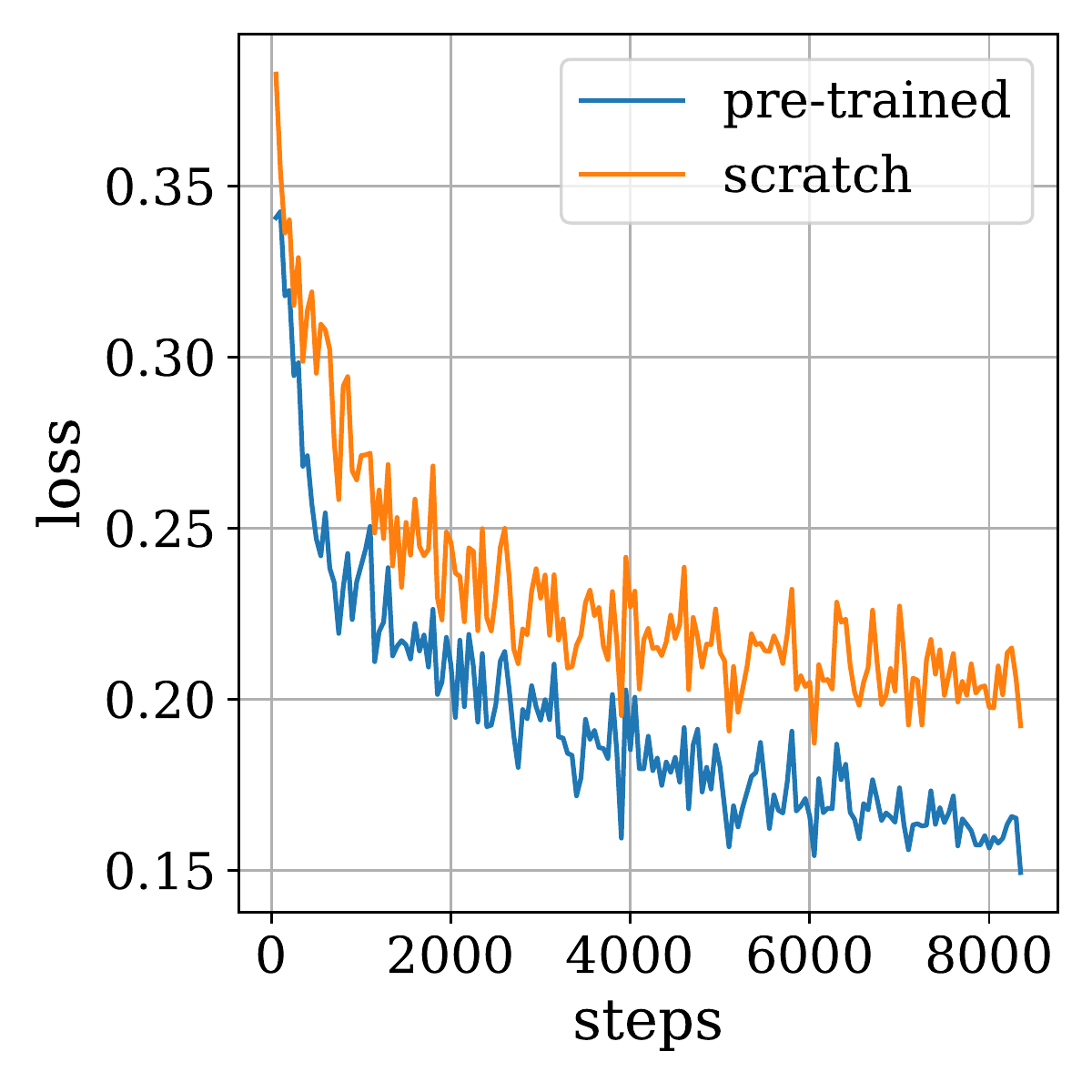}
			\caption{stability}
		\end{subfigure}
		\begin{subfigure}[t]{0.245\linewidth}
			\centering
			\includegraphics[width=\linewidth]{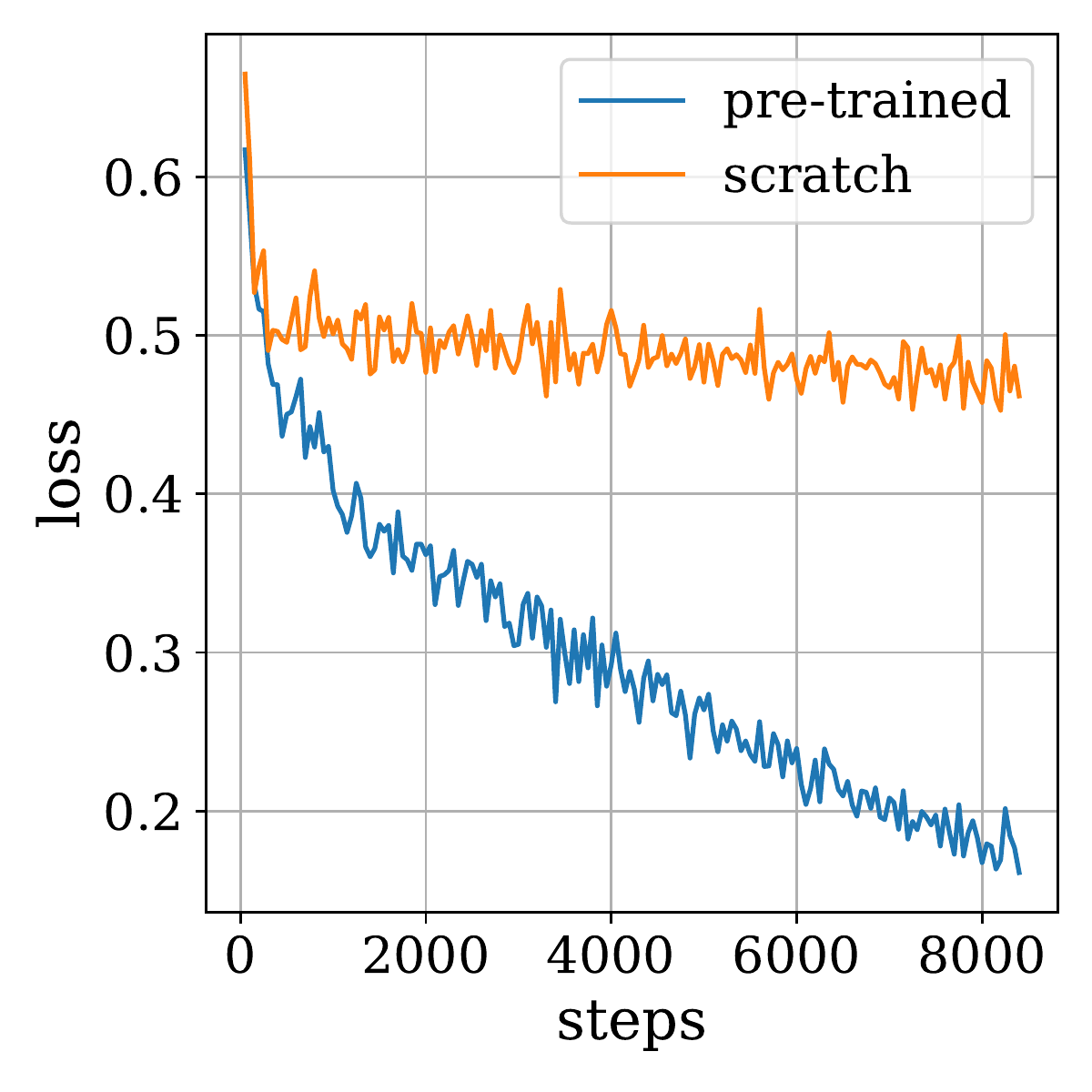}
			\caption{H3}
		\end{subfigure}
		\begin{subfigure}[t]{0.245\linewidth}
			\centering
			\includegraphics[width=\linewidth]{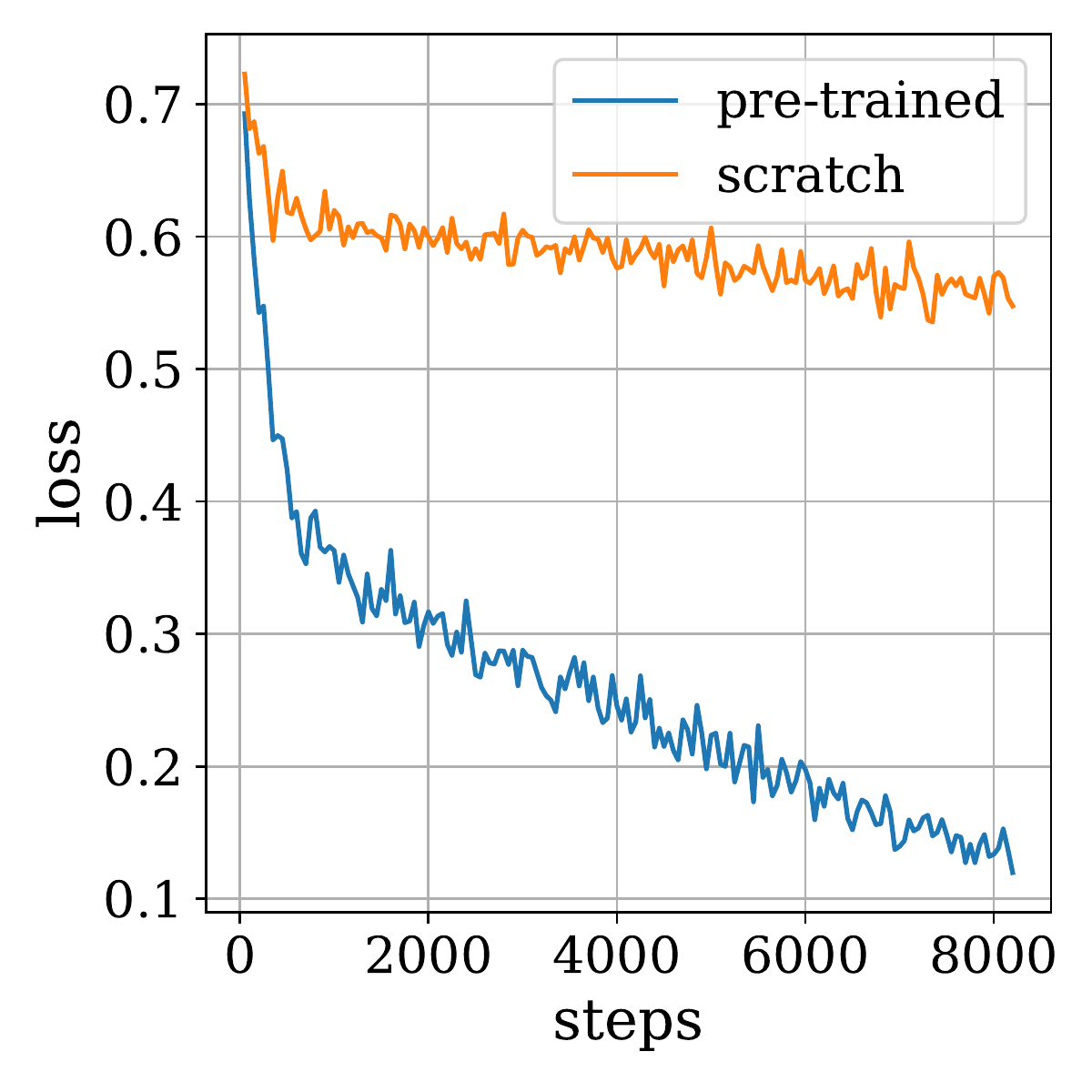}
			\caption{H4}
		\end{subfigure}
				\begin{subfigure}[t]{0.245\linewidth}
			\centering
			\includegraphics[width=\linewidth]{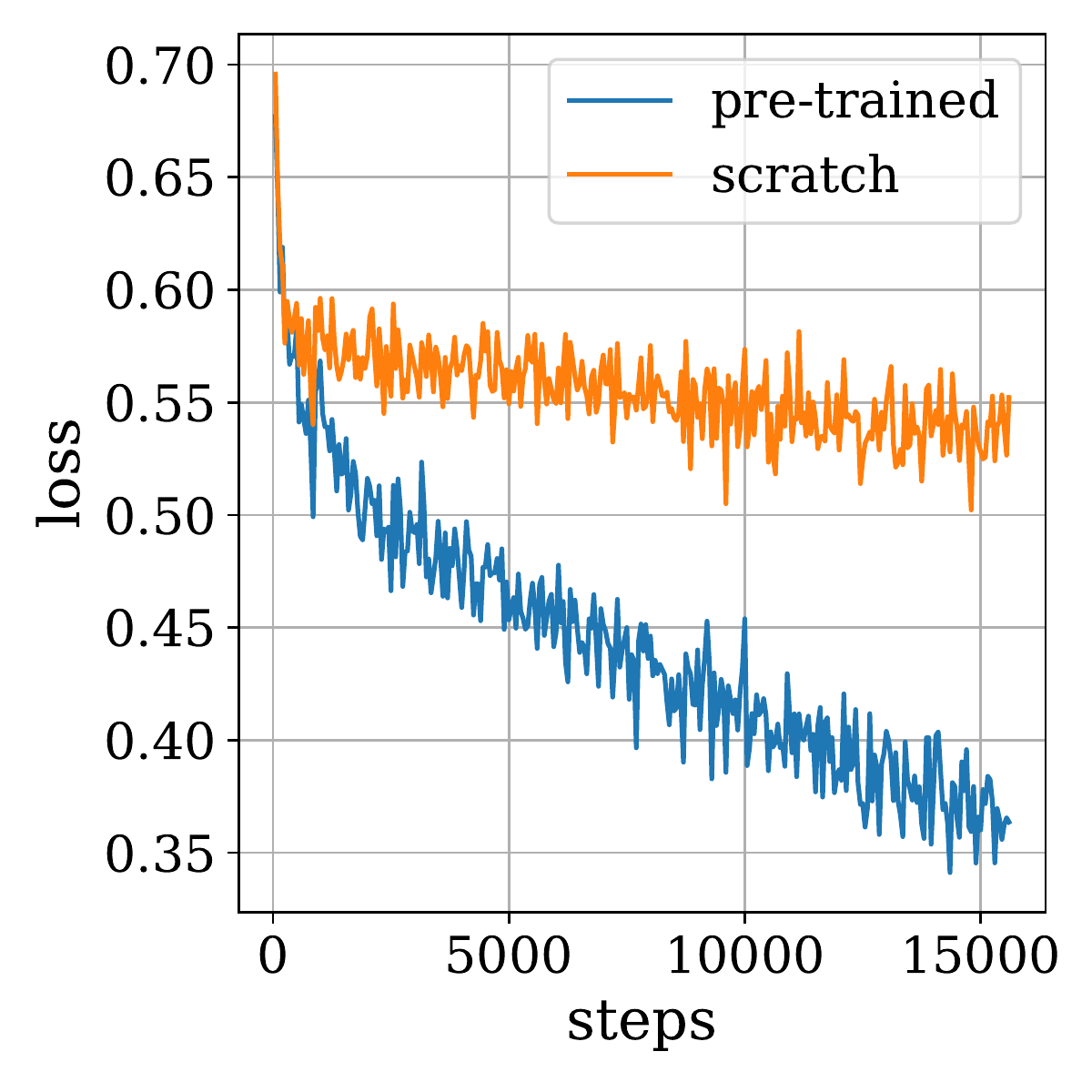}
			\caption{H3K9ac}
		\end{subfigure}
		\begin{subfigure}[t]{0.245\linewidth}
			\centering
			\includegraphics[width=\linewidth]{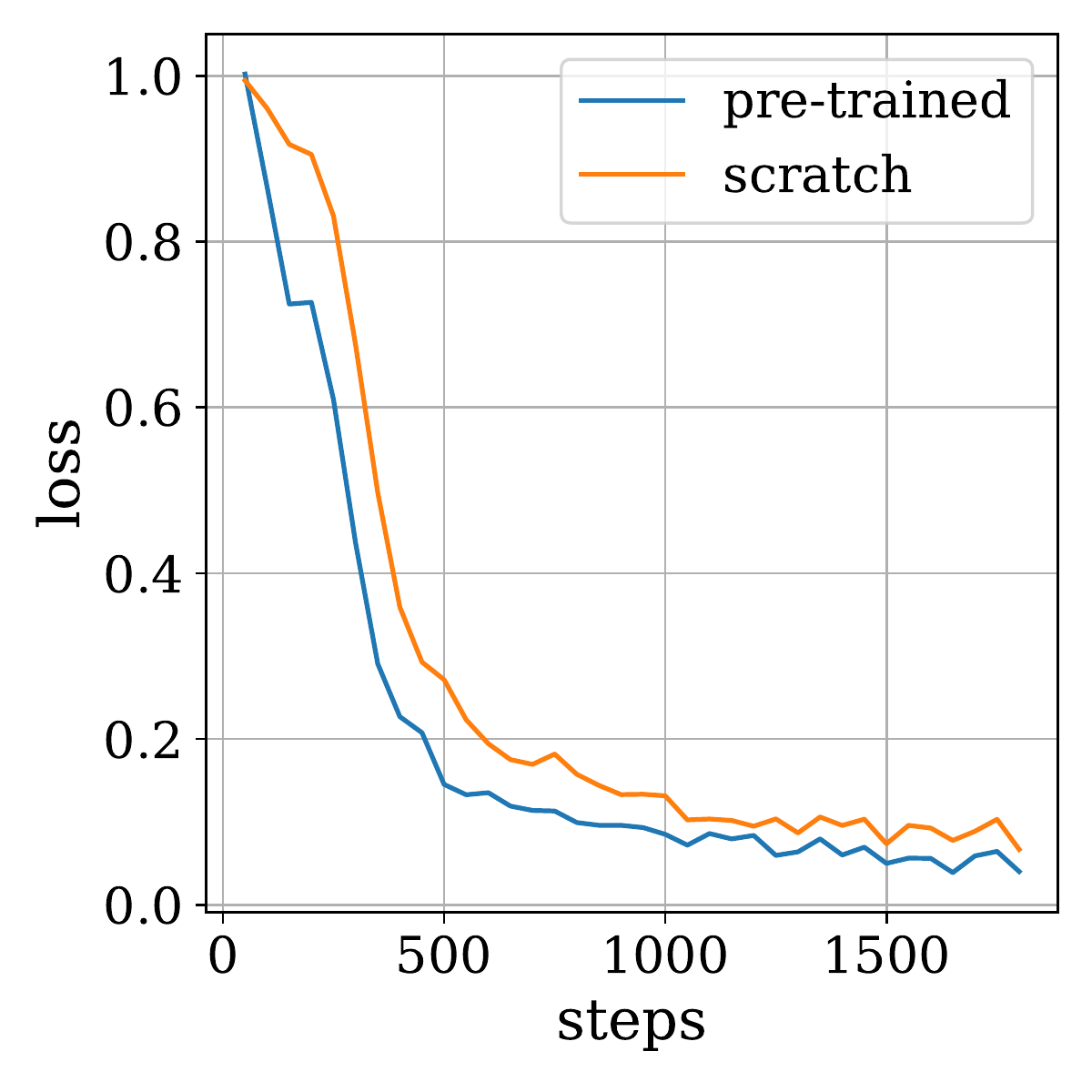}
			\caption{splice}
		\end{subfigure}
		\begin{subfigure}[t]{0.245\linewidth}
			\centering
			\includegraphics[width=\linewidth]{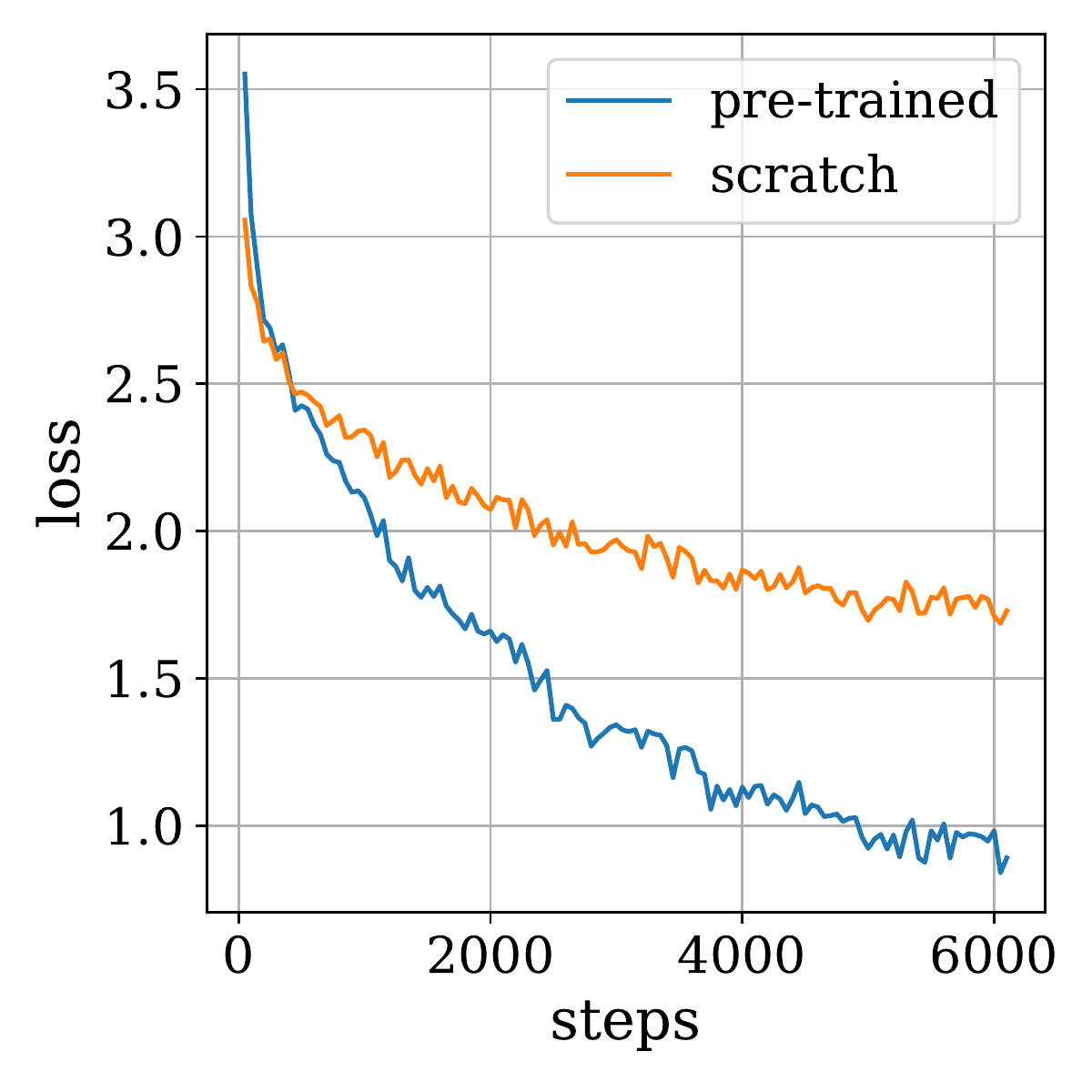}
			\caption{maestro-v1}
		\end{subfigure}
		
		\caption{Training loss of BERT (blue line) and the models trained from scratch (orange line) on the other non-text datasets.}
		\label{app:train_speed_non_text}
\end{figure*}

Figure \ref{app:train_speed_glue} and \ref{app:train_speed_non_text} show the training loss of BERT and the models trained from scratch on the other GLUE tasks and the other non-text datasets, respectively. BERT can reduce the training loss more quickly than the models trained from scratch except for the SST-2 task, on which BERT performs worse. The results are consistent over disciplines.

\section{generalization experiments for other tasks}

\begin{figure*}[t]
    \centering
    \begin{subfigure}[t]{0.245\linewidth}
			\centering
			\includegraphics[width=\linewidth]{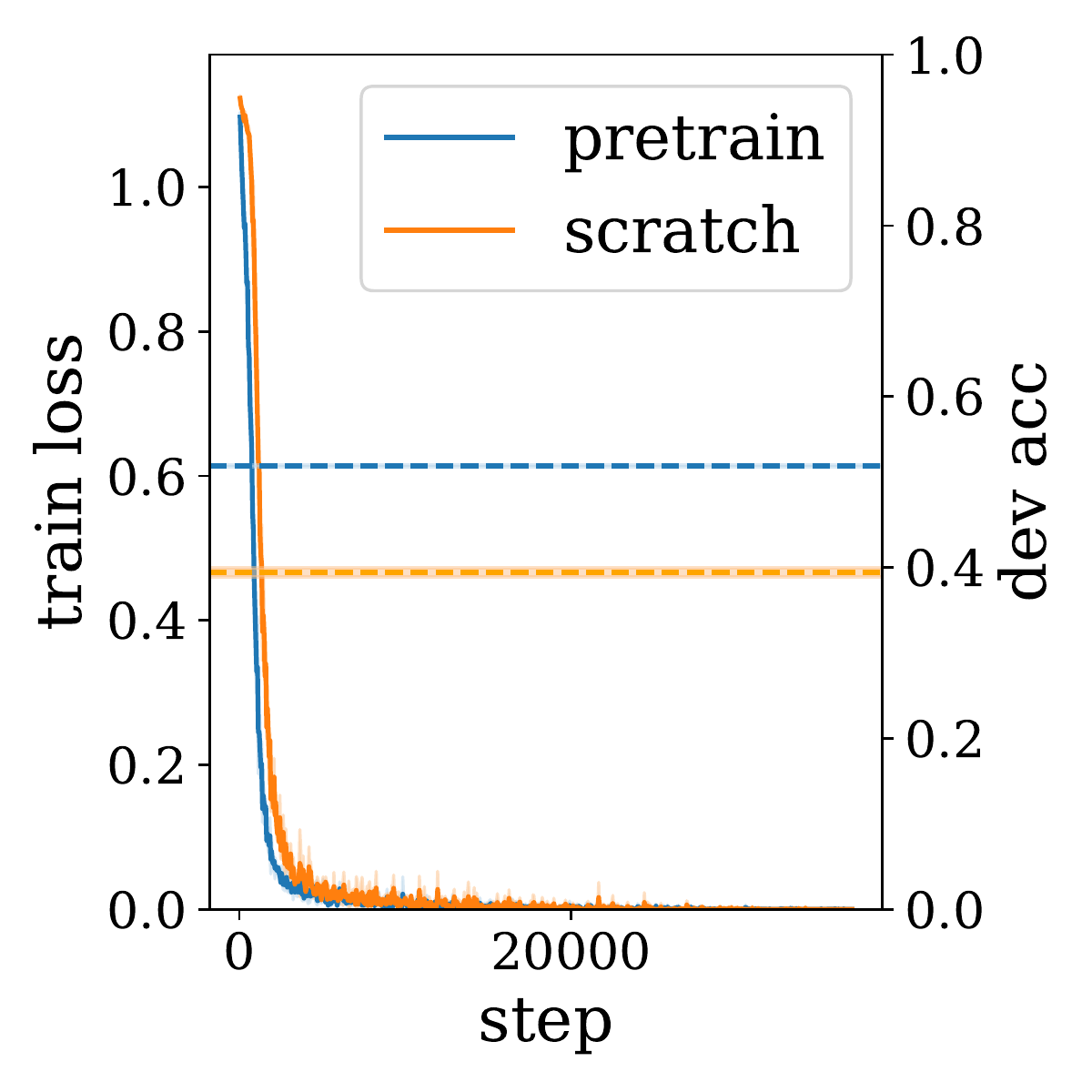}
			\caption{MNLI}
	\end{subfigure}
	\begin{subfigure}[t]{0.245\linewidth}
			\centering
			\includegraphics[width=\linewidth]{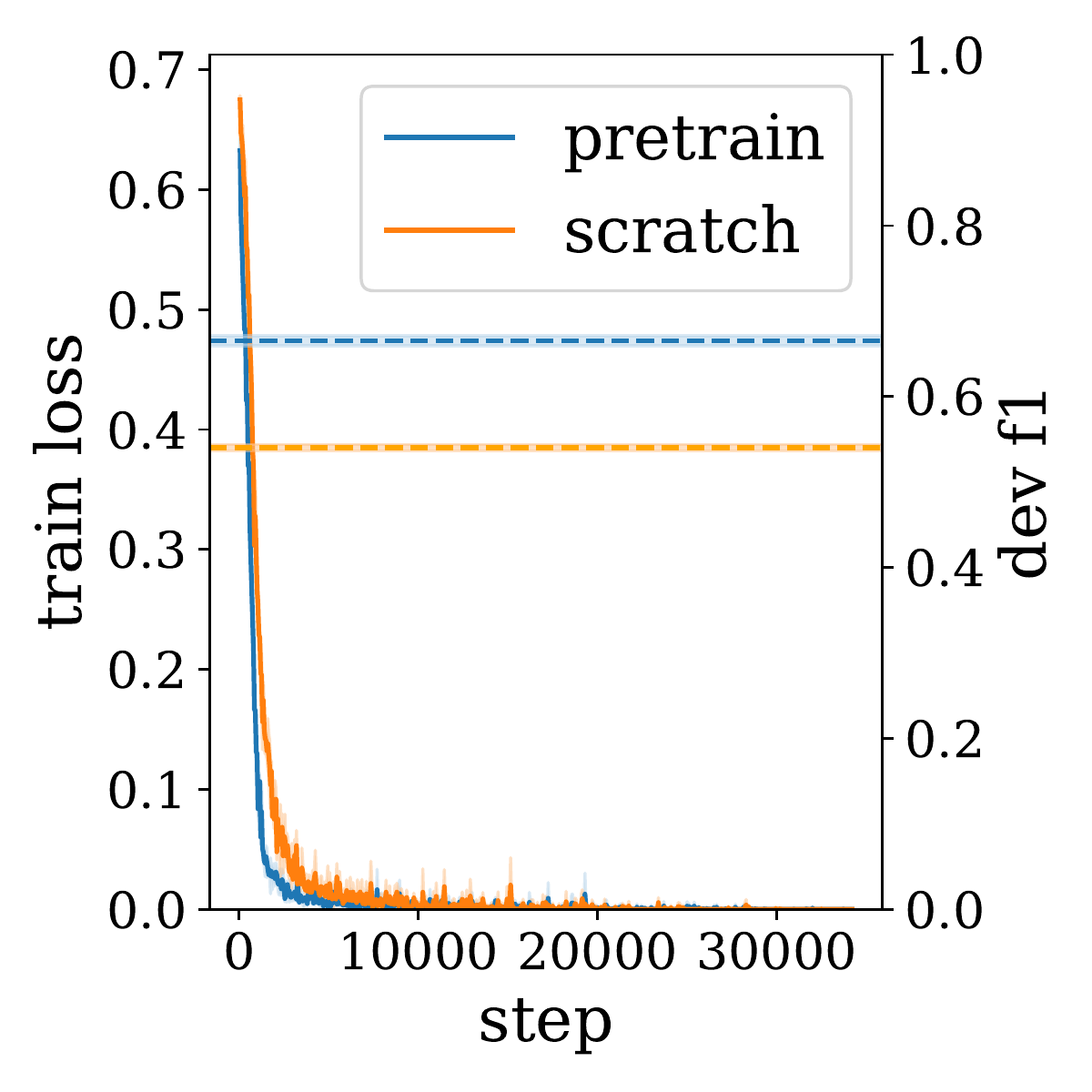}
			\caption{QQP}
	\end{subfigure}
	\begin{subfigure}[t]{0.245\linewidth}
			\centering
			\includegraphics[width=\linewidth]{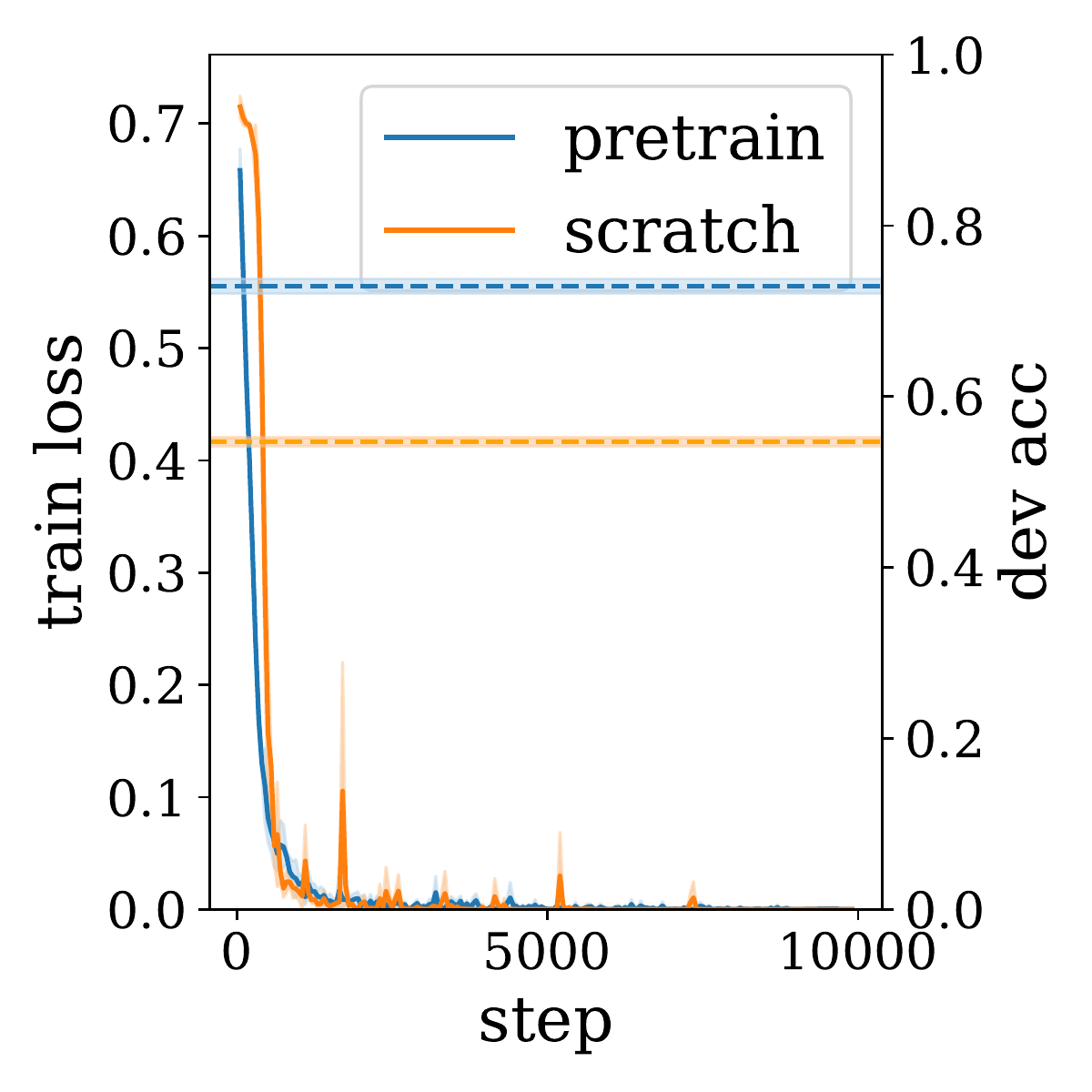}
			\caption{QNLI}
	\end{subfigure}
	\begin{subfigure}[t]{0.245\linewidth}
			\centering
			\includegraphics[width=\linewidth]{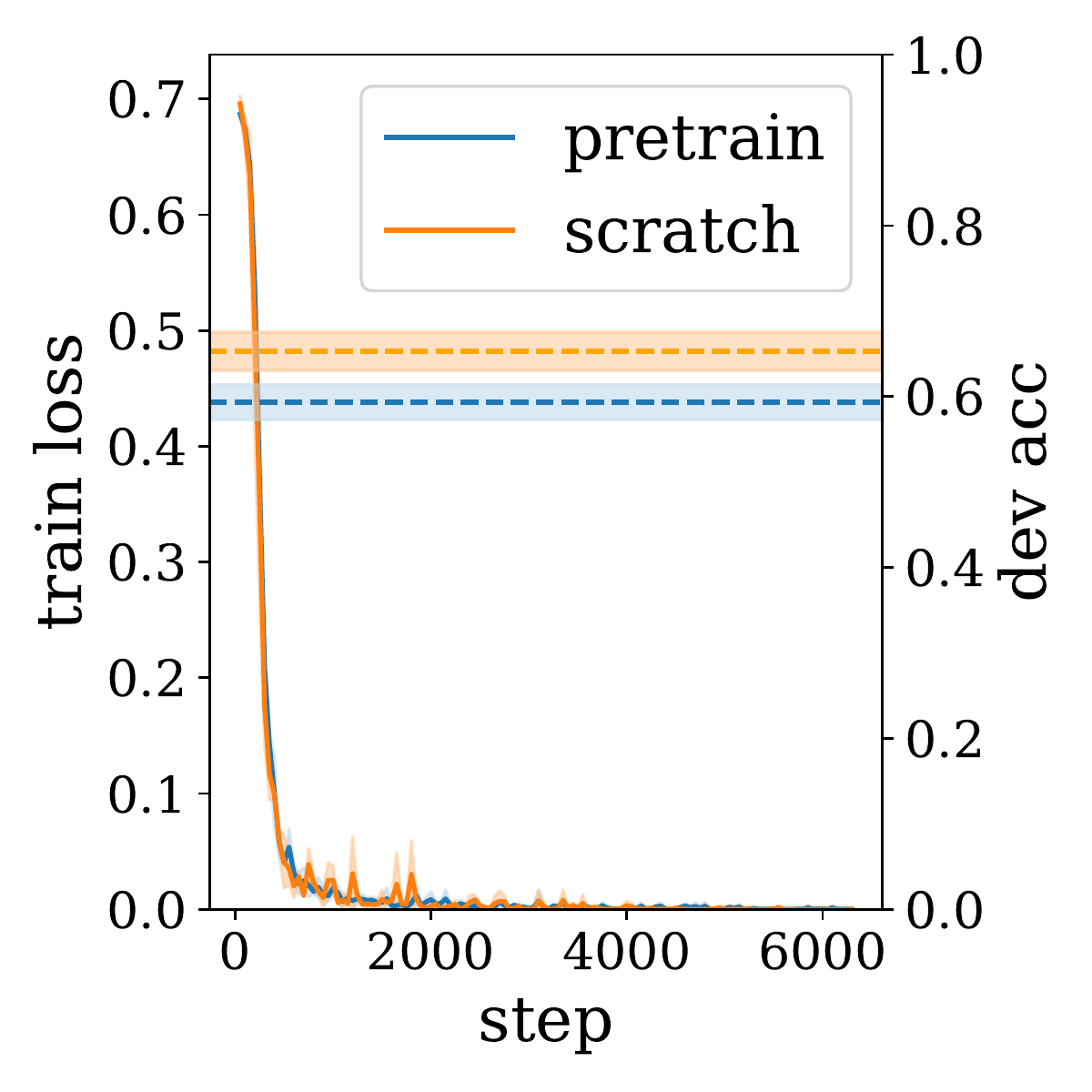}
			\caption{SST-2}
	\end{subfigure}
	\begin{subfigure}[t]{0.245\linewidth}
			\centering
			\includegraphics[width=\linewidth]{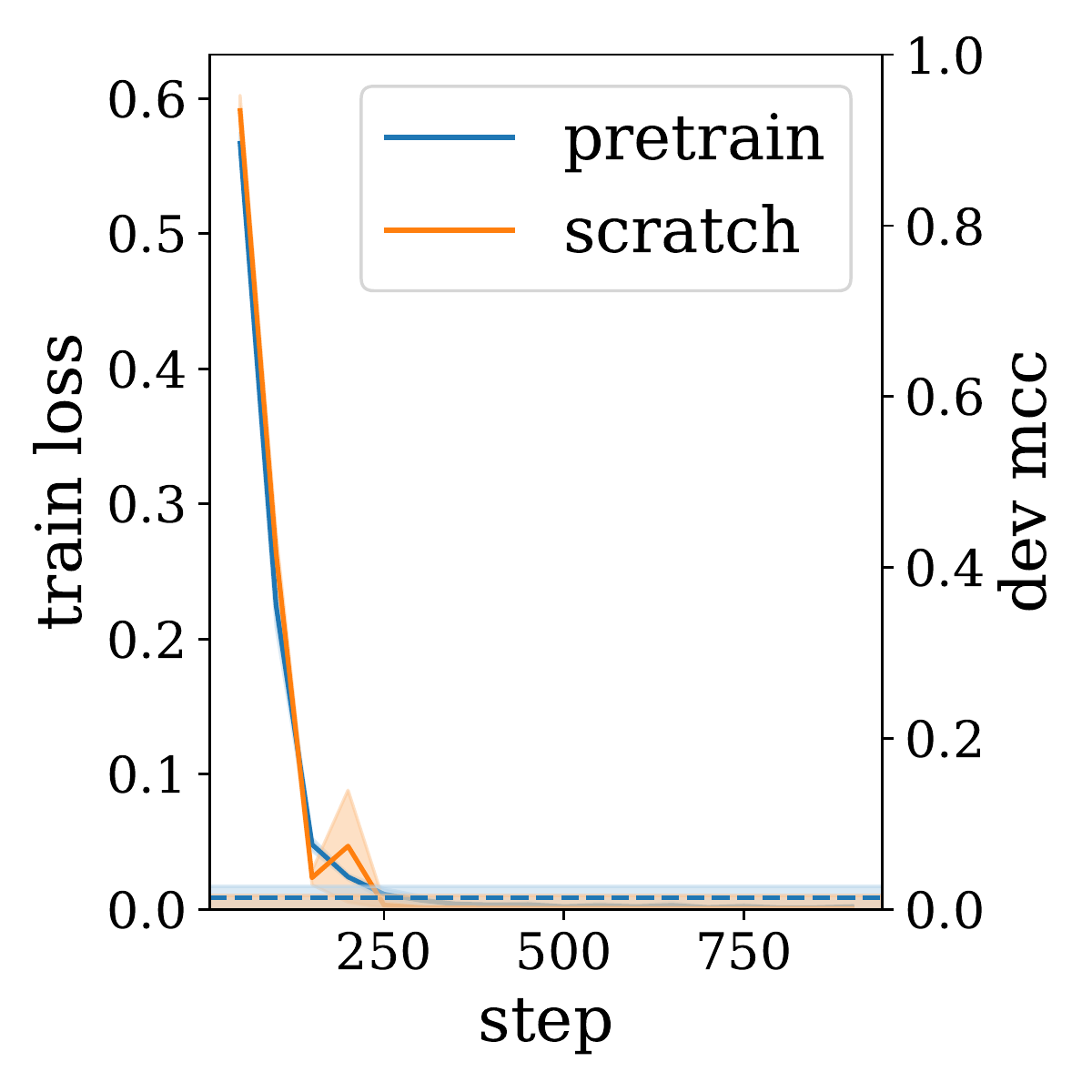}
			\caption{CoLA}
	\end{subfigure}
	\begin{subfigure}[t]{0.245\linewidth}
			\centering
			\includegraphics[width=\linewidth]{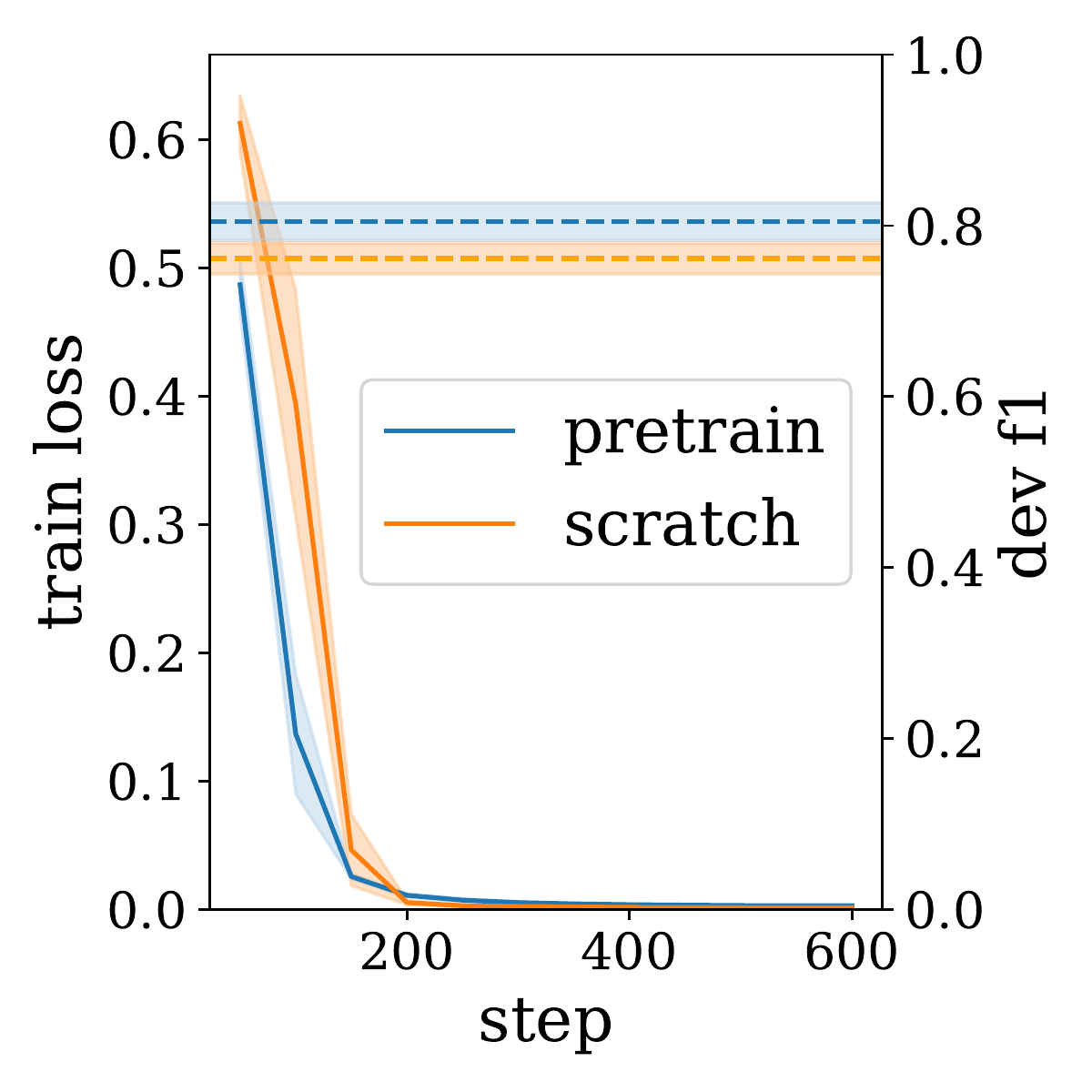}
			\caption{MRPC}
	\end{subfigure}
	\begin{subfigure}[t]{0.245\linewidth}
			\centering
			\includegraphics[width=\linewidth]{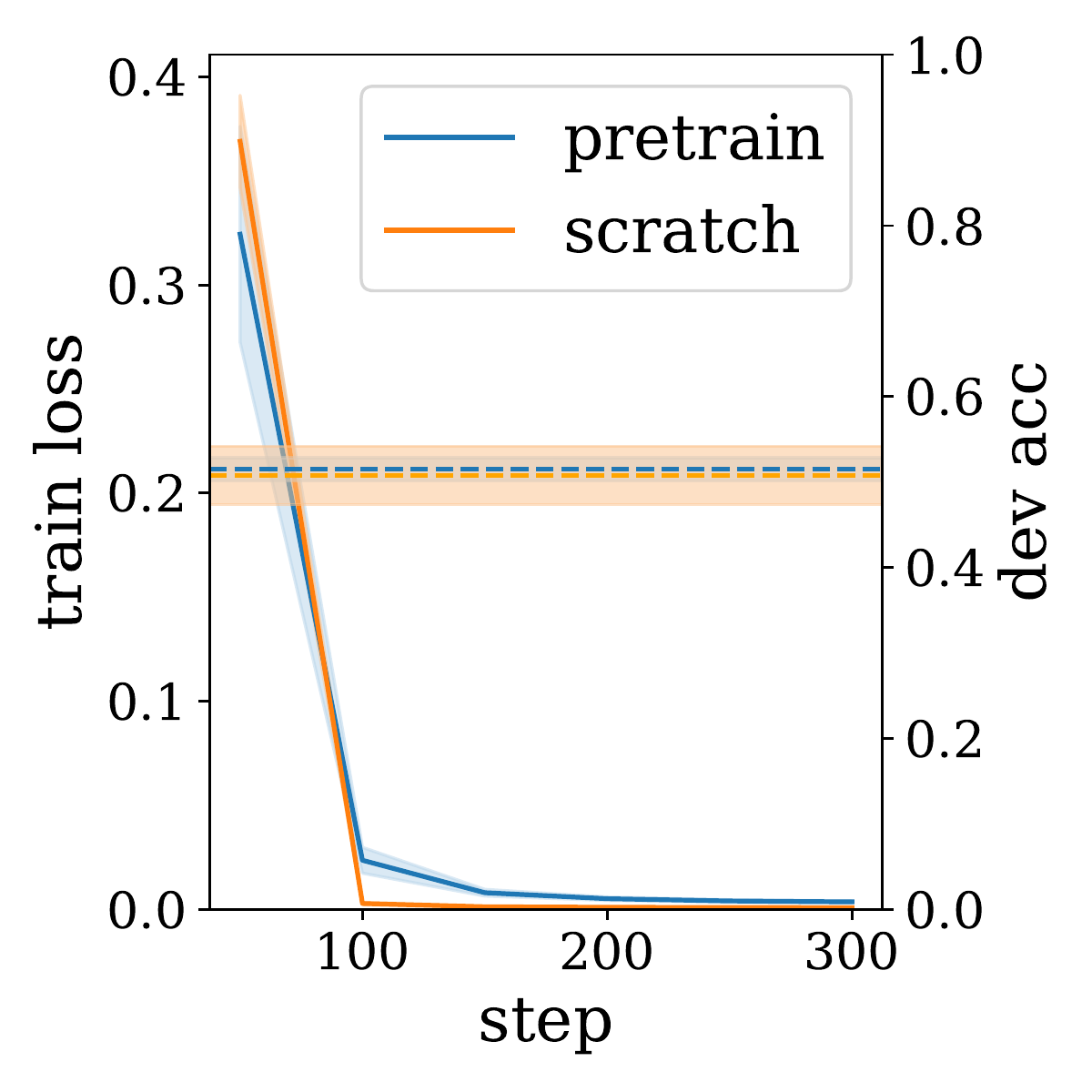}
			\caption{RTE}
	\end{subfigure}
    
    \caption{Training loss (the solid lines) and validation performance (the dashed lines) of BERT (blue line) and the models trained from scratch (orange line) on the other GLUE tasks using only 1\% training data. The last checkpoints are used to perform validation.}
    \label{app:small_glue}
\end{figure*}

\begin{figure*}[t]
        \centering
        \begin{subfigure}[t]{0.245\linewidth}
			\centering
			\includegraphics[width=\linewidth]{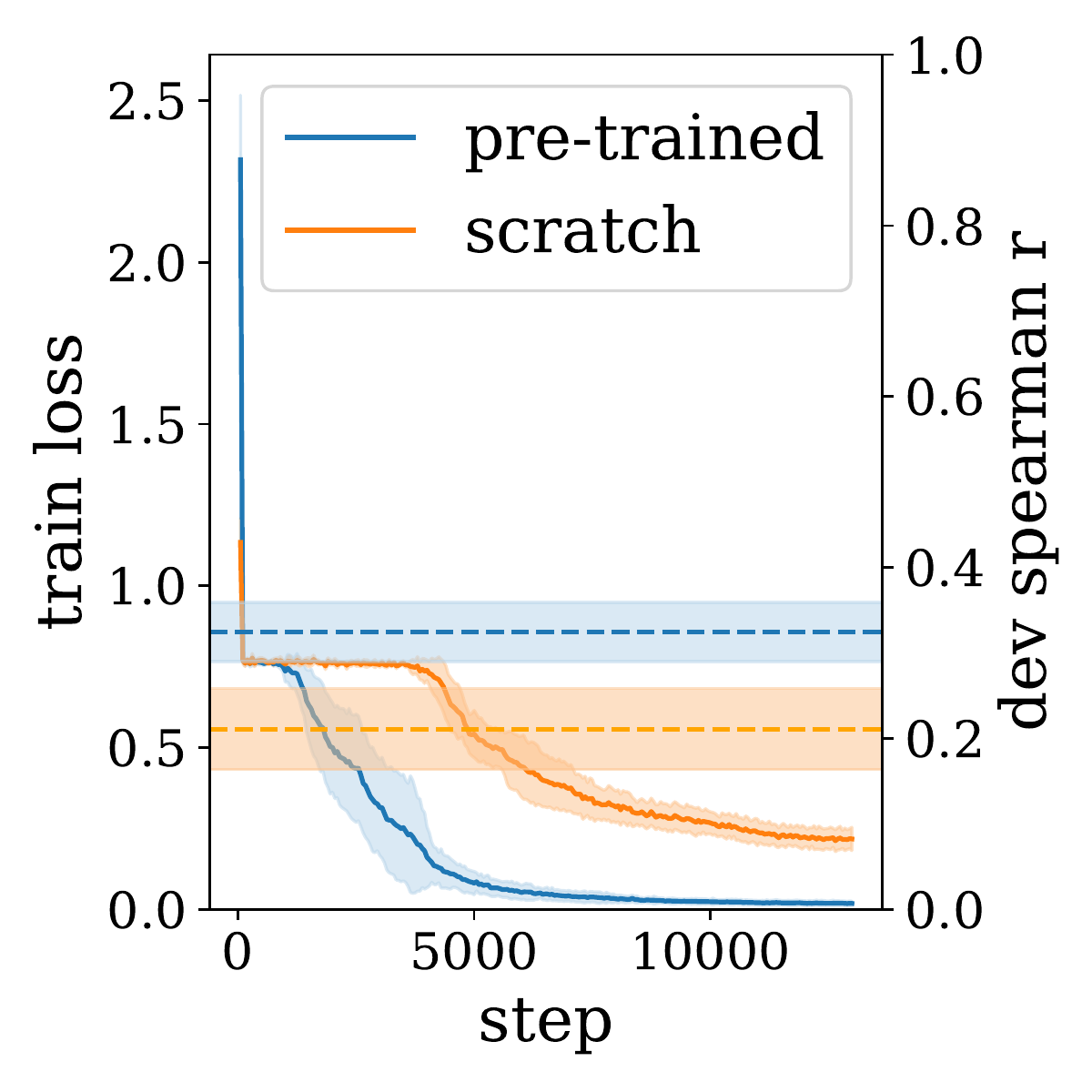}
			\caption{fluorescence}
		\end{subfigure}
		\begin{subfigure}[t]{0.245\linewidth}
			\centering
			\includegraphics[width=\linewidth]{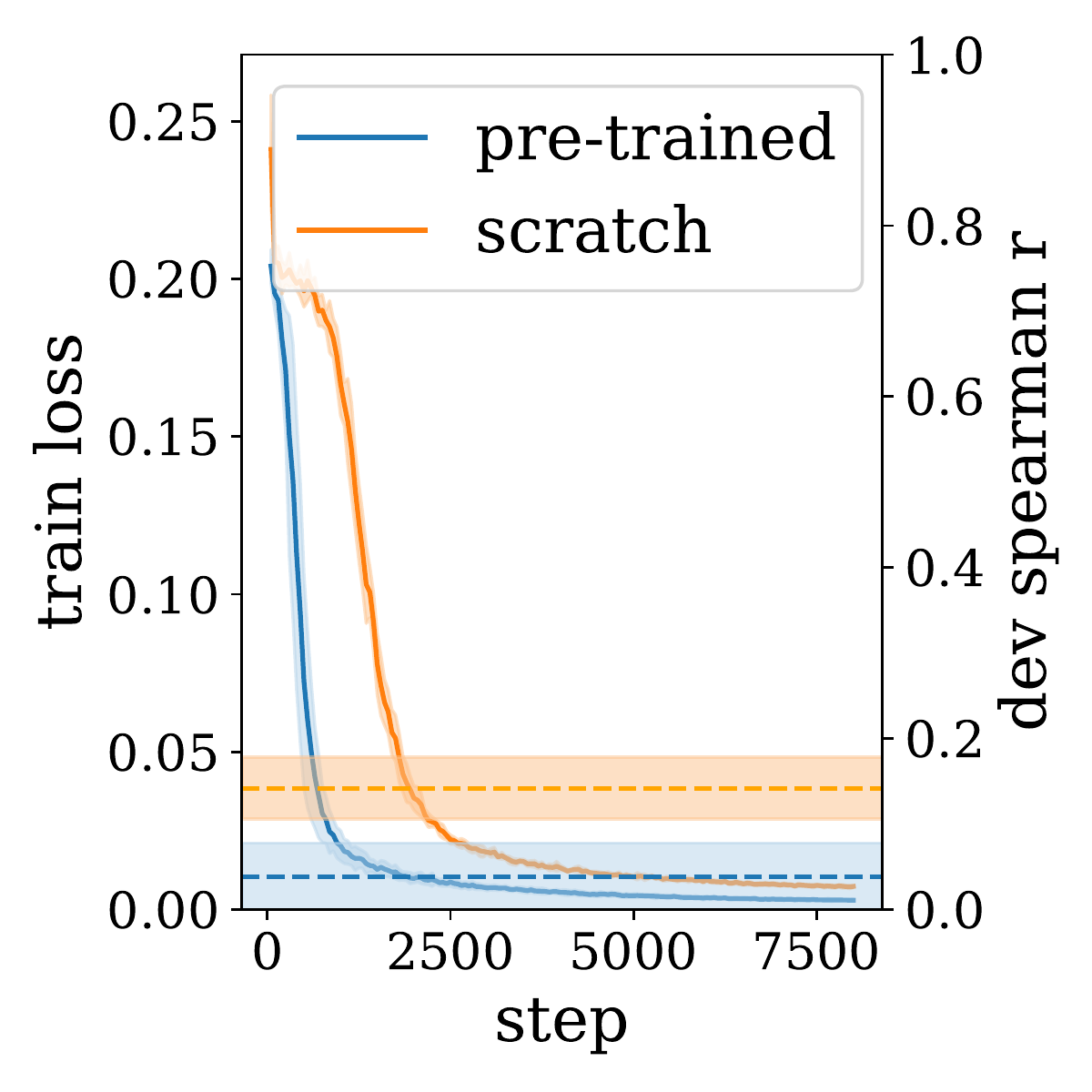}
			\caption{stability}
		\end{subfigure}
		\begin{subfigure}[t]{0.245\linewidth}
			\centering
			\includegraphics[width=\linewidth]{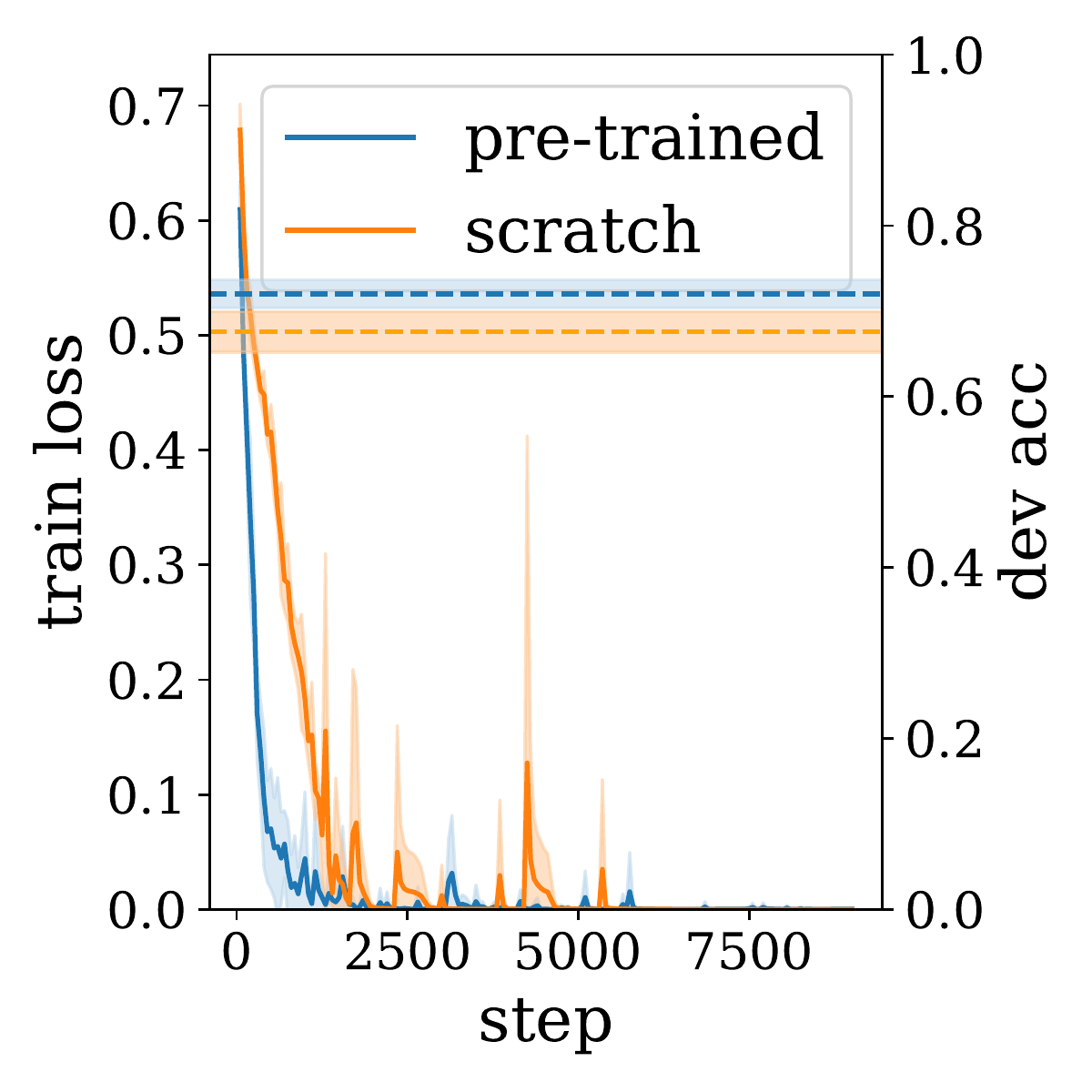}
			\caption{H3}
		\end{subfigure}
		\begin{subfigure}[t]{0.245\linewidth}
			\centering
			\includegraphics[width=\linewidth]{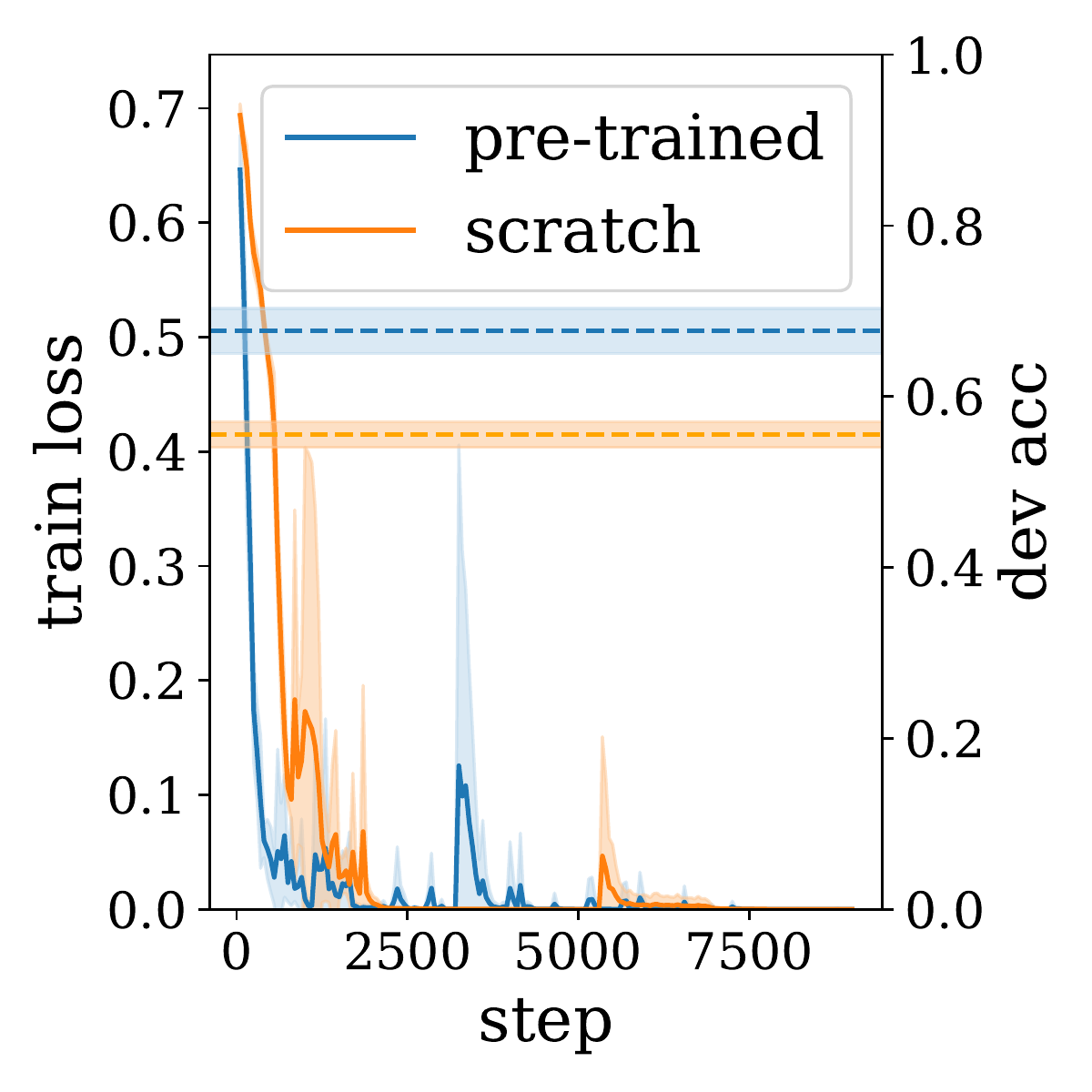}
			\caption{H4}
		\end{subfigure}
		
		\caption{Training loss (the solid lines) and validation performance (the dashed lines) of BERT (blue line) and the models trained from scratch (orange line) on the other non-text datasets using only 1\% training data. We do not conduct the experiments on splice and maestro-v1 since the 1\% training sets are too small.}
		\label{app:small_non_text}
\end{figure*}

Figure \ref{app:small_glue} and \ref{app:small_non_text} shows the results of BERT and the model trained from scratch with only 1\% training data of the GLUE dataset and the non-text datasets. We do not conduct the experiment on splice and maestro-v1 datasets due to the limitation of the size of the training sets. For most of the tasks, BERT generalizes better than the models trained from scratch.

\section{Detailed results of section 4.4}

\subsection{Dynamical isometry}

\begin{figure*}[t]
    \centering
    \begin{subfigure}[t]{0.32\linewidth}
        \centering
        \includegraphics[width=\linewidth]{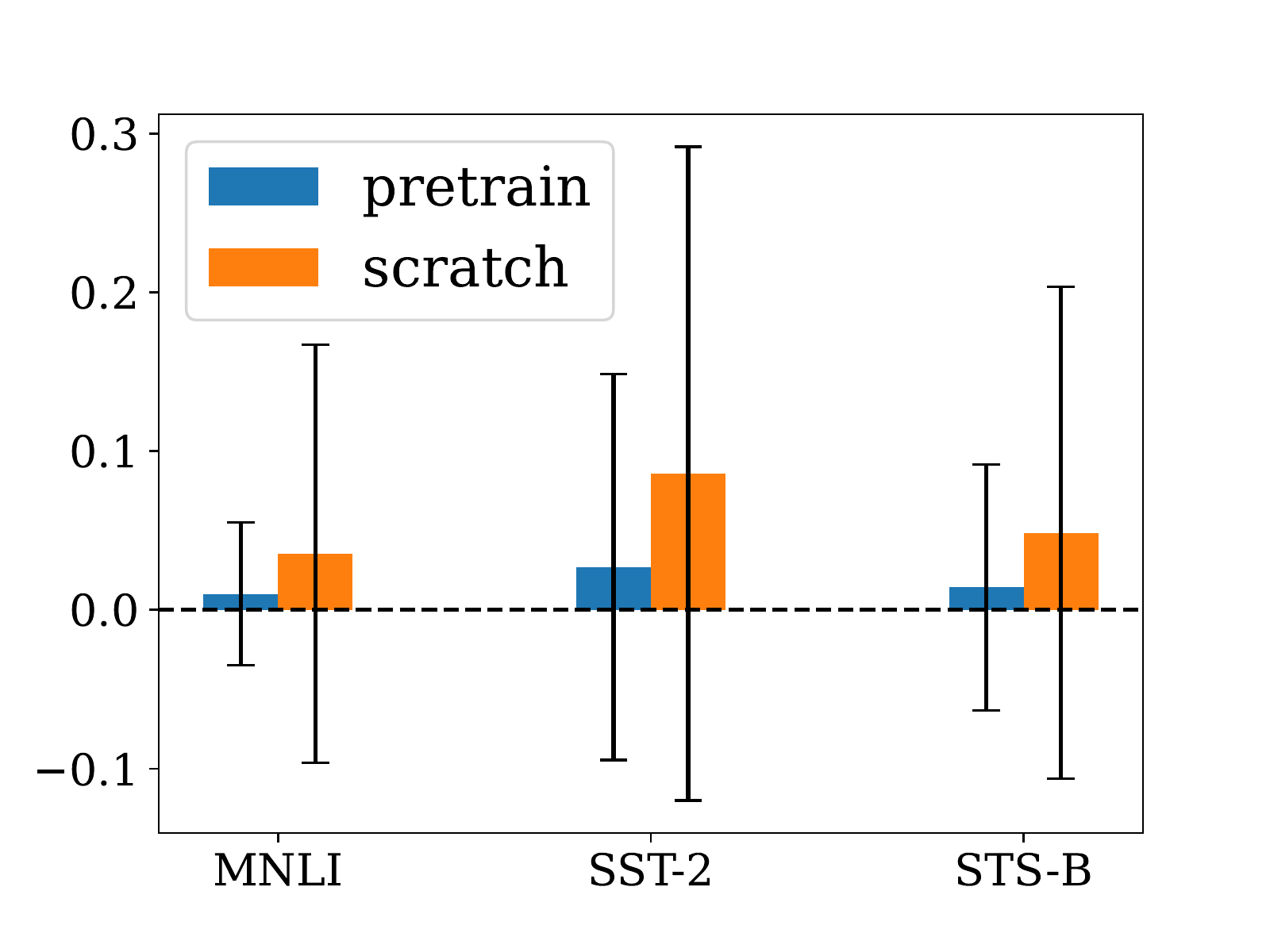}
        \caption{BERT-base}
    \end{subfigure}
    \begin{subfigure}[t]{0.32\linewidth}
        \centering
        \includegraphics[width=\linewidth]{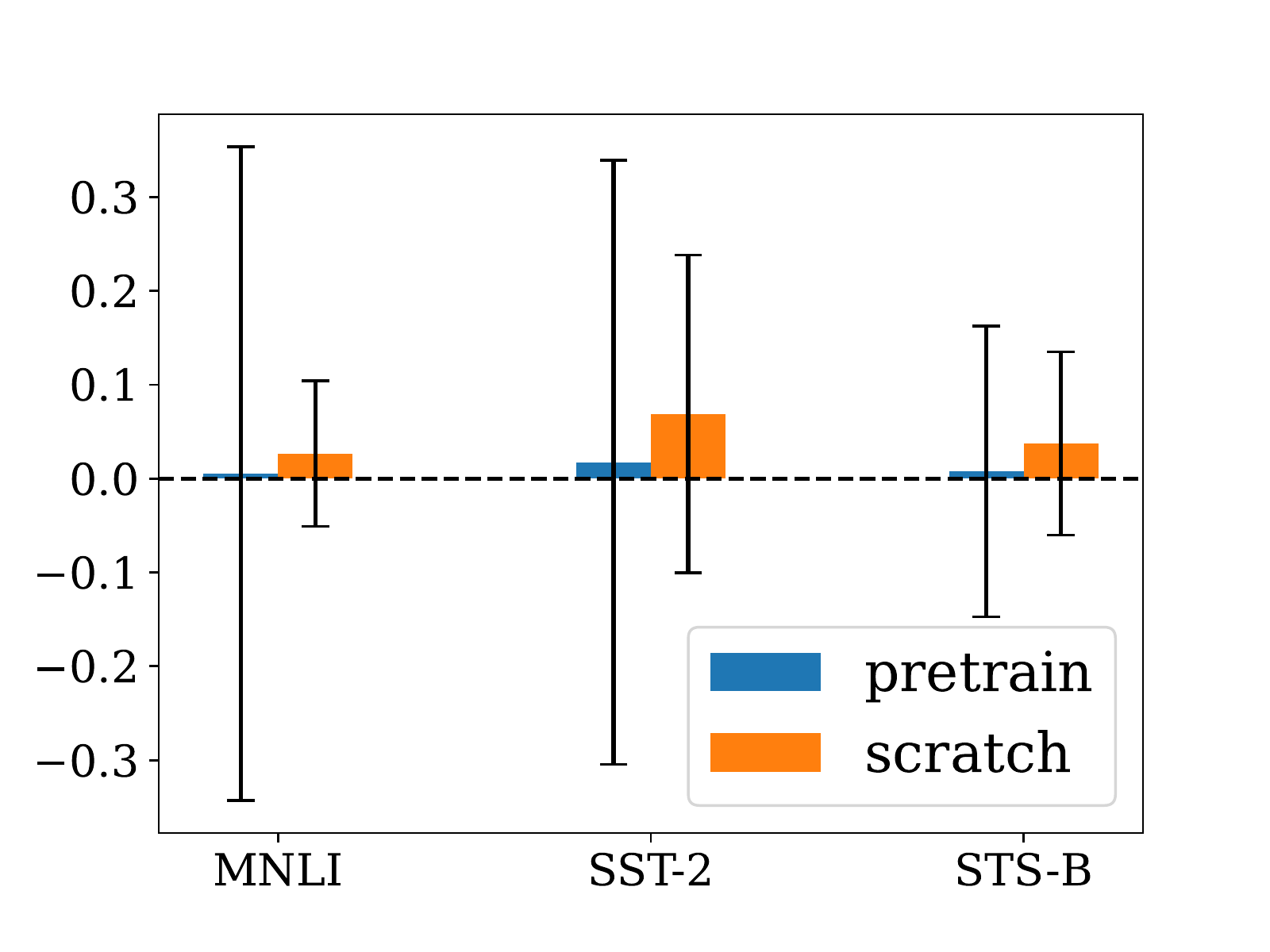}
        \caption{BERT-large}
    \end{subfigure}
    \begin{subfigure}[t]{0.32\linewidth}
        \centering
        \includegraphics[width=\linewidth]{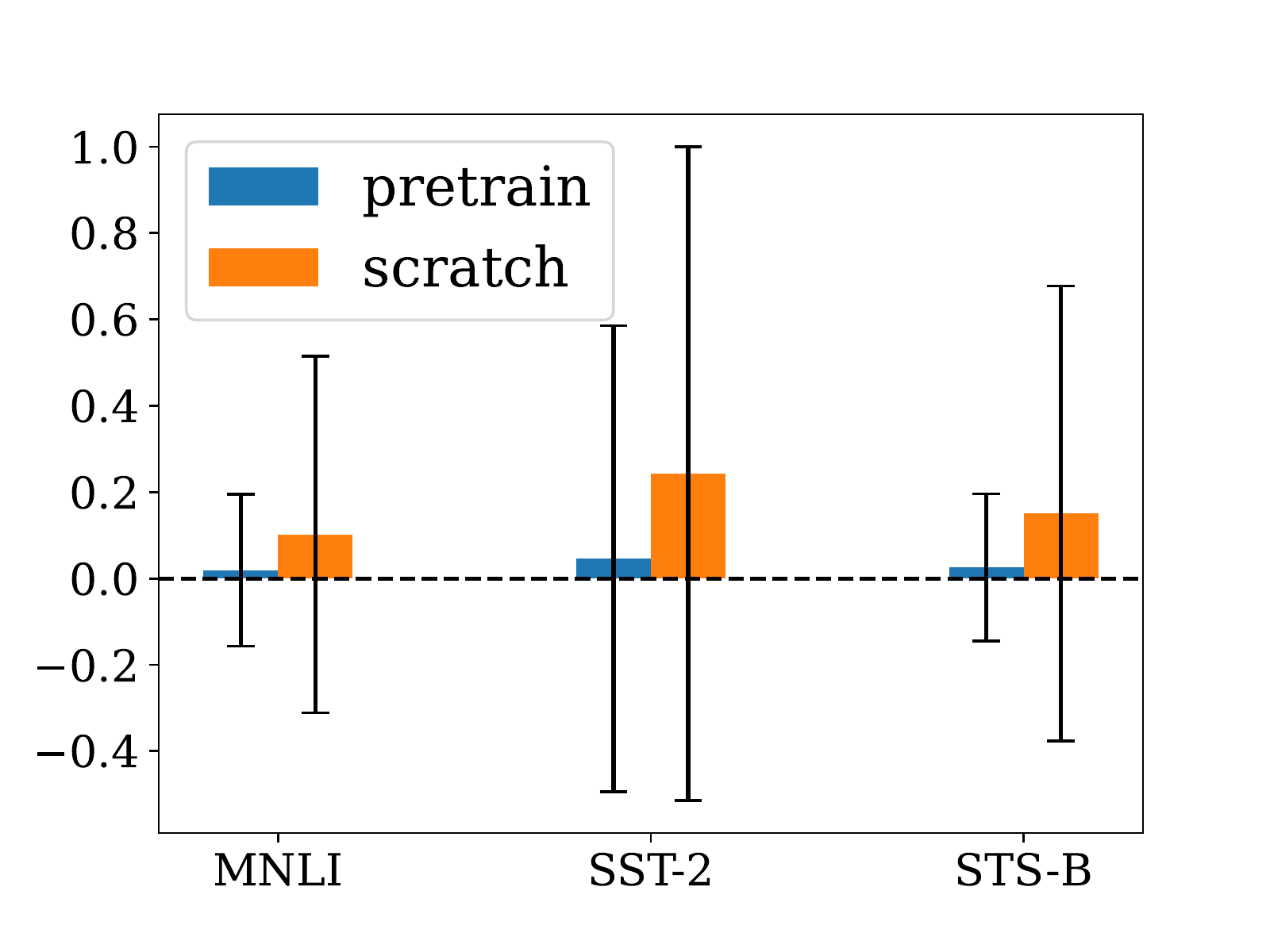}
        \caption{albert-base}
    \end{subfigure}
    \caption{Distributions of the singular values of the output-input jacobian matrices of BERT and ALBERT. The singular values are calculated on GLUE dataset. The bars stand for mean and the error bars stand for standard deviation.}
    \label{app:dynamical}
\end{figure*}

Figure \ref{app:dynamical} shows the distribution of the singular values of the output-input jacobian matrices of BERT-base, BERT-large, and ALBERT-base. The jacobian matrices are computed by calculating the derivative of the representation from the last layer with respect to the input word embeddings. And the input data is from normal GLUE dataset. Compared to the random initialization (scratch in fig \ref{app:dynamical}), the singular values of BERT and ALBERT concentrate at zero but not one, which is opposite to the hypothesis of dynamical isometry. Therefore, it is hard to claim that the power of BERT and ALBERT originates from dynamical isometry.

\subsection{Gradient confusion}

\begin{figure*}[t]
    \centering
    \begin{subfigure}[t]{0.49\linewidth}
        \centering
        \includegraphics[width=0.9\linewidth]{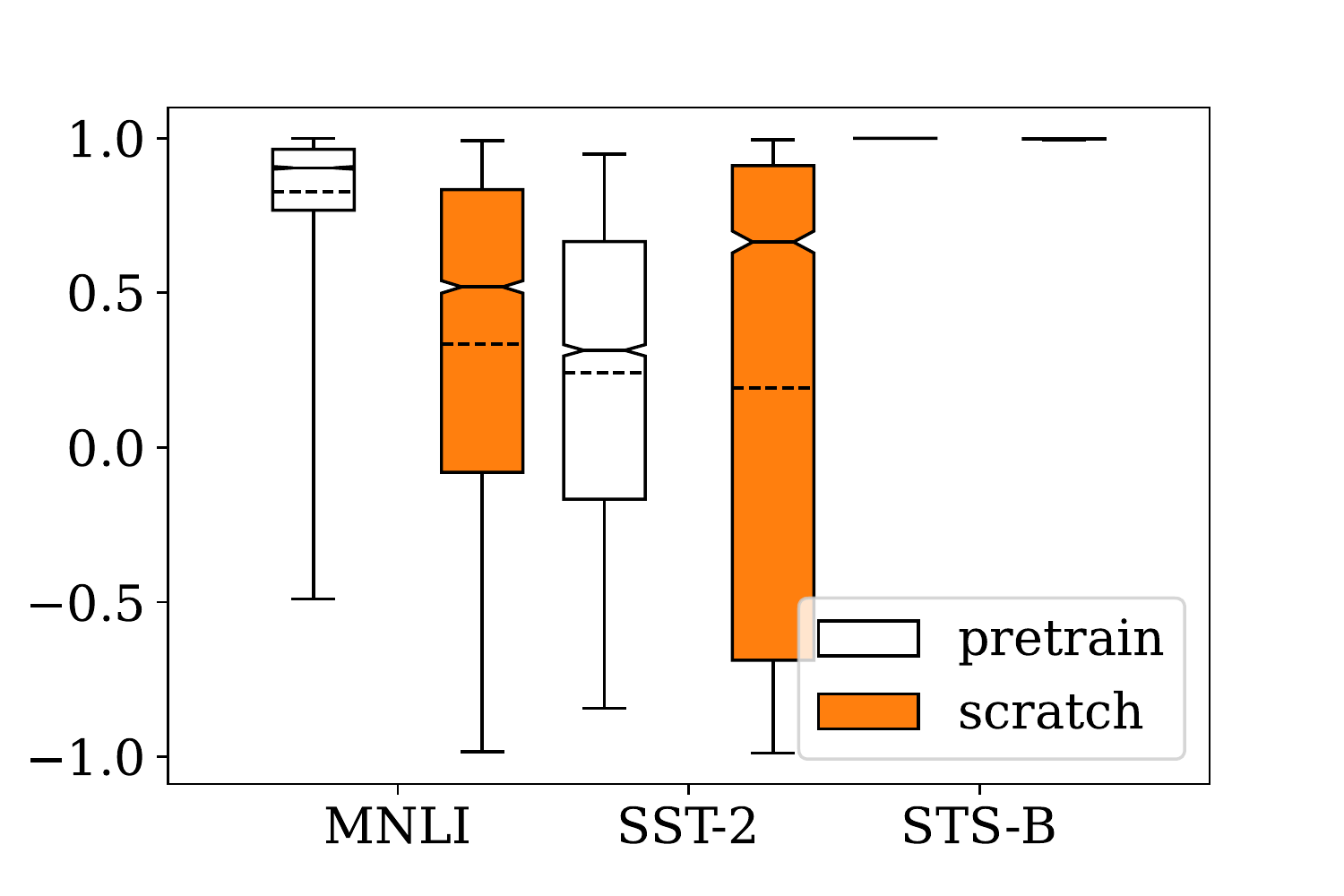}
        \caption{BERT-base}
    \end{subfigure}
    \begin{subfigure}[t]{0.49\linewidth}
        \centering
        \includegraphics[width=0.9\linewidth]{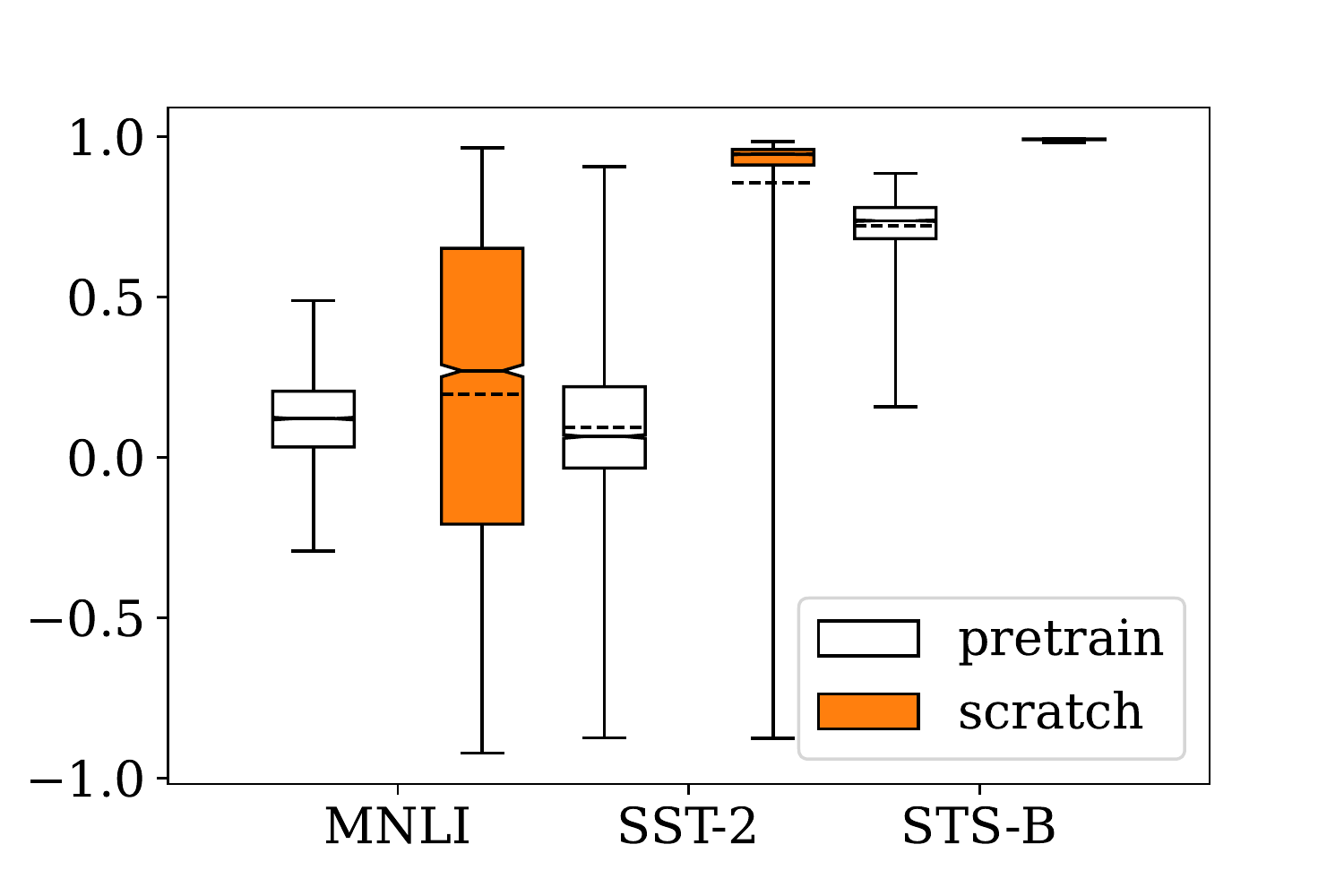}
        \caption{ALBERT-base}
    \end{subfigure}
    \caption{gradient cosine similarity of (a) BERT and (b) ALBERT on the synthetic GLUE dataset. The notches stand for median, and the dashed lines without notches stand for mean.}
    \label{app:confusion}
\end{figure*}

Figure \ref{app:confusion} shows the cosine similarity of gradients produced by different data points in synthetic GLUE dataset. Although the cosine similarity of BERT is larger than the random initialized (scratch) counterpart, ALBERT shows adverse trends. The cosine similarity of pre-trained ALBERT is smaller than the scratch counterpart. But pre-trained ALBERT still outperforms the random initialization, which indicates that avoiding gradient confusion may not be the key to pre-trained MLMs' \textit{discipline adaptability}. 

\subsection{Output variance under perturbation}

\begin{figure*}[t]
    \centering
    \begin{subfigure}[t]{0.32\linewidth}
        \centering
        \includegraphics[width=\linewidth]{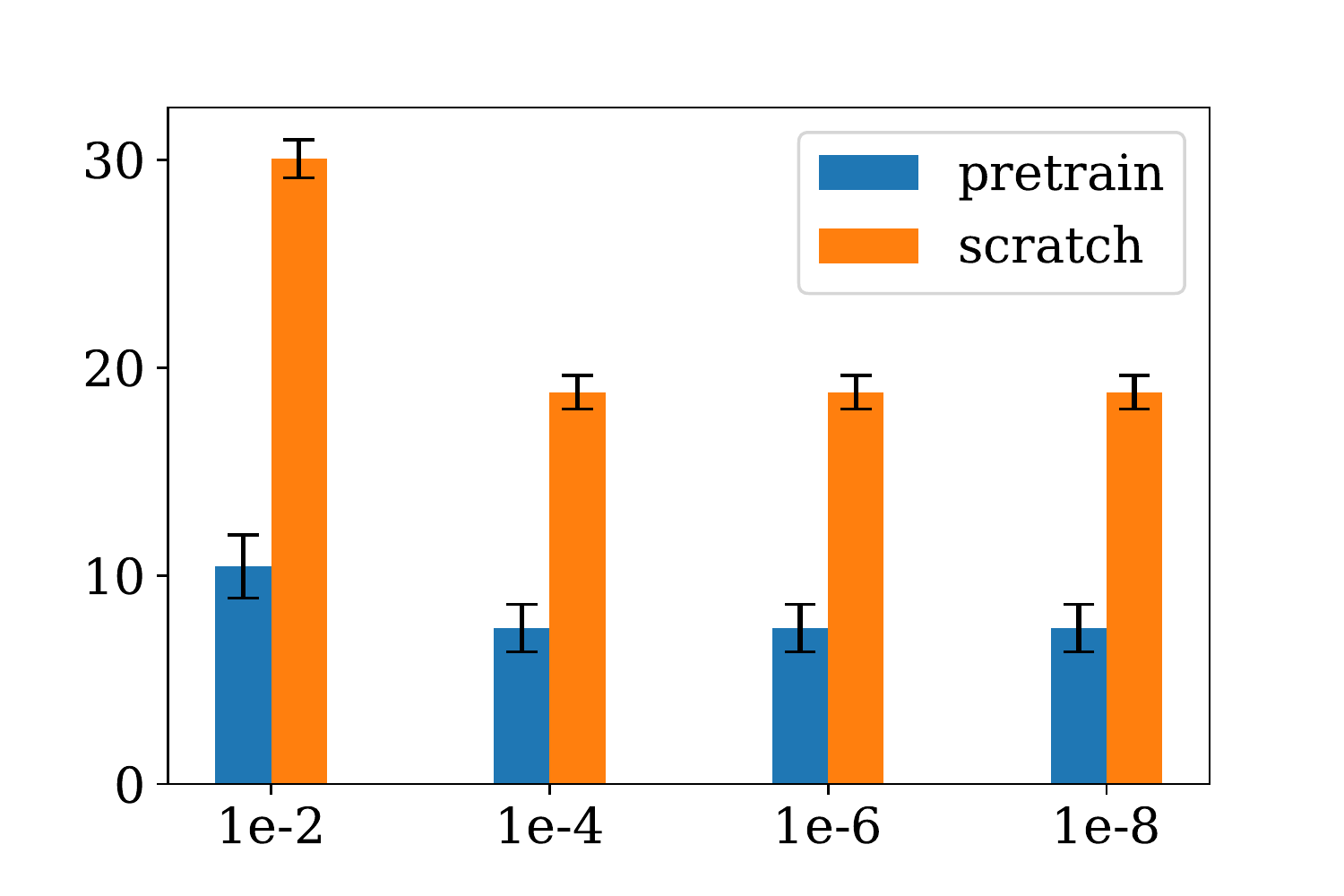}
        \caption{MNLI}
    \end{subfigure}
    \begin{subfigure}[t]{0.32\linewidth}
        \centering
        \includegraphics[width=\linewidth]{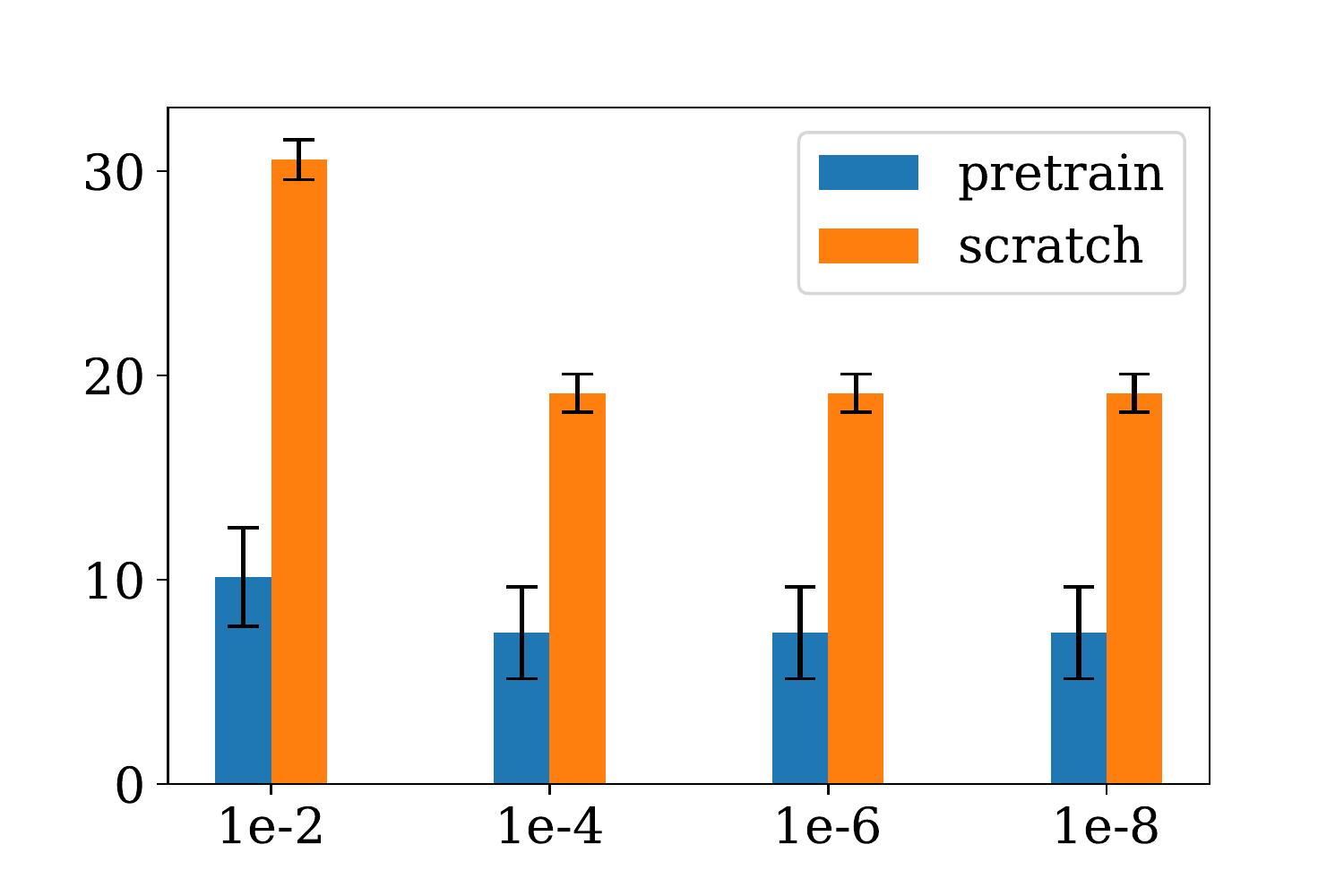}
        \caption{SST-2}
    \end{subfigure}
    \begin{subfigure}[t]{0.32\linewidth}
        \centering
        \includegraphics[width=\linewidth]{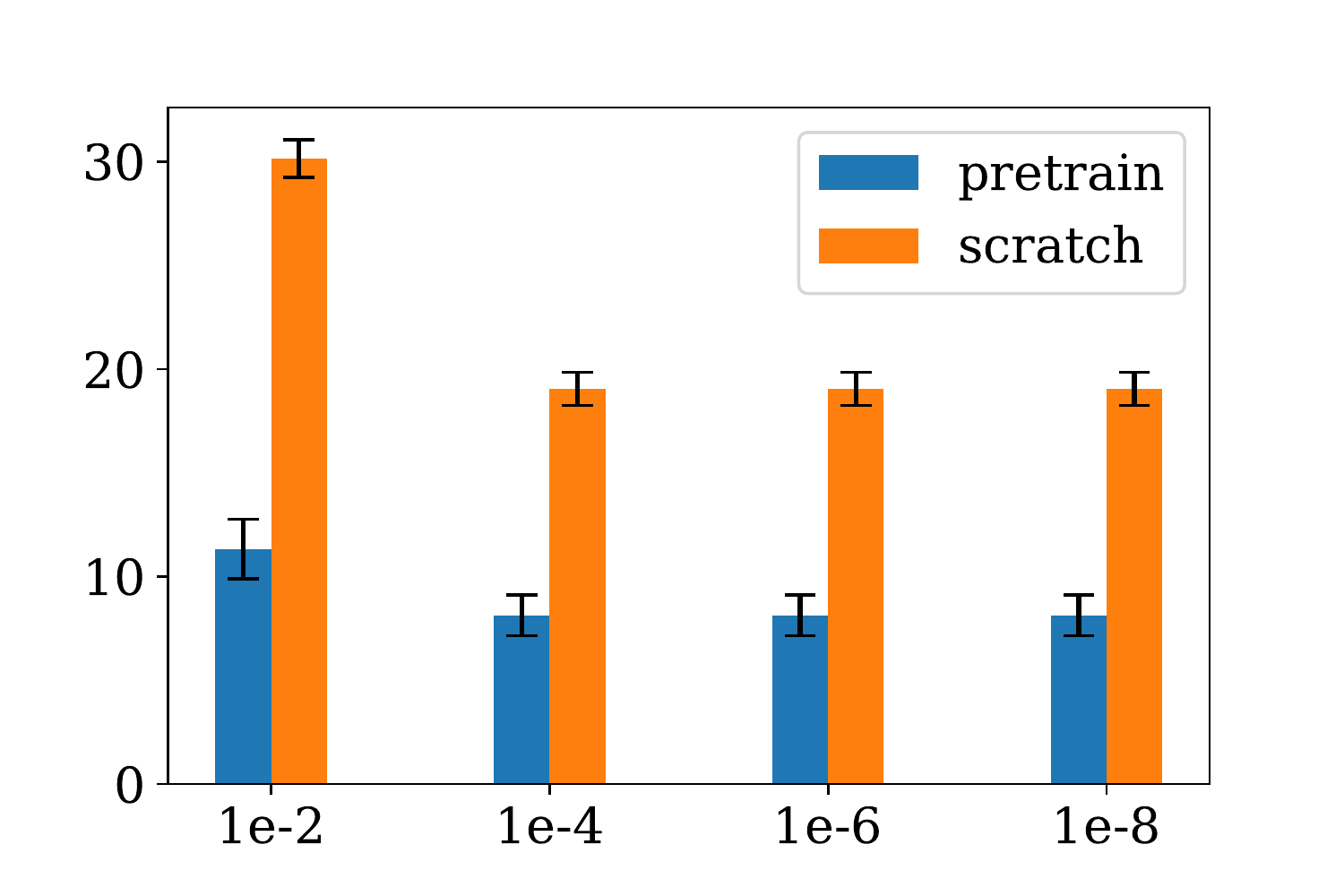}
        \caption{STS-B}
    \end{subfigure}
    \caption{The mean (bar) and std (error bar) of the L2 distance between BERT's outputs with and without adding noise to the model parameters. "scratch" stands for the randomly initialized parameters.}
    \label{app:per_bert}
\end{figure*}

\begin{figure*}[t]
    \centering
    \begin{subfigure}[t]{0.32\linewidth}
        \centering
        \includegraphics[width=\linewidth]{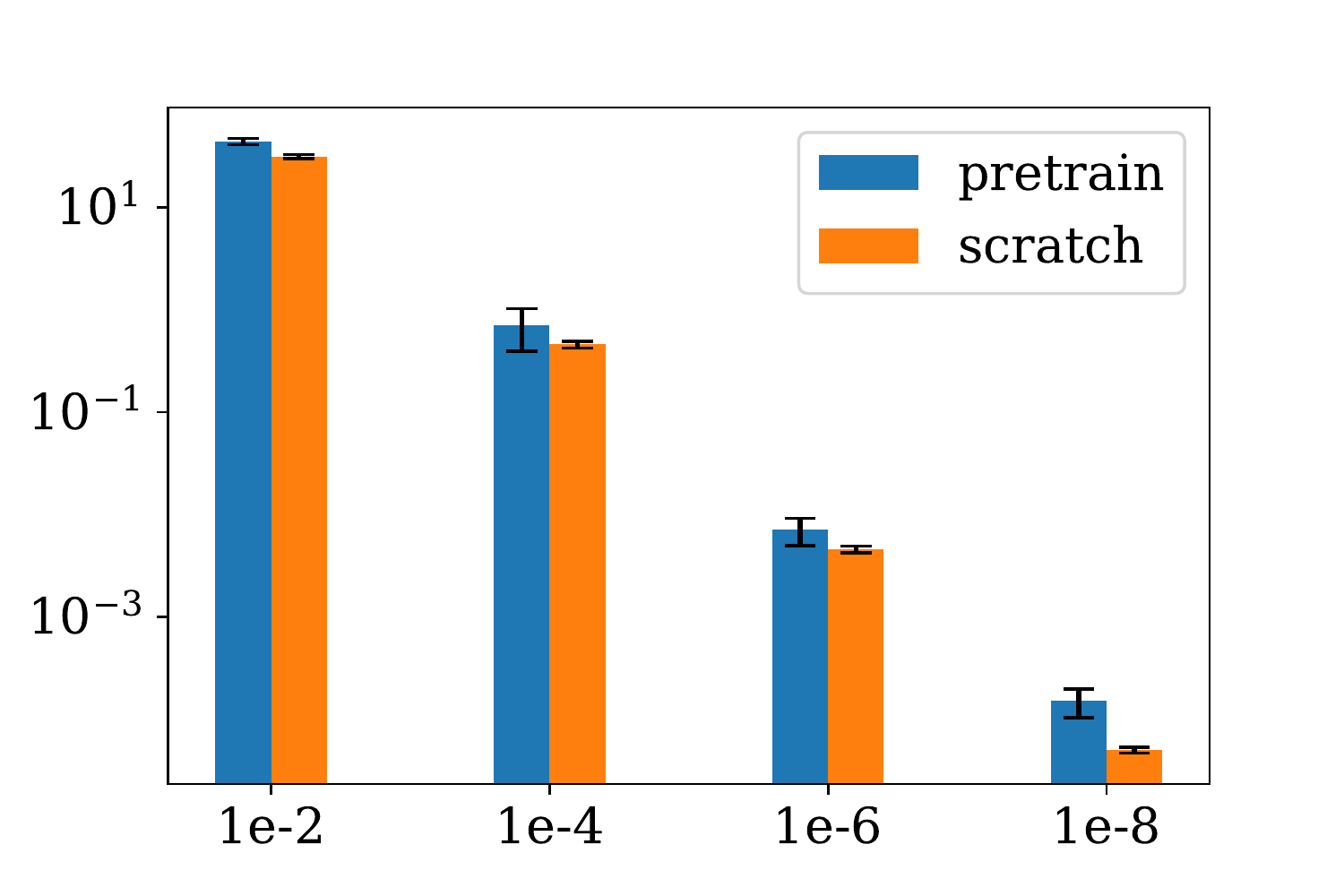}
        \caption{MNLI}
    \end{subfigure}
    \begin{subfigure}[t]{0.32\linewidth}
        \centering
        \includegraphics[width=\linewidth]{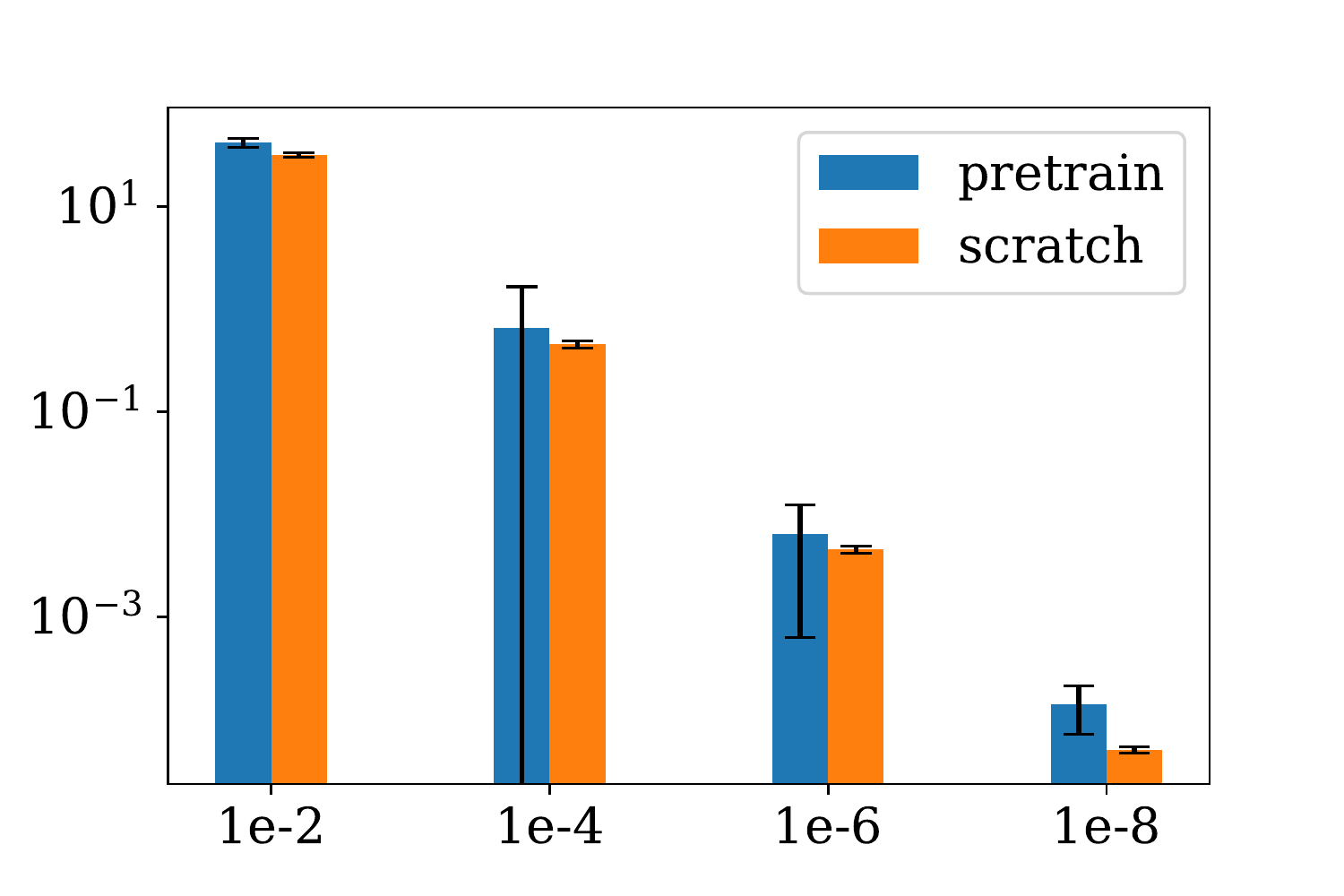}
        \caption{SST-2}
    \end{subfigure}
    \begin{subfigure}[t]{0.32\linewidth}
        \centering
        \includegraphics[width=\linewidth]{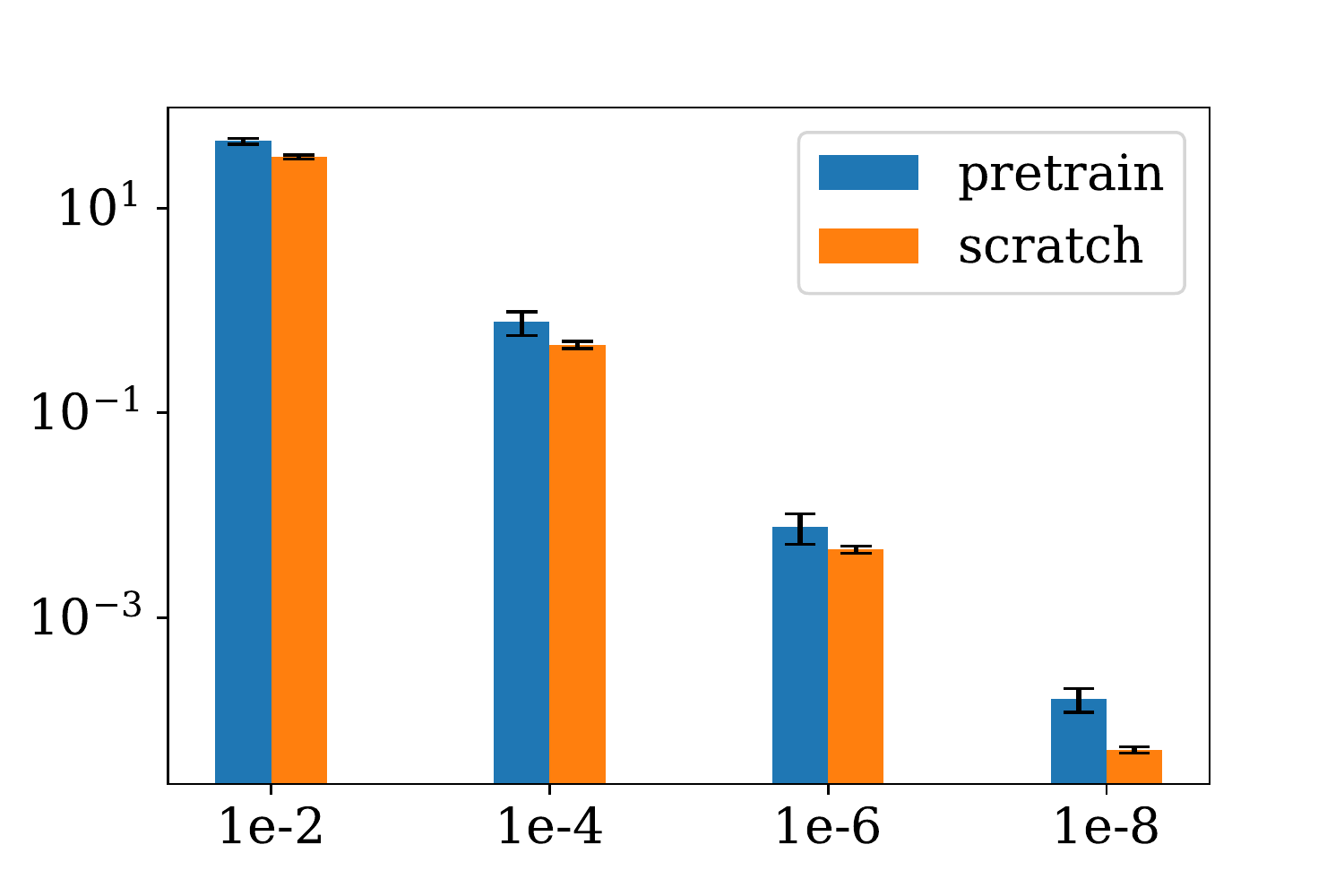}
        \caption{STS-B}
    \end{subfigure}
    \caption{The mean (bar) and std (error bar) of the L2 distance between ALBERT's outputs with and without adding noise to the model parameters. "scratch" stands for the randomly initialized parameters.}
    \label{app:per_albert}
\end{figure*}

We inject zero mean gaussian noise to the model parameters to calculate the variation of the model's outputs under the noise. The variation is represented by the L2 distance of the model's outputs before and after adding the noise. We choose the magnitude of standard deviation to be $10^{-2}, 10^{-4}, 10^{-6}$, and $10^{-8}$. Figure \ref{app:per_bert} and \ref{app:per_albert} show the results of BERT-base and ALBERT-base on the three synthetic GLUE tasks, respectively. We find that BERT and ALBERT show contrary trends: The variation of BERT is smaller than the randomly initialized counterpart, while the one of ALBERT is larger than the counterpart. So this hypothesis is not sufficient to explain the \textit{discipline adaptability} of the pre-trained models. 

\section{Statistics of datasets}
\subsection{GLUE}
\begin{table}[t]
\centering

\begin{tabular}{llll}
\toprule
  dataset  & train & validation \\
\midrule
  CoLA  &  8551 & 1043 \\
  SST-2 &  67349 & 872 \\
  MRPC & 3668 & 408 \\
  QQP & 363849 & 40430 \\
  STS-B & 5749 & 1500 \\
  MNLI & 392702 & 9815/9832 \\
  QNLI & 104743 & 5463 \\
  RTE & 2490 & 277 \\
\bottomrule
\end{tabular}
\caption{train/validation examples of GLUE dataset. The numbers of MNLI validation set are the matched subset and the mismatched subset respectively. Data can be downloaded at \url{https://gluebenchmark.com}}
\label{tab:glue_stat}
\end{table}

GLUE is an English dataset that consists of several tasks. Table \ref{tab:glue_stat} shows the statistics of GLUE. We use the validation set as the test set in our experiments. The train/validation split can be found in the downloaded data. 

\subsection{Protein classification}
\begin{table}[t]
\centering

\begin{tabular}{llll}
\toprule
  dataset  & train & validation & test\\
\midrule
  fluorescence & 21446 & 5362 & 27217\\
  stability & 53614 & 2512 & 12851\\
  localization & 9977 & 1108 & 2773\\
\bottomrule
\end{tabular}
\caption{train/validation/test examples of protein classification datasets. Data can be downloaded at \url{http://ailab.snu.ac.kr/PLUS/}. The train/validation/test can be found in the downloaded files.}
\label{tab:protein_stat}
\end{table}

Table \ref{tab:protein_stat} shows the statistics of protein classification datasets. For pre-processing, we truncate the length of input sequences to 512.

\subsection{DNA classification}

\begin{table}[t]
\centering

\begin{tabular}{ll}
\toprule
  dataset & \#samples\\
\midrule
  H3  &  14965 \\
  H4 &  14601 \\
  H3K9ac & 27782 \\
  Splice & 3190 \\
\bottomrule
\end{tabular}
\caption{Numbers of samples of DNA classification datasets. Data can be downloaded at \url{https://github.com/Doulrs/Hilbert-CNN}}
\label{tab:dna_stat}
\end{table}

Table \ref{tab:dna_stat} shows the statistics of DNA classification datasets. For the train/validation/test splits, we use randomly chosen 90\% samples as training data, 5\% samples as validation data, and 5\% samples as testing data as Hilbert-CNN does. We do not apply any additional pre-processing for these datasets.

\subsection{Music composer classification}
\begin{table}[t]
\centering

\begin{tabular}{llll}
\toprule
  dataset  & train & validation & test\\
\midrule
  MAESTRO-v1 & 954 & 105 & 125\\
\bottomrule
\end{tabular}
\caption{train/validation examples of MAESTRO-v1 dataset. Data can be downloaded at \url{https://magenta.tensorflow.org/datasets/maestro}}
\label{tab:maestro_stat}
\end{table}

Table \ref{tab:maestro_stat} shows the statistics of MAESTRO-v1 dataset. The train/validation/test splits can be found in the downloaded files. We read the midi data and convert it to pitch sequence. For sequences longer than 128, we divide them into several segments of length 128. For training data, each segment is one training example. For validation and testing, we inference on all the segments and decide the final output by voting.

\end{document}